\pdfoutput=1

\documentclass{article}
\usepackage{ijcai24}

\usepackage{times}
\usepackage{soul}
\usepackage{url}
\usepackage[hidelinks]{hyperref}
\usepackage[utf8]{inputenc}
\usepackage[small]{caption}
\usepackage{graphicx}
\usepackage{amsmath}
\usepackage{amsthm}
\usepackage{booktabs}
\usepackage[switch]{lineno}

\usepackage[linesnumbered,ruled,vlined]{algorithm2e}
\usepackage{amssymb}
\usepackage{color}
\usepackage{xspace}
\usepackage{bm}
\usepackage{natbib}
\usepackage{lipsum}
\usepackage{latexsym}

\newtheorem{lem}{Lemma}
\newtheorem{thm}{Theorem}
\newtheorem{dfn}{Definition}

\newtheorem{remark}{Remark}
\newtheorem{example}{Example}
\newtheorem*{thm*}{Theorem}

\newcommand{\algone}[1]{{\sf TS-episodic\xspace}}
\newcommand{\algtwo}[1]{{\sf TS-dynamic\xspace}}
\newcommand{\pregret}[1]{{problem-dependent regret}}
\newcommand{\Poi}{\mathrm{Poi}}
\newcommand{\Ga}{\mathrm{Ga}}

\newcommand{\e}{\mathrm{e}}
\newcommand{\expv}{\mathbb{E}}
\newcommand{\expvp}[2]{\mathbb{E}^{\pi}\left[\left. #1 \right|#2  \right]}

\newcommand{\rev}[3]{{\rm Rev}^{*}\left[#1, #2, #3\right]}

\DeclareMathOperator*{\argmin}{argmin}
\newcommand\blfootnote[1]{%
  \begingroup
  \renewcommand\thefootnote{}\footnote{#1}%
  \addtocounter{footnote}{-1}%
  \endgroup
}
\title{Learning with Posterior Sampling for Revenue Management under Time-varying Demand}

\author{
Kazuma Shimizu$^1$
\and
Junya Honda$^{2,3}$\and
Shinji Ito\footnote{
    He is currently affiliated with the University of Tokyo. 
}$^{1,3}$\And
Shinji Nakadai$^{1,4}$\\
\affiliations
$^1$NEC Corporation\\
$^2$Kyoto University\\
$^3$RIKEN AIP\\
$^4$Intent Exchange, Inc.\\
\emails
smzkzm2019@nec.com,
honda@i.kyoto-u.ac.jp,
shinji@mist.i.u-tokyo.ac.jp,
nakadai@intent-exchange.com
}
\begin{document}

\maketitle

\begin{abstract}
    This paper discusses the revenue management (RM) problem to
    maximize revenue by pricing items or services. 
    One challenge in this problem is that
    the demand distribution is unknown and varies over time
    in real applications such as airline and retail industries.
    In particular, the time-varying demand has not been well studied under scenarios of unknown demand
    due to the difficulty of jointly managing the remaining inventory and estimating the demand.
    To tackle this challenge, we first introduce an episodic generalization of the RM problem motivated by typical application scenarios. 
    We then propose a computationally efficient algorithm based on posterior sampling,
    which effectively optimizes prices by solving linear programming. 
   We derive a Bayesian regret upper bound of this algorithm for general models where demand parameters can be correlated between time periods, 
   while also deriving a regret lower bound for generic algorithms.
    Our empirical study shows that the proposed algorithm performs better than other benchmark algorithms
    and comparably to the optimal policy in hindsight. 
    We also propose a heuristic modification of the proposed algorithm, 
    which further efficiently learns the pricing policy in the experiments. \blfootnote{
        This paper is an extended version of the paper accepted by the 33rd International Joint
        Conference on Artificial Intelligence (IJCAI 2024). The appendixes can be found at the end of this paper.  
    }
\end{abstract}

%%%%%%%%%%%%%%%%%%%%%%%%%%%

\section{Introduction and Motivation}
Maximizing revenue by pricing items or services is a key problem in many industries such as airline and retail industries. 
This kind of problem is known as a price-based revenue management (RM) problem, which has been extensively studied in operations research and management science.
To determine optimal prices, it is essential to capture the relation between the selling price and demand, which is called a demand curve.
However, in the real world,
such a curve is not only unavailable to a seller in advance, but is also stochastic.  
Thus, maximizing revenue in real-world businesses requires the seller to deal with both unknown and stochastic demand.

Another challenge in the RM problem
is that demand may depend
not only on the offered price but also on the time.
Such a dynamic nature of the demand can particularly appear
in applications where sold items or services have some deadlines~
\citep{gallego1997multiproduct,su2007intertemporal}.
For example,
more business trippers reserve airplane seats as the departure date approaches 
as their schedules become clearer~\citep{lazarev2013welfare,williams2020dynamic}.

Despite its potential importance of addressing such dynamic demand,
most studies on RM problems with unknown demand have focused on the stationary demand
whose distribution solely depends on the price.
This discrepancy might come from the typical formulation of the problem,
where a seller experiences a single selling season.
In such a scenario, the seller can observe the actual demand for each time period only once,
and thus it is impossible to correctly estimate the future demand from the observed one.
One could mitigate this limitation if some model on the time dependency of the demand is assumed.
Still, it is not realistic to construct a model that precisely predicts future demands since the demand often drastically changes especially 
around the end or start of selling seasons. 
\citet{su2007intertemporal} discusses this dynamic demand changes 
in the fashion and travel industries.

A clue to address this difficulty from the practical viewpoint is that
a seller often has multiple seasons to sell items or services with an independent amount of inventories. 
In the case of the hotel industry, for example,
the demands for rooms of the same day of the week almost do not vary among weeks unless there is a special event near the hotel \citep{bandalouski2021dynamic}. 
Thus, the seller can learn demand for rooms by pricing them across multiple weeks.

In consideration of such applications, 
we introduce a generalization of the RM problem,
which we call {\it an episodic price-based revenue management problem}.
In this problem, we consider the setting where selling seasons (or episodes) are repeated multiple times as follows.
At the beginning of each selling season, a seller is given a fixed amount of inventory, which is not replenished during the season.
At each time period of a season, the seller prices the item and the number of sold items is determined according to
the demand whose distribution depends on the price and time period but is independent of the season.

In this episodic setting,
we can learn the demand distribution depending on both the price and time period if we have infinitely many seasons.
Our goal is to design a policy that efficiently explores the demand of each time period 
while exploiting the current knowledge to optimize the revenue in a finite number of seasons.
Such exploration-exploitation trade-off is much more complicated under the limited inventory with time-varying demand
since the desired prices between different time periods affect each other through the inventory constraint and the prior model on the time-varying demand.

\subsection{Our Contributions}

For the proposed generalization of RM problems, we develop two algorithms that combine
the pricing based on linear programming (LP)
with the technique of posterior sampling
called Thompson sampling \citep{thompson1933likelihood} in online-learning. 
This combination successfully leads to balancing the exploration-exploitation trade-off with inventory constraint
to efficiently achieve high performance.

The first algorithm, \algone{}, determines the schedule of prices during a season at the beginning of each season,
which is computationally efficient while having a reasonable theoretical guarantee.
The second algorithm, \algtwo{}, is a heuristic modification of \algone{}, which dynamically updates
the pricing at each time period of a season.
This modification can quickly reflect the observed data in the prices and enables robustness to stochastic demand. 
We numerically demonstrate that the proposed algorithms become close to the oracle policy that is optimal in hindsight as the number of seasons increases,
and this convergence is particularly fast under \algtwo{}.

On the theoretical contributions, we derive both upper and lower bounds on Bayesian regret. 
We first derive an upper bound on the Bayesian regret for \algone{} of $\mathcal{O}( T\sqrt{SK\log(K)} + S\sqrt{T})$, 
where $S$ is the number of episodes, $T$ is the selling horizon (the number of time periods) of each episode, and $K$ is the number of feasible prices.
We also derive a Bayesian regret lower bound for generic algorithms of $\Omega(T\sqrt{SK})$. 
The regret upper bound is sublinear in the total number $ST$ of time steps
when $S$ and $T$ jointly increase. 
In particular, the first term $\mathcal{O}(T\sqrt{SK})$
matches the lower bound up to a logarithmic factor and 
is unavoidable. 
The second term is linear in $S$, which can be regarded as a cost for computational efficiency
coming from the fact that the optimal pricing needs a dynamic programming with a very large table even if
the seller completely knows the demand distributions.
This kind of linear regret often appears in the settings where the problem involves a complicated optimization problem
(see the discussion below Theorem~\ref{thm:upper}).

 \subsection{Literature Review and Distinctions}
\label{sub:Lit}

The literature on RM problems primarily focuses on cases with known demand information. 
In such settings, RM problems with dynamic demand are well investigated by \citet{gallego1994optimal,gallego1997multiproduct,zhao2000optimal,anjos2005optimal,malighetti2009pricing}.
Other topics of price-based RM problems are reviewed in \citet{bitran2003overview, chiang2007overview}. 
RM problems with demand learning are studied later, which are reviewed in \citet{den2015dynamic}.
 In addition, there exist other types of revenue management problems called quantity-based revenue management. 
 We refer the reader to \citet{STRAUSS2018375, KLEIN2020397} and references therein for recent developments in such RM problems.

The episodic RM problem is related to reinforcement learning (RL) in a time-inhomogeneous Markov decision process (MDP)
\citep{hao2022regret,moradipari2023improved}. 
The regret analysis of the RL literature typically relies on an oracle to exactly compute the optimal pricing policy 
whereas our algorithms use LP for computational efficiency. 
This difference requires us to carefully evaluate the gap between the optimal pricing schedule and the approximated schedule by LP 
as will be discussed in Section~\ref{sec:regretana}.

For other episodic settings, \citet{den2015dynamic2} discuss an episodic RM for unknown stationary demand.
In addition, \citet{chen2022dynamic} consider a related episodic setting with unknown non-stationary demand.  
However, their setting allows inventory shortages (negative inventory) with a small extra cost,\footnote{
To be more specific, \citet{chen2022dynamic} consider the setting where the cost is incurred depending on the possibly negative 
remaining inventory at the end of the episode, and this cost is Lipschitz continuous in the remaining inventory.
}
meaning that inventory shortage does not drastically affect the revenue unlike our setting.  
Furthermore, their algorithm relies on more restricted demand conditions than ours,  
such as a sub-Gaussian demand distribution.

The most closely related work is \citet{ferreira2018online}, 
which focuses on a generalization of a RM problem with unknown stationary demand. 
They provide algorithms based on Thompson sampling
that balance the exploration-exploitation trade-off under inventory constraints. 
They derive upper bounds of Bayesian regret for their algorithms and demonstrate their outstanding performance in numerical experiments.
However, their algorithms heavily rely on the stationary demand setting, 
and extending them to non-stationary settings is highly non-trivial.

%%%%%%%%%%%%%%%%%%%%%%%%%

\section{Problem Setting}
\label{sec:problemsetting}
In this section, we formulate the episodic RM problem and propose algorithms with posterior sampling.

\subsection{Revenue Management Process}
\label{sec:revmana}
We consider the setting where a seller deals in a single item
and repeats selling seasons $S$ times.
Each selling season, indexed by $s \in [S]=\{1,2,\dots,S\}$, consists of $T \in \mathbb{Z}$ time periods.
At the beginning of each selling season, the seller has $n_{0}$ units of initial inventory, 
which is not replenished during the selling season. 
The inventory at the end of period $t$ is denoted by $n_{t;s}$, and $n_{0;s}=n_{0}$.
Each time period $t \in [T]$ of the $s$-th season consists of the following procedures:

(i) The seller chooses a price $P_{t;s}$ from the set of $K+1$ prices $\mathcal{P}\cup \{p_{\infty}\}$.
Here, $\mathcal{P}=\{p_k\}_{k \in [K]}\in[0,\infty)^{K}$ is a set of feasible prices  and
$p_{\infty}$ is a  ``shut-off" price, which is commonly used in dynamic pricing literature.
Under the shut-off price $p_{\infty}$, the demand is zero and no revenue is obtained with probability one.

(ii)
The seller observes random demand $D_{t;s}\ge 0$ that is independent of the past prices and demands.
The distribution under the offered price $p_k$ is denoted by $\mathcal{D}_{t,k}(\theta)$,
where $\theta\in\Theta$ is a parameter unknown to the seller.

(iii) The inventory is consumed according to the observed demand $D_{t;s}$ to yield revenue. 
The seller can consume at most $n_{t-1;s}$ units of the inventory
and the inventory at the end of the current time period is expressed as $n_{t;s} = \max( n_{t-1;s} - D_{t;s}, 0)$, 
which yields revenue $P_{t;s}\min(D_{t;s}, n_{t-1;s})$.

\begin{remark}{\rm
This formulation considers the case where there are units of only one item to sell,
whereas \citet{ferreira2018online} consider the case of multiple items.
We focus on the case of a single item just to highlight the
difficulty of the dynamic demand 
although the extension to the case of multiple items
is straightforward as discussed in Appendix~\ref{sec:multiple}.
}\end{remark}

We adapt a Bayesian approach in our demand model. 
We assume that a demand parameter $\theta\in\Theta$ follows some given prior distribution $f$.
For any $t \in[T]$ and $s \in [S]$, 
the posterior distribution is determined by the history $H_{t-1}^{s}$ of offered prices and observed demands up to the current time period of the season,
which is expressed as 
$H_{t-1}^{s} =
\{(P_{\tau ;\sigma}, D_{\tau;\sigma})\}_{\tau \in [T], \sigma \in [s-1]} \cup \{(P_{\tau ;s}, D_{\tau;s})\}_{\tau \in [t-1]}$. 
Given a history $H_{t-1}^{s}$, the posterior distribution is then expressed as $f(\theta |H_{t-1}^{s}) \propto \mathbb{P}(H_{t-1}^{s} | \theta) f(\theta)$, 
where $\mathbb{P}(H_{t-1}^{s} | \theta)$ is the likelihood function.

Note that we impose no assumption on the demand model $\{\mathcal{D}_{t,k}(\theta)\}_{t\in[T],k\in[K],\theta\in\Theta}$ and the prior distribution $f(\theta)$
as far as $f$ is a well-defined probability distribution (that is, an improper prior in the Bayesian statistics is not used).
In particular, we allow models where the demands among different prices and time periods are correlated.
This is a special strength of our framework capturing wide models, while causing technical difficulties 
since the estimator or the posterior distribution of the demand at each price and time period might complicatedly
depend on the past observations.
We introduce two examples of demand models below, both of which are used in the numerical analysis in Section~\ref{sec:Numa}.

\begin{example}[{\bf Poisson Demand with Independent Gamma Priors}]
\label{sec:exa1}
{\rm 
This is a simple model of the demand distributions,
in which
the demand distribution $\mathcal{D}_{t,k}(\theta)$ is expressed as $\mathcal{D}_{t,k}(\theta) = \Poi(\cdot|\lambda_{t,k})$ for $k \in [K]$ and time $t \in [T]$, 
where $\Poi(\cdot|\lambda)$ is a Poisson distribution with intensity $\lambda\in \mathbb{R}^+$.
The intensity parameters $\{\lambda_{t,k}\}_{t\in [T], k\in[K] }$ are
assumed to be independently and identically distributed by gamma distributions $\Ga(\alpha,\beta)$ with shape $\alpha>0$ and scale $\beta>0$.
Since gamma distributions are conjugate to Poisson distributions, 
the posterior distribution remains a gamma distribution that can be easily computed.
Still, this model requires estimation of $KT$ parameters $\theta=\{\lambda_{t,k}\}_{k\in[K],t\in[T]}$ independent of each other
and is not always sample-efficient in realistic settings. 
}
\end{example}

\begin{example}[{\bf Poisson Demand with Gaussian Process Prior}]
\label{sec:exa2}
{\rm
In this model, $\mathcal{D}_{t,k}(\theta)=\Poi(\cdot|\lambda_{t,k})$ is assumed
as in the first example,
but the intensity parameters $\{ \lambda_{t,k}\}_{t\in [T], k \in [K]}$ are modeled using a Gaussian process.   
To be more specific, this model assumes that the intensity parameter is expressed as
$\lambda_{t,k}= \exp(g(t,p_{k}))>0$ for $g(t,p): [T]\times \mathcal{P}\to \mathbb{R}$ following a Gaussian process
with some mean function $\mu(\cdot):[T]\times \mathcal{P}\to \mathbb{R}$ and kernel function $K(\cdot,\cdot): ([T]\times \mathcal{P}) \times ([T]\times \mathcal{P})\to \mathbb{R}$. 
Under this model, the posterior distribution of $\{\lambda_{t,k}\}_{t\in[T], k \in [K]}$ has no closed form 
but can be approximated by, e.g.,
Laplace approximation and Markov chain Monte Carlo method
(see Appendix~\ref{sec:gaussianP} 
 and \citet{10.7551/mitpress/3206.001.0001} for further details).
}
\end{example}

\subsection{Proposed Algorithms}

For the problem we stated so far, 
we propose two algorithms, \algone{} and \algtwo{}.
These algorithms use a mean demand function for a demand parameter $\theta \in \Theta$, 
which is defined as
\begin{linenomath}
\begin{align*}
  \lambda_{t,k}(\theta) = \mathbb{E}\left[D_{t;s}|P_{t;s} = p_{k}, \theta\right].
\end{align*}
\end{linenomath}
With this mean demand function,
the proposed algorithms solve the linear optimization problem
${\rm LP}(\theta,t, n)$ over
$\{x_{\tau,k}\}_{t \leq\tau\leq T,\ k\in[K]} \in [0,1]^{(T-t+1)K}$, defined as follows:
\begin{linenomath}
\begin{align}
   \text{maximize:\,} &\sum_{\tau =t}^{T}\sum_{k=1}^{K}x_{\tau,k}\lambda_{\tau,k}(\theta)p_{k}  \nonumber\\
    \text{subject to:\,} &\sum_{\tau=t}^{T}\sum_{k=1}^{K}x_{\tau,k}\lambda_{\tau, k}(\theta) \leq n, \nonumber \\
    &\sum_{k=1}^{K} x_{\tau,k} \leq 1, \quad \forall \tau \in \{t,t+1, \dots, T\}.
\label{def:LP}
\end{align}
\end{linenomath}
In this problem, $x_{\tau,k}$ intuitively corresponds to the probability of choosing price $p_k$ at time period $\tau$.
This optimization problem corresponds to a kind of
LP relaxation of the revenue optimization problem in our setting.
To be more specific, if the domain of $\{x_{\tau,k}\}_{t \leq\tau\leq T,\ k\in[K]}$ is
restricted to be binary and $D_{t,k}$ is deterministic then
\eqref{def:LP} gives the optimal pricing policy under a parameter $\theta$ 
when the inventory is $n$ at time period $t$. 
We use $\{x_{\tau,k}(\theta)\}_{\tau\in \{t,\dots,T\}, k\in[K]}$ to denote
the optimal solution of ${\rm LP}(\theta,t,n)$.

Although relying on linear programming instead of dynamic programming may result in a suboptimal algorithm, 
LP is commonly used particularly in bandit with knapsack problems for both stationary and non-stationary settings
 (see, for example, \citet{badanidiyuru2013bandits,immorlica2022adversarial,liu2022non}).

The first algorithm, \algone{}, is given in Algorithm~\ref{Fixed estimation on posterior sampling}.
At the beginning of the selling season, this algorithm randomly samples
a demand parameter $\theta_{s}$ from its posterior distribution.
Then,
the algorithm solves the ${\rm LP}(\theta_s, 1, n_{0})$ 
to obtain a solution $\{x_{\tau,k}(\theta_s)\}_{\tau\in \{1,\dots,T\}, k\in[K]}$.
At every time period $t$, \algone{} randomly chooses the prices according to this
solution computed at the beginning of the season.

\begin{algorithm}[t]
\caption{\algone{}}
\label{Fixed estimation on posterior sampling}
\DontPrintSemicolon
\For{$s=1,\dots, S$}
{
Sample a demand parameter $\theta_{s} \in \Theta $ from the posterior distribution $f(\cdot|H_{0}^{s})$ of $\theta$.\;
Solve LP$(\theta_s, 1,n_0)$.\;
\For{$t=1,\dots, T$}
{
Offer price $P_{t;s}= p_{k}$ with probability $x_{k,t}(\theta_{s})$
and $P_{t;s}=p_{\infty}$ with probability $1 - \sum_{k=1}^{K}x_{t,k}(\theta_s )$.\;
Observe realized demand $D_{t;s}$ and update the history as $H_{t}^{s} = H_{t-1}^{s} \cup \{P_{t;s}, D_{t;s}\}$.\;
}
}
\end{algorithm}

\begin{algorithm}[t]
\caption{\algtwo{}}
\label{Updating estimation on Posterior Sampling}
\DontPrintSemicolon
\For{$s=1,\dots, S$}
{
\For{$t=1,\dots, T$}
{
Sample a demand parameter $\theta_{t;s} \in \Theta $ from the posterior distribution $f(\cdot|H_{t-1}^{s})$ of $\theta$.\;
Solve LP$(\theta_{t;s}, t,n_{t-1;s})$.\;
Offer price $P_{t;s}= p_{k}$ with probability $x_{k,t}(\theta_{t;s})$
    and $P_{t;s}=p_{\infty}$ with probability $1 - \sum_{k=1}^{K}x_{t,k}(\theta_{t;s})$.\;
Observe realized demand $D_{t;s}$ and update the history  $H_{t}^{s} = H_{t-1}^{s} \cup \{P_{t;s}, D_{t;s}\} $.\; 
}
}
\end{algorithm}

The second algorithm, \algtwo{}, samples a parameter $\theta_{t;s}$ and solves ${\rm LP}(\theta_{t;s}, t, n_{t-1;s})$ at every time period.
The algorithm then determines the price according to $x_{t,k}(\theta_{t;s})$.
These procedures enable us to immediately exploit the demand observation at each time period.

\algone{} has a simpler structure than \algtwo{} since \algone{} samples a demand parameter 
and solves the LP only once at the beginning of each episode, whose solution is used throughout the season.
Still, due to the randomness of the offered price and the demand, 
the remaining inventory sometimes becomes unstable, which might cause discrepancy from the optimal pricing policy.
\algtwo{}, on the other hand, samples the demand parameter
and solves the LP based on the current inventory at each time period.
It thus can dynamically control the inventory during a selling season,
and can immediately exploit the demand information soon after the observation. 
These properties possibly enable \algtwo{} to learn demand faster than \algone{} and to have better performance 
at the cost of the computational time about $T$ times larger than that of \algone{}.

The proposed algorithms balance the exploration-exploitation trade-off through
the randomness of the samples from the posterior distribution with pricing by LP.
In particular, the important characteristic of the proposed algorithms is that
the future demand is taken into account through the LP when determining the price for the current time period.
Although pricing by LP is also considered in \citet{ferreira2018online}, 
it assigns the inventory uniformly into each time period, which results in no more optimal revenue under the time-varying demand. 
This difference is drastically reflected in the performances of the proposed algorithms and benchmarks as shown in Section \ref{sec:Numa}.

%%%%%%%%%%%%%%%%%%%%%%%%%%%%%%%%

\section{Regret Analysis}
\label{sec:regretana}

In this section, 
we first analyze the Bayesian regret of \algone{} 
and then derive a Bayesian regret lower bound for generic algorithms. 
The \pregret{} is the difference between the total expected revenue obtained by 
the algorithm and
that by the optimal policy $\pi^*(\theta)$ in hindsight, which is defined as follows.
 \begin{dfn}
 {\rm  The \pregret{} of an algorithm $\pi$ under a demand parameter $\theta \in \Theta$ is
 }
    \begin{align}
    \label{def:regret}
    {\rm Regret}\left(T, S, \theta, \pi\right) &= \sum_{s=1}^{S}\sum_{t=1}^{T}\mathbb{E}^{\pi^{*}(\theta)}\left[P_{t;s}\tilde{D}_{t;s}|\theta \right]\nonumber \\
    &\quad - \sum_{s=1}^{S}\sum_{t=1}^{T}\mathbb{E}^{\pi}\left[P_{t;s}\tilde{D}_{t;s}|\theta\right] \nonumber \\
    &= S {\rm Rev}^{*}\left(T,\theta\right) - {\rm Rev}^{\pi}\left(T,S,\theta\right),
    \end{align}  
{\rm where} $\tilde{D}_{t;s}=\min \{n_{t-1;s}, D_{t;s}\}$. 
  \end{dfn}
  Here, the optimal pricing policy in hindsight $\pi^{*}(\theta)$ serves as a chosen competitive algorithm.
  We then define the Bayesian regret as the expectation of the \pregret{} with respect to a prior distribution $f$ over $\theta$.
  \begin{dfn}{\rm
The Bayesian regret of an algorithm $\pi$ for a prior $f$ is}
\begin{linenomath}
  \begin{align}
  \label{def:Bregret}
  {\rm BRegret}(T,S,f,\pi) = \mathbb{E}_{\theta}[{\rm Regret}(T, S, \theta, \pi)],
  \end{align}
\end{linenomath}
  {\rm where $\mathbb{E}_{\theta}[\cdot]$ denotes the expectation taken for $\theta$ following $f$. }
  \end{dfn}
  
The Bayesian regret is a typical performance measure of online Bayesian algorithms. 
See \citet{russo2014learning} for further interpretations of Bayesian regret. 

\begin{thm}
\label{thm:upper}  

Assume that $4K \leq S$ and there exists $\bar{d} > 0$ such that  
the support of the distribution $\mathcal{D}_{t,k}(\theta)$ is finite and included in $[0, \bar{d}]$ for all $\theta$. 
Then, the Bayesian regret \eqref{def:Bregret} of \algone{} satisfies
\begin{linenomath}
\begin{align*}
{\rm BRegret}(T,S,f, \pi) & \leq p_{M}\bar{d}\bigg(S\sqrt{T} + 54T\sqrt{SK\log(K)}\bigg),
\end{align*}
\end{linenomath}
where  $p_{M} = \max_{k \in [K] }{p_k}$.
\end{thm}

From the upper bound of this theorem, we see that
the regret is sublinear in $ST$ when $S$ and $T$ jointly increase.
Here the first term of $\mathcal{O}(S\sqrt{T})$ is linear in $S$ but it can be regarded as a cost for computational efficiency.
Even if we know the true demand parameter $\theta$,
the optimal pricing policy $\pi^*(\theta)$
requires to compute dynamic programming,
which is sometimes costly in practice though it is polynomial time.
Furthermore, the optimal pricing policy $\pi^*(\theta)$ essentially depends on the demand distributions
$\{\mathcal{D}_{t,k}(\theta)\}_{t,k}$ themselves rather than their expectations.
On the other hand, the proposed algorithms use the linear programming,
which can be computed efficiently in practice by off-the-shelf solvers,
and behave stably since we only need to estimate the expected demands.

Note that this kind of linear regret often implicitly appears in the online learning problems
involving complicated optimization problems.
In such problems, the notion of $\alpha$-regret is often introduced instead,
which corresponds to the regret when the optimal algorithm $\pi^*(\theta)$ in \eqref{def:regret}
is replaced with an approximate algorithm with approximation ratio of $\alpha$ \citep{regret_linear,regret_influence}.
The regret of $\mathcal{O}(S\sqrt{T})$ corresponds to this gap between the optimal algorithm $\pi^*(\theta)$
and the approximation algorithm based on $\mathrm{LP}(\theta,1,n_0)$ for the true parameter $\theta$. 
\begin{remark}{\rm
The linear term might be avoidable if we exactly optimize the pricing schedule by dynamic programming instead of LP and combine techniques in RL literature.
They often consider the regret analysis under exact optimization of the schedule, which might be applicable
if we regard our problem as an MDP.
However, such an analysis is highly non-trivial since our model allows complex prior distributions 
unlike existing studies of RL literature that focus on independent or specific prior distributions \citep{osband2017posterior,lu2019information,hao2022regret,moradipari2023improved}. 
}
\end{remark}

\paragraph*{Key Points in the Proof of Theorem~1.}

The key challenge in the analysis of Theorem~\ref{thm:upper} is that the inventory constraint affects the offered prices and the total revenue in a very complicated way,
through the dynamic programming in the optimal policy in hindsight and the LP in the proposed policy.
Due to this challenge, standard regret analysis approaches for Thompson sampling algorithms, such as the one in \citet{russo2014learning},
cannot be directly applied to our problem.   
Though the inventory constraint is already considered in \citet{ferreira2018online},  
their argument is limited to the static demand 
because the best policy is to uniformly assign the inventory to each time period. 
To address this difficulty with the inventory constraint, 
we employ an approach to reduce the problem into $T$ instances of MAB problems with $K$ arms and $S$ rounds.
These $T$ problems are highly correlated through the inventory constraint and the posterior distribution,
but we show by a careful analysis that the regret can be decomposed into that of $T$ independent instances and that arising from the correlation.
We will give more details of the proof in Section~\ref{sec:proofsketch} (the formal proof is given in Appendix~\ref{sec:proofth1}
).

The following theorem ensures that the second term $\mathcal{O}(T\sqrt{SK\log K})$ of the regret upper bound in Theorem~\ref{thm:upper}, 
which is linear in $T$, is essential in the Bayesian regret:

\begin{thm}
  \label{thm:lower}  
  Assume the same condition as that in Theorem~1 and $n_{0}\geq \bar{d}$. Then, 
  there exists a demand model $\{\mathcal{D}_{t,k}(\theta)\}_{t\in[T],k\in[K], \theta\in\Theta}$ 
  and a prior distribution $f_0$ over $\Theta$ such that the Bayesian regret \eqref{def:Bregret} of any algorithm $\pi$ satisfies
\begin{linenomath}
  \begin{align*}
  {\rm BRegret}(T,S,f_0, \pi) \geq \Omega\left(T\sqrt{SK}\right).
  \end{align*}
\end{linenomath}
\end{thm}

\paragraph*{Key Points in the Proof of Theorem~2.}
The main difficulty in deriving this lower bound arises from the inventory constraint, 
which can vary the optimal price depending on the remaining inventory. 
This prohibits the use of common toolkits for evaluating lower bounds in multi-armed bandit problems.   
To address this challenge, we provide an instance where 
demand can appear in chosen $m=\left\lceil \frac{n_0}{\bar{d}} \right\rceil$ time periods and no demand in other time periods. 
This setting decomposes our problem into $m$ independent dynamic pricing tasks since the inventory never runs out.    
For each of the $m$ tasks,  
we can then use techniques for lower bound analysis of MAB problems (see, for example, \citet{lattimore2020bandit}).  
These techniques provide potential instances where any algorithm must incur at least $\Omega(\sqrt{SK})$ regret.     
The existence of such instances for each of the $m$ tasks allows us to choose a prior distribution $f$ for which the Bayesian regret 
is bounded below for any algorithms. 
The formal proof of Theorem~\ref{thm:lower} is given in Appendix~\ref{sec:proofofthm2}
.

\subsection{Proof Sketch of Theorem~1}

\label{sec:proofsketch}
First, we decompose the expected total revenue ${ \rm Rev}^{\pi}\left(T,S,\theta\right)$ into
the (virtual) total revenue ignoring the inventory limit
and the lost sales due to the lack of inventory.
To be more specific, we decompose the regret by

\begin{align}
\lefteqn{
\mathrm{BRegret}(T,S,f,\pi)
}
\nonumber\\
  &=
\sum_{t=1}^{T}\sum_{s=1}^{S}
\mathbb{E}_{\theta}\left[\mathbb{E}^{\pi^*(\theta)}\left[P_{t;s}\tilde{D}_{t;s} \bigg |\theta\right]- \mathbb{E}^{\pi}\left[P_{t;s}\tilde{D}_{t;s}\bigg|\theta\right]\right]
\nonumber \\
  &=
\
\underset{(A)}{\underline{\sum_{t=1}^{T}\sum_{s=1}^{S}\mathbb{E}_{\theta}\left[\mathbb{E}^{\pi^*(\theta)}\left[P_{t;s}\tilde{D}_{t;s} \bigg |\theta\right]- \mathbb{E}^{\pi}\left[P_{t;s}D_{t;s}\bigg|\theta\right]\right] }}
\nonumber \\
&\quad+\underset{(B)} {\underline{\sum_{s=1}^{S}
\mathbb{E}_{\theta}\left[\mathbb{E}^{\pi}\left[
\sum_{t=1}^{T}
\left(P_{t,s}D_{t;s} - P_{t;s}\tilde{D}_{t,s} \right) \Bigg| \theta \right] \right]
}}.
\label{eq:breg}
\end{align}
We refer to the underlined parts $(A)$ and $(B)$ as { \it the revenue-difference} and {\it the lost sales} parts, respectively. 
In the rest of this section, we will briefly sketch the derivation of upper bounds for these revenue-difference and lost sales parts. 
In what follows,  we will use the notation $\mathbb{E}^{\pi}\left[ \cdot \right] = \mathbb{E}_{\theta}\left[\mathbb{E}^{\pi}\left[\cdot | \theta\right]\right]$.

\subsubsection{Evaluation of the Revenue-difference Part}
As we will show in Lemma~\ref{lem:rev*vsoptLP} 
in Appendix~\ref{sec:regretdecomposed}
, ${\rm Rev}^{*}\left(T, \theta\right)$ is bounded above by the optimal value of ${\rm LP}(\theta, 1, n_{0})$. 
This fact allows us to have
\begin{linenomath}
    \begin{align*}
\!\!  (A) \leq \sum_{t=1}^{T}\mathbb{E}^{\pi}
\left[\sum_{s=1}^{S}\sum_{k=1}^{K}\left(x_{t,k}(\theta)-x_{t,k}(\theta_{s})\right)p_k \lambda_{t,k}(\theta) \right].
\end{align*}
\end{linenomath} 
However, the right-hand side of this inequality remains difficult to analyze since 
the solution of the LP depends on the inventory allocation across time periods.

To address this difficulty related to the complex inventory allocation, 
we introduce a set of upper confidence bounds $\{ U_{t,k;s}\}_{t\in[T], k\in[K], s\in[S]}$ (the definition of $U_{t,k;s}$ is given in Appendix~\ref{sec:defucb}
).
The upper confidence bound $U_{t,k;s}$ bounds the mean demand $\lambda_{{t,k}}(\theta)$ above 
with high probability. 
By combining these upper confidence bounds and the regret decomposition technique of \citet{russo2014learning}, we have 
\begin{align*}
\!\!  (A) &\leq \sum_{t=1}^{T}\left(\sum_{s=1}^{S}\sum_{k=1}^{K}  \left(\mathbb{E}^{\pi}\left[p_kx_{t,k}(\theta) \left(\lambda_{t,k}(\theta) - U_{t,k;s}\right)  \right]\right) \right.
\\ & \left. \quad\quad \quad\quad + \sum_{s=1}^{S}\sum_{k=1}^{K}\left(\mathbb{E}^{\pi}\left[\left(U_{t,k;s} - \lambda_{t,k}(\theta)\right)p_k x_{t,k}(\theta_s) \right]\right) \right). 
\end{align*}
For any fixed $t \in [T]$ and  $k\in[K]$, $U_{t,k;s}$ will decrease and converge to $\lambda_{{t,k}}(\theta)$ as the $k$-th price is offered.
Thus, both $\left(\lambda_{t,k}(\theta) - U_{t,k;s}\right)x_{t,k}(\theta)$ and 
$\left(U_{t,k;s} - \lambda_{t,k}(\theta)\right)x_{t,k}(\theta_s)$ can vanish as the number of seasons increases, 
which eventually results in an upper bound for $(A)$:
    \begin{align}
  \label{eq:boundA}
(A)  &\leq 18 p_{M}\bar{d}T\sqrt{SK\log(K)}.
\end{align}
The complete argument to derive this bound is given in the proof of 
Lemma~\ref{lem:revenuediff} 
in Appendix~\ref{sec:regretdecomposed}
.

\subsubsection{Evaluation of the Lost Sales Part}
The lost sales part in $(B)$ is first bounded by
$(B)\le
p_{M}\sum_{s=1}^{S}\mathbb{E}^{\pi}\left[
\left(\sum_{t=1}^{T}D_{t;s} - n_{0}\right)^{+}\right]$, 
where $x^{+} = \max\{x, 0\} $,
which is detailed in Appendix~\ref{sec:proofth1}
.
This expectation is further bounded by
\begin{align*} 
(B) \leq p_{M}\sum_{s=1}^{S}\underset{(B1)}{\underline {\mathbb{E}^{\pi}\left[
\left(\sum_{t=1}^{T}D_{t;s} -\mathbb{E}^{\pi}\left[\left. \sum_{t=1}^{T}D_{t;s} \right| \theta\right] \right)^+  \right]}}\nonumber  \\
 +p_M \underset{(B2)}{\underline{\sum_{s=1}^{S}\mathbb{E}^{\pi}\left[
\left(\mathbb{E}^{\pi}\left[\left.\sum_{t=1}^{T}D_{t;s} \right| \theta\right] - n_{0}\right)^{+}  \right]}},
\end{align*}
where the inequality follows from $(x+y)^{+} \leq x^{+} + y^{+}$. 
The underlined part $(B1)$ is bounded above by the conditional variance of $\sum_{t=1}^{T}D_{t;s}$ on $\theta$ from the Cauchy-Schwarz inequality, $ \mathbb{E}_{X}[|X|]^2 \leq \mathbb{E}_{X}[X^2]$. 
$(B1)$ is then bounded by an upper bound of the variance $\bar{d}\sqrt{T}$. Then, we have
  \begin{linenomath}
    \begin{align}
  \label{eq:lostupper1}
  (B1) \leq \bar{d}\sqrt{T}.
\end{align}
\end{linenomath} 
For the underlined part $(B2)$, recall that
the solution of ${\rm LP}(\theta_{t}, 1, n_{0})$ satisfies
$\sum_{t=1}^{T}\sum_{k=1}^{K}\lambda_{t,k}(\theta_{s})x_{t,k}(\theta_{s}) \leq n_{0}$
due to the inventory constraint. 
Combining this relation with $ \mathbb{E}^{\pi}\!\left[\left.\sum_{t=1}^{T}D_{t;s}  \right| \theta \right] 
= \sum_{k=1}^{K}\sum_{t=1}^{T} \lambda_{t,k}(\theta)x_{t,k}(\theta_s)$,  we have 
\begin{align*}  
(B2)
%&=
%\sum_{s=1}^{S}\mathbb{E}^{\pi}\left[
%\left(\sum_{t=1}^{T}\sum_{k=1}^{K}\lambda_{t,k}(\theta)x_{t,k}(\theta_{s}) - n_{0}\right)^{+}  \right]
%\nonumber\\
&\leq \sum_{t=1}^{T}\mathbb{E}^{\pi}\left[\sum_{s=1}^{S}\sum_{k=1}^{K}\left(\lambda_{t,k}(\theta)- \lambda_{t,k}(\theta_{s})\right)^{+}x_{t,k}(\theta_{s}) \right].
\end{align*}
Following a similar argument to that used in the derivation of \eqref{eq:boundA}, we obtain 
  \begin{align}
\label{eq:lostupper2}
(B2) \leq \bar{d}\left(36T\sqrt{SK\log(K)} \right). 
  \end{align}
The detailed derivation of this bound is given in Lemma~\ref{lem:lostsales2} 
in Appendix~\ref{sec:lostsalesanalysis}
.

From \eqref{eq:boundA}, \eqref{eq:lostupper1} and \eqref{eq:lostupper2},  
we can bound $(A)$ and $(B)$ in \eqref{eq:breg} and thus have proved Theorem~\ref{thm:upper}.

\begin{remark}{\rm
  Proving a theoretical upper bound is more challenging for \algtwo{} than for \algone{}. 
  This is because \algtwo{} repeatedly solves LP and the resulting total revenue depends on dynamical changes in remaining inventory. 
  This distinction complicates the regret analysis, 
  which requires a new technique for theoretical analyses. 
  However, we believe that \algtwo{} could have a theoretical bound of the Bayesian regret with the same order as that of \algone{} 
  due to its using LP and the lower bound in Theorem~2. 
  The additional evidence of this belief is the empirical results presented in Section~\ref{sec:Numa}, 
  where there is no significant difference in the performance between the two algorithms. 
 }
  \end{remark}

\section{Experiments}
\begin{figure*}[t!]
  \includegraphics[width=1.0\textwidth]{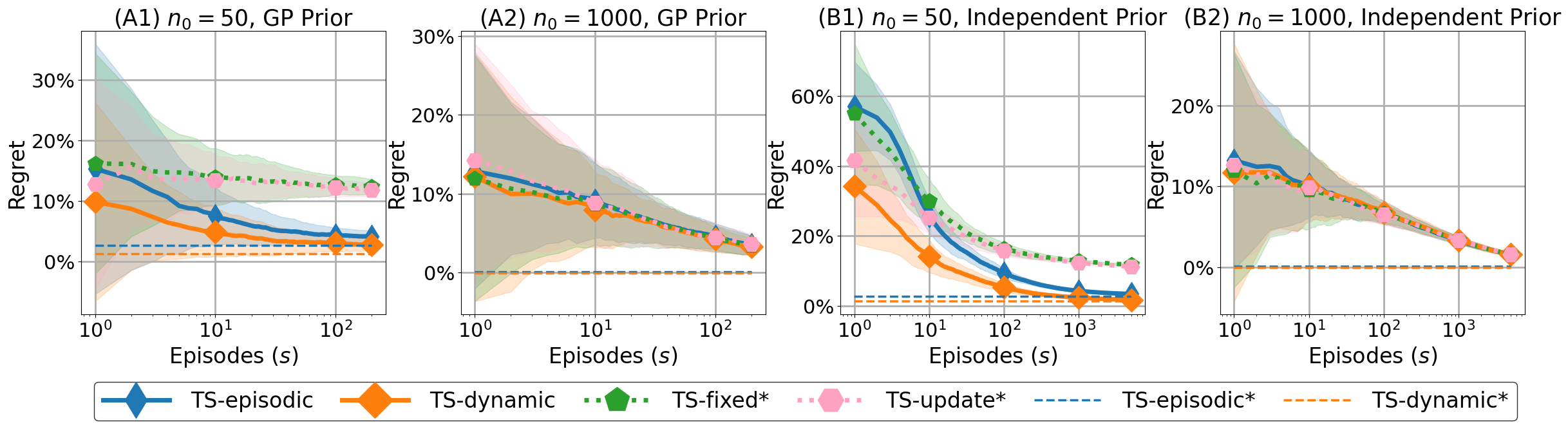}
     \caption{The numerical results for regret of \algone{}, \algtwo{}, {\sf TS-fixed}*, and {\sf TS updated}*,  \algone{}*, and \algtwo{}*. 
  (A1) and (A2) show the results for the GP prior, and (B1) and (B2) show the results for the independent prior.
   (A1) and (B1) show the results for $n_0=50$, and (A2) and (B2) show those for $n_0=1000$.
   The lines represent the averages of the regret and the shaded regions indicate the standard errors across independent $100$ trials. 
   The standard errors for \algone{}* and \algtwo{}* are omitted here and given in Appendix~\ref{sec:benches}
   .}
   \label{fig:bench1}
\end{figure*}

\label{sec:Numa}
In this section, we show numerical results\footnote{
The code of the experiments is available at: \url{https://github.com/NECDSresearch2007/RM-TSepisodic-and-dynamic}. 
} on the expected regret of the proposed algorithms and benchmark ones. 
Additional results are given in Appendix~\ref{sec:addnum}.

\subsubsection{Experimental Settings}
We consider the set of $K=9$ prices $\mathcal{P} = \{1,2,\dots,9\}$
with a shut-off price $p_{\infty}$.
The selling horizon is set to $T=10$.
The true demand distribution is set to Poisson distributions with mean demand parameters  
$\lambda(t,p) = 50\exp\left( - \frac{p + t}{5}\right)$, 
depending on the time $t$ and price $p$. 
These demand parameters can be viewed as an exponential-type demand curve \citep{gallego1994optimal}
with an exponentially decreasing coefficient with time,
which results in the discounted revenue case \citep{gallego1997multiproduct}. 
The initial inventory is set to
$n_0=1000$ and $50$, corresponding to the cases where there is enough inventory 
and where the inventory is quite limited, respectively.
In the former setting, pricing can be made almost independently between time periods 
and the problem becomes easier. 

For the prior on the demand distributions in the proposed algorithms and benchmarks,
we used the two models discussed in Examples 1 and 2, respectively.
The details of these priors are as follows.
\begin{description}
\item[Independent Gamma Prior:]
For example 1 in Section \ref{sec:revmana},  
  we set prior gamma distributions with shape $\alpha=10$ and scale $\beta=1$ for all $k\in [K]$ and $t\in [T]$.  
\item[Gaussian Process (GP) Prior:] For example 2 in Section \ref{sec:revmana},  
 we took the mean function $\mu$ as a zero function and the kernel function as
an anisotropic radial basis function kernel defined as,  
$K\left((p,t), (p',t')\right)= \exp\left( -(t-t')^2/\sigma^2_t  - (p-p')^2/\sigma^2_p \right)$ 
where $\sigma_t = 3$ $\sigma_p = 2.5$. 
\end{description}
We consider the number of episodes $S=5000$ for the settings with the independent gamma prior (referred to as independent prior hereafter) 
and $S=200$ for the GP prior.
We run independent $100$ trials for each setting.

\subsubsection{Comparison Targets}
In the considered setting, we can compute the optimal policy $\pi^*(\theta)$ using the dynamic programming
in manageable time though not practically efficient.
Then, we measure the performance of the algorithms based on the relative ratio
of the cumulative revenue compared with $\pi^*(\theta)$.

We compared the performance of the proposed algorithms with four benchmark algorithms.
The first two algorithms are generalizations of {\sf TS-fixed} 
and {\sf TS-update} in \citet{ferreira2018online} to our problem, which are
denoted by {\sf TS-fixed* } and {\sf TS-update*}, respectively.
To measure the gap between the performance of the optimal policy $\pi^*(\theta)$ and the policy based on linear relaxation,
we also consider two oracle algorithms, which are denoted by \algone{}* and \algtwo{}*.
In these algorithms, the true demand parameter is used instead of the one sampled from the posterior.
Then, \algone{}* and \algtwo{}* are independent of episodes and Bayesian settings by production. 
We compute their regret in an episode over $10000$ trials. 
The details of these benchmarks are given in Appendix~\ref{sec:benches}
.

\subsubsection{Numerical Results}
Figure \ref{fig:bench1} provides numerical results of the relative expected regrets
$1 - \allowbreak{\rm {Rev}}^{\pi}\left(T,s,\theta\right) / \left(s {\rm {Rev}}^{*}\left(T, \theta\right) \right)$
for $s\in[S]$.

For the case of $n_{0}=1000$ shown in Figures~\ref{fig:bench1} (A2) and (B2), 
our proposed algorithms and the benchmark algorithms show almost the same performance. 
This is because greedily optimizing the price at each round leads to the optimal policy 
and all the algorithms can learn efficient pricing without considering the inventory allocation across time periods.

Figures \ref{fig:bench1} (A1) and (B1) show the results for the case of $n_0=50$.
They demonstrate that the proposed algorithms can successfully learn an efficient pricing policy 
to maximize total revenue over the episode unlike the benchmark algorithms. 
In particular, \algtwo{} learns an efficient pricing policy faster than \algone{} 
and shows better performance as expected.

The results for \algone{}* and \algtwo{}* in Figure~\ref{fig:bench1} for $n_{0} = 50$
show that their expected regret does not become zero even under the knowledge of the true demand parameters,
which corresponds to the term
$\mathcal{O}(S\sqrt{T})$ in the regret bound in Theorem 1, which is linear in $S$.
This is the inevitable cost of relying on the linear optimization instead of dynamic programming. 

\begin{remark}{\rm
Whereas \algtwo{} learns an efficient policy faster than \algone{},
dynamically recomputing LP does not always contribute to a better allocation after demand parameters are learnt well.
In fact, we give a result in Appendix~\ref{sec:addnum} 
where \algtwo{}* becomes slightly worse than \algone{}* in certain settings. 
In this way, behavior of the allocation based on LP relaxation with recomputation becomes complicated even under known parameters, 
which makes the analysis of \algtwo{} particularly difficult.
}\end{remark}

We next discuss the effect of prior distributions. 
Our proposed algorithms with both the independent prior and the GP prior achieve almost the same performance level around the final episode.
However, the algorithms with the independent prior spend more episodes to learn an efficient pricing policy
than the ones with the GP prior. 
This result arises from the prior model where the mean demand function $\{\lambda_{t,k}(\theta)\}_{t \in [T], k \in [K]}$ must be independently estimated.
In contrast, our algorithms under the GP prior 
can learn faster through the kernel function utilizing the dependency of the demands between time periods.

%%%%%%%%%%%%%%%%%%%%%%%%%%%%%%%%%%%%%%%%%%

\section{Conclusion}

In this paper, we investigated a price-based revenue management problem, 
in which a seller tries to maximize the total revenue over a finite selling season with finite inventory of an item.
In particular, we considered the episodic scenario with unknown and time-varying demand with the real-world applicability in mind. 
For this problem, we proposed \algone{}, which combines the pricing based on linear programming relaxation
with posterior sampling.
We derived a regret guarantee for \algone{} and confirmed its effectiveness by numerical simulation.
We also proposed \algtwo{}, which is a heuristic modification of \algone{} and dynamically updates the posterior sample
and the pricing.
We numerically confirmed this algorithm can further quickly learn an effective pricing policy.

Finally, we present two relevant future directions.
The first direction is to find an algorithm that can achieve an upper bound of the Bayesian regret without the
$\mathcal{O}(S\sqrt{T})$ term. 
Such an algorithm would need to minimize lost sales as much as the optimal policy in hindsight does.  
The other direction is to find a precise theoretical analysis for \algtwo{}.  
However, this direction is challenging 
due to the difficulty of analyzing the algorithm repeating LP 
even if the true demand parameters are known as in \algtwo{}*. 
Therefore, it may be possible to analyze the difference of the regret of \algtwo{} 
compared with \algtwo{}* instead of that with the optimal hindsight policy, $\pi^{*}(\theta)$.

%%%%%%%%%%%%%%%%%%%%%%%%%%%%%%%%%

\section*{Acknowledgements}
The authors appreciate helpful and constructive comments for the anonymous reviewers.  
JH was supported by JSPS, KAKENHI Grant Number JP21K11747, Japan.

%%%%%%%%%%%%%%%%%%%%%%%%%%%%%%%%%
\bibliographystyle{named}
\bibliography{ijcai24}

\begin{thebibliography}{}

\bibitem[\protect\citeauthoryear{Anjos \bgroup \em et al.\egroup }{2005}]{anjos2005optimal}
Miguel~F Anjos, Russell~CH Cheng, and Christine~SM Currie.
\newblock Optimal pricing policies for perishable products.
\newblock {\em European Journal of Operational Research}, 166(1):246--254, 2005.

\bibitem[\protect\citeauthoryear{Audibert and Bubeck}{2010}]{audibert2010regret}
Jean-Yves Audibert and S{\'e}bastien Bubeck.
\newblock Regret bounds and minimax policies under partial monitoring.
\newblock {\em The Journal of Machine Learning Research}, 11:2785--2836, 2010.

\bibitem[\protect\citeauthoryear{Badanidiyuru \bgroup \em et al.\egroup }{2013}]{badanidiyuru2013bandits}
Ashwinkumar Badanidiyuru, Robert Kleinberg, and Aleksandrs Slivkins.
\newblock Bandits with knapsacks.
\newblock In {\em 2013 IEEE 54th Annual Symposium on Foundations of Computer Science}, pages 207--216. IEEE, 2013.

\bibitem[\protect\citeauthoryear{Bandalouski \bgroup \em et al.\egroup }{2021}]{bandalouski2021dynamic}
Andrei~M Bandalouski, Natalja~G Egorova, Mikhail~Y Kovalyov, Erwin Pesch, and S~Armagan Tarim.
\newblock Dynamic pricing with demand disaggregation for hotel revenue management.
\newblock {\em Journal of Heuristics}, 27(5):869--885, 2021.

\bibitem[\protect\citeauthoryear{Bitran and Caldentey}{2003}]{bitran2003overview}
Gabriel Bitran and Ren{\'e} Caldentey.
\newblock An overview of pricing models for revenue management.
\newblock {\em Manufacturing \& Service Operations Management}, 5(3):203--229, 2003.

\bibitem[\protect\citeauthoryear{Bubeck and Liu}{2013}]{bubeck2013prior}
S{\'e}bastien Bubeck and Che-Yu Liu.
\newblock Prior-free and prior-dependent regret bounds for thompson sampling.
\newblock {\em Advances in neural information processing systems}, 26, 2013.

\bibitem[\protect\citeauthoryear{Chen \bgroup \em et al.\egroup }{2022}]{chen2022dynamic}
Boxiao Chen, Menglong Li, and David Simchi-Levi.
\newblock Dynamic pricing with infrequent inventory replenishments.
\newblock {\em Available at SSRN 4240137}, 2022.

\bibitem[\protect\citeauthoryear{Chiang \bgroup \em et al.\egroup }{2007}]{chiang2007overview}
Wen-Chyuan Chiang, Jason~CH Chen, and Xiaojing Xu.
\newblock An overview of research on revenue management: current issues and future research.
\newblock {\em International journal of revenue management}, 1(1):97--128, 2007.

\bibitem[\protect\citeauthoryear{den Boer and Zwart}{2015}]{den2015dynamic2}
Arnoud~V den Boer and Bert Zwart.
\newblock Dynamic pricing and learning with finite inventories.
\newblock {\em Operations research}, 63(4):965--978, 2015.

\bibitem[\protect\citeauthoryear{Den~Boer}{2015}]{den2015dynamic}
Arnoud~V Den~Boer.
\newblock Dynamic pricing and learning: historical origins, current research, and new directions.
\newblock {\em Surveys in operations research and management science}, 20(1):1--18, 2015.

\bibitem[\protect\citeauthoryear{Ferreira \bgroup \em et al.\egroup }{2018}]{ferreira2018online}
Kris~Johnson Ferreira, David Simchi-Levi, and He~Wang.
\newblock Online network revenue management using thompson sampling.
\newblock {\em Operations research}, 66(6):1586--1602, 2018.

\bibitem[\protect\citeauthoryear{Gallego and Van~Ryzin}{1994}]{gallego1994optimal}
Guillermo Gallego and Garrett Van~Ryzin.
\newblock Optimal dynamic pricing of inventories with stochastic demand over finite horizons.
\newblock {\em Management science}, 40(8):999--1020, 1994.

\bibitem[\protect\citeauthoryear{Gallego and Van~Ryzin}{1997}]{gallego1997multiproduct}
Guillermo Gallego and Garrett Van~Ryzin.
\newblock A multiproduct dynamic pricing problem and its applications to network yield management.
\newblock {\em Operations research}, 45(1):24--41, 1997.

\bibitem[\protect\citeauthoryear{Garber}{2017}]{regret_linear}
Dan Garber.
\newblock Efficient online linear optimization with approximation algorithms.
\newblock In {\em Advances in Neural Information Processing Systems}, volume~30, 2017.

\bibitem[\protect\citeauthoryear{{Gurobi Optimization, LLC}}{2023}]{gurobi}
{Gurobi Optimization, LLC}.
\newblock {Gurobi Optimizer Reference Manual}, 2023.

\bibitem[\protect\citeauthoryear{Hao and Lattimore}{2022}]{hao2022regret}
Botao Hao and Tor Lattimore.
\newblock Regret bounds for information-directed reinforcement learning.
\newblock In {\em Advances in Neural Information Processing Systems}, volume~35, pages 28575--28587, 2022.

\bibitem[\protect\citeauthoryear{Hoeffding}{1963}]{hoeffding1963probability}
Wassily Hoeffding.
\newblock Probability inequalities for sums of bounded random variables.
\newblock {\em Journal of the American Statistical Association}, 58(301):13--30, 1963.

\bibitem[\protect\citeauthoryear{Immorlica \bgroup \em et al.\egroup }{2022}]{immorlica2022adversarial}
Nicole Immorlica, Karthik Sankararaman, Robert Schapire, and Aleksandrs Slivkins.
\newblock Adversarial bandits with knapsacks.
\newblock {\em Journal of the ACM}, 69(6):1--47, 2022.

\bibitem[\protect\citeauthoryear{Klein \bgroup \em et al.\egroup }{2020}]{KLEIN2020397}
Robert Klein, Sebastian Koch, Claudius Steinhardt, and Arne~K. Strauss.
\newblock A review of revenue management: Recent generalizations and advances in industry applications.
\newblock {\em European Journal of Operational Research}, 284(2):397--412, 2020.

\bibitem[\protect\citeauthoryear{Lattimore and Szepesv{\'a}ri}{2020}]{lattimore2020bandit}
Tor Lattimore and Csaba Szepesv{\'a}ri.
\newblock {\em Bandit algorithms}.
\newblock Cambridge University Press, 2020.

\bibitem[\protect\citeauthoryear{Lazarev}{2013}]{lazarev2013welfare}
John Lazarev.
\newblock The welfare effects of intertemporal price discrimination: an empirical analysis of airline pricing in us monopoly markets.
\newblock {\em New York University}, 2013.

\bibitem[\protect\citeauthoryear{Liu \bgroup \em et al.\egroup }{2022}]{liu2022non}
Shang Liu, Jiashuo Jiang, and Xiaocheng Li.
\newblock Non-stationary bandits with knapsacks.
\newblock {\em Advances in Neural Information Processing Systems}, 35:16522--16532, 2022.

\bibitem[\protect\citeauthoryear{Lu and Van~Roy}{2019}]{lu2019information}
Xiuyuan Lu and Benjamin Van~Roy.
\newblock Information-theoretic confidence bounds for reinforcement learning.
\newblock In {\em Advances in Neural Information Processing Systems}, volume~32, 2019.

\bibitem[\protect\citeauthoryear{Malighetti \bgroup \em et al.\egroup }{2009}]{malighetti2009pricing}
Paolo Malighetti, Stefano Paleari, and Renato Redondi.
\newblock Pricing strategies of low-cost airlines: The ryanair case study.
\newblock {\em Journal of Air Transport Management}, 15(4):195--203, 2009.

\bibitem[\protect\citeauthoryear{Moradipari \bgroup \em et al.\egroup }{2023}]{moradipari2023improved}
Ahmadreza Moradipari, Mohammad Pedramfar, Modjtaba Shokrian~Zini, and Vaneet Aggarwal.
\newblock Improved {B}ayesian regret bounds for {T}hompson sampling in reinforcement learning.
\newblock In {\em Advances in Neural Information Processing Systems}, volume~36, pages 23557--23569, 2023.

\bibitem[\protect\citeauthoryear{Osband and Van~Roy}{2017}]{osband2017posterior}
Ian Osband and Benjamin Van~Roy.
\newblock Why is posterior sampling better than optimism for reinforcement learning?
\newblock In {\em International Conference on Machine Learning}, pages 2701--2710. PMLR, 2017.

\bibitem[\protect\citeauthoryear{Rasmussen and Williams}{2005}]{10.7551/mitpress/3206.001.0001}
Carl~Edward Rasmussen and Christopher K.~I. Williams.
\newblock {\em {Gaussian Processes for Machine Learning}}.
\newblock The MIT Press, 11 2005.

\bibitem[\protect\citeauthoryear{Russo and Van~Roy}{2014}]{russo2014learning}
Daniel Russo and Benjamin Van~Roy.
\newblock Learning to optimize via posterior sampling.
\newblock {\em Mathematics of Operations Research}, 39(4):1221--1243, 2014.

\bibitem[\protect\citeauthoryear{Strauss \bgroup \em et al.\egroup }{2018}]{STRAUSS2018375}
Arne~K. Strauss, Robert Klein, and Claudius Steinhardt.
\newblock A review of choice-based revenue management: Theory and methods.
\newblock {\em European Journal of Operational Research}, 271(2):375--387, 2018.

\bibitem[\protect\citeauthoryear{Su}{2007}]{su2007intertemporal}
Xuanming Su.
\newblock Intertemporal pricing with strategic customer behavior.
\newblock {\em Management Science}, 53(5):726--741, 2007.

\bibitem[\protect\citeauthoryear{Thompson}{1933}]{thompson1933likelihood}
William~R Thompson.
\newblock On the likelihood that one unknown probability exceeds another in view of the evidence of two samples.
\newblock {\em Biometrika}, 25(3-4):285--294, 1933.

\bibitem[\protect\citeauthoryear{Wen \bgroup \em et al.\egroup }{2017}]{regret_influence}
Zheng Wen, Branislav Kveton, Michal Valko, and Sharan Vaswani.
\newblock Online influence maximization under independent cascade model with semi-bandit feedback.
\newblock In {\em Advances in Neural Information Processing Systems}, volume~30, 2017.

\bibitem[\protect\citeauthoryear{Williams}{2020}]{williams2020dynamic}
Kevin~R Williams.
\newblock Dynamic airline pricing and seat availability.
\newblock Technical report, Cowles Foundation for Research in Economics, Yale University, 2020.

\bibitem[\protect\citeauthoryear{Zhao and Zheng}{2000}]{zhao2000optimal}
Wen Zhao and Yu-Sheng Zheng.
\newblock Optimal dynamic pricing for perishable assets with nonhomogeneous demand.
\newblock {\em Management science}, 46(3):375--388, 2000.

\end{thebibliography}

\appendix
\onecolumn

\section{Summary of Notations}
\label{sec:notation}
Table \ref{tab:notations} summarizes the notations used in this paper.
\begin{table}[h!]
    \begin{tabular}{ll } \toprule 
    Notation    & Meaning \\ \midrule\midrule
    $[x]$ & the set of integers ranging from 1 to $x$\\
    $\mathbf{1} (P)$ & the indicator  function, which returns 1 if the proposition $P$ is true, or 0 otherwise \\
    $\log_{+}(x)$ &  a function defined as $\log(x) \mathbf{1}(x \geq 1)$ \\
    $\mathbb{E}$ & expectation \\
    \midrule\midrule
    $S$ & the number of selling season (episodes) \\
    $T$ & the number of time periods in each selling season \\
    $s$ & an index of a selling season in $[S]$ \\
    $t$ & an index of time period in $[T]$ \\
    $\mathcal{P}$ & the set of feasible prices \\
    $K$ & the number of the feasible prices \\
    $p_{k}$ & an element of $\mathcal{P}$ \\
    $p_{\infty}$ & a shut-off price \\
    $n_{0}$ & an initial inventory  provided at the beginning of each selling season\\ 
    $n_{t;s}$ & the remaining inventory at the end of time period $t$ on the $s$-th selling season\\ 
    $P_{t;s}$ & the offered price at time period $t$ on hte $s$-th selling season \\
    $D_{t;s}$ & the demand at time period $t$ on hte $s$-th selling season \\
    $\tilde{D}_{t;s}$ & actual sales, which is defined as $\tilde{D}_{t;s} = \min(D_{t;s}, n_{t-1;s})$ \\
    $H_{t-1}^{s}$ & a history available at the beginning of time period $t$ on the $s$-th selling season    \\
    $\Theta$ & a parameter space \\
    $\theta$ & a demand parameter of $\Theta $\\
    $\mathcal{D}_{t,k}(\cdot| \theta)$ & the distribution function followed by $D_{t;s}$ when the $k$-th price $p_{k}$ is offered \\ 
    $\lambda_{t,k}(\theta)$ & the mean of the demand distribution $\mathcal{D}_{t,k}(\cdot| \theta)$ \\
    ${ \rm LP}(\theta, t, n)$ & the linear optimization problem defined in \eqref{def:LP}\\
    $ f$ & a prior distribution defined on  $\Theta$ \\  
    $f(\cdot | H_{t-1}^{s})$ & the posterior distribution given history $H_{t-1}^{s}$\\
    $\theta_{t;s}$ & a sampled demand parameter from the posterior distribution given history $H_{t-1}^{s}$ \\
    $\{ x_{t,k}(\theta)\}_{t\in\{t, \cdots, T\}, k \in [K] 
    } $ & an optimal solution of ${\rm LP}\left(\theta, t, n_{t-1}\right)$ \\
    $\pi$ & an algorithm for the episodic revenue management \\
    $\pi ^{*}(\theta)$ & the optimal policy under a demand parameter $\theta$ in hindsight. \\
    ${\rm Rev}^{*}\left(T,\theta\right)$ &  the optimal expected revenue obtained by $\pi^{*}(\theta)$ when a selling contain $T$ time periods \\
    ${\rm Rev}^{\pi}\left(T,\theta\right)$ & the optimal expected revenue obtained by $\pi$ when a selling contain $T$ time periods given $\theta$\\
    ${\rm OPT_{LP}}\left(T, \lambda\right)$ & the value of ${ \rm LP}(\theta, 1, n_{0})$ \\
    ${\rm Regret}(S,T,\theta,\pi)$ & the problem dependent regret for an algorithm $\pi$ under a demand parameter $\theta$ \\
    ${\rm BRegret}(S,T,f,\pi)$ & the Bayesian regret for an algorithm $\pi$ for a prior $f$ \\
    \midrule\midrule
    $k(t;s) \in [K]$ & the index  of the price offered at time period $t$ on the $s$-th selling season \\
    $ N_{t,k}(s)$ &  the counts of times when $k$-th price is offered at the time period $t$ until $s$-th selling season \\
    $ \hat{D}_{t,k;s}$ & the normalized sample mean demand for $k$-th price at the time period $t$ on $s$-th selling season \\ 
    $\hat{D}_{t,k}(n)$ & the normalized sample mean demand with first $n$ samples at time $t$ for $k$-th price. \\
    $\beta_n$ & the confidence width depending on $n$, $S$ and $K$ defined on \eqref{eq:widthb} \\
    $\gamma_n$ & another confidence width depending on $n$, $S$ and $K$ defined \eqref{eq:widthg} \\
    $U_{t,k;s}$ & the upper confidence  bounds with $\beta_n$, defined in \eqref{def:UCB} \\
    $L_{t,k;s}$ & the lower confidence bounds with $\beta_n$, defined in \eqref{def:LCB} \\
    $\hat{U}_{t,k}$ & the upper confidence bounds with $\gamma_n$, defined in \eqref{def:hatUCB}  \\
    $\hat{L}_{t,k}$ & the lower confidence bounds with $\gamma_n$, defined in \eqref{def:hatUCB} \\ 
    \midrule\midrule
    $D(P,Q)$ & the KL-divergence of probability measures $P$ and $Q$\\
    ${\rm Regret_{sta}}(S,\theta_0, \pi)$ & the problem dependent regret under $\theta_0$ in the RM problem in static demand \\
    ${\rm BRegret_{sta}}(S, f_0, \pi)$ & the Bayesian regret for a prior $f_0$ in the RM problem in static demand \\
    \bottomrule
    \end{tabular}
     \caption{Summary of the notation used in this paper.}
     \label{tab:notations}
    \end{table}

\section{Definitions of Confidence Bounds}
\label{sec:defucb}
\label{sec:UCBdef}

In this appendix, we introduce the definitions of the confidence bounds that we will use in analysis of the Bayesian regret. 

To clarify the upcoming discussion in the proofs of lemmas, we first introduce upper confidence bounds based on the sample mean demand.
To define the sample mean demand,  let $\{ D_{t,k,s}\}_{s\in \mathbb{N}, t\in[T], k\in [K]}$ be a collection of independent random variables 
with the distribution of $D_{t,k,s}$ equal to $\mathcal{D}_{t,k}(\theta)$. 
This collection of random variables provides the alternative view of 
the demand observing process as follows:
for every $s \in [S]$, $k\in [K]$ and $t \in [T]$, $D_{t,k,s}$ is drawn from the corresponding distribution before the first selling season starts.
For each subsequent time period, the seller chooses a price without knowing the drawn values in advance.
The demand for the chosen price can be identified with $D_{t,k(t;s),N_{t,k(t;s)}(s)}$ where $k(t;s) \in [K]$ is the index of the arm chosen at time $t$ on the $s$-th selling season 
and $N_{t,k}(s)$ represents the counts of occasions when the $k$-th price is offered at time period $t$ until the $s$-th selling season. 
The definition of $N_{t,k}(s)$ is 
\begin{linenomath}
\begin{align*}
  N_{k;t}(s) = & \sum_{u=1}^{s} \mathbf{1}\left(P_{t,u} = p_{k}\right). 
\end{align*}
\end{linenomath}
Based on this observation, let $\hat{D}_{t,k}(n)$ denote the normalized mean demand for $k$-th price at the time period $t$ with $n$ samples, which is formulated as 
\begin{align*}
  \hat{D}_{t,k}(n) =& \frac{1}{\bar{d}} \frac{\sum_{u=1}^{n}D_{t,k,u}}{n}.
\end{align*} 
Using this definition, for all $k\in [K]$ and $t \in [T]$, we define $U_{t,k}(n)$ as a upper confidence bound for demand associated with the $k$-th price when the price appears $n$ times, which is expressed as 
\begin{align}
  \label{def:UCBtilde}
  \tilde{U}_{t,k}\left( n\right) =& \hat{D}_{t,k}(n) + \sqrt{\frac{\log_{+}\left(\frac{SK}{n}\right)}{n}},  \\
  U_{t,k}(n) =& \min(1, \tilde{U}_{t,k}\left( n\right)),
\end{align}
where $\log_{+}(x) = \log(x)\mathbf{1}\left(x\geq 1\right)$, and for later use, we write the confidence width as 
\begin{align}
\label{eq:widthb}
\beta_n = \sqrt{\frac{\log_{+}\left(\frac{SK}{n}\right)}{n}}, 
\end{align}
and define $U_{t,k}(0) = 1$. 

From these definitions, we will define an upper confidence associate with the sample mean of the demand at time period $t$ for the $k$-th price. 
Let $\hat{D}_{t,k;s}$ denote the normalized mean demand for $k$-th price at the time period $t$ until $s$-th selling season, which is formulated as 
\begin{align*}
  \hat{D}_{t,k; s} =& \frac{1}{\bar{d}} \frac{\sum_{u=1}^{s-1} \mathbf{1}\left(P_{t;u}=p_k\right)D_{t;u}}{N_{t,k}(s-1)}, 
\end{align*} 
which means that $\hat{D}_{t,k;s} = \hat{D}_{t,k}(N_{t,k}(s-1))$. 
We note that $\hat{D}_{t,k}(n)$ uses the number of samples, which is the constant, 
whereas $\hat{D}_{t,k;s}$ is a random variable.  
Consequently, we define an upper confidence bound for the demand associated with the $k$-th price up to time period $t$ in the $s$-th selling season as 
\begin{align}
\label{def:UCB}
U_{t,k;s} = U_{t,k}(N_{t,k}(s-1)). 
\end{align}
Note that $N_{t,k}(s-1)$ is completely determined by a given history at the beginning of the $s$-th selling season, $H_{0}^{s}=H_{T}^{s-1}$, and then $U_{t,k;s}$ is also determined.  
We use the notation $U_{t,k;s}(H_{0}^{s})$ to emphasize that the value is a constant under a given history $H_{0}^{s}$.

Similarly, let $L_{t,k}(n)$ denote a lower confidence bounds associated with $k$-th prices when the price appears $n$ times, which is expressed as 
\begin{align}
  \label{def:LCBtilde}
    \tilde{L}_{t,k}\left( n\right) =& \hat{D}_{t,k}(n) - \beta_{n}, \\
    L_{t,k}(n) =& \max(0, \tilde{L}_{t,k}\left( n\right)).
  \end{align}
Then, we define a lower confidence bound for the demand associated with the $k$-th price up to the $s$-th selling season as 
  \begin{align}
  \label{def:LCB}
  L_{t,k;s} = L_{t,k}(N_{t,k}(s-1)),
  \end{align}
We also define other confidence bounds for the demand associated with the $k$-th price when the price appears $n$ times, which is expressed as 
\begin{align}
\label{def:hatUCB}   
    \hat{U}_{t,k}(n) =& \hat{D}_{t,k}(n) + \sqrt{\frac{\log_{+}\left(\frac{S}{Kn}\right)}{n}},\quad 
    &&\hat{L}_{t,k}(n) = \hat{D}_{t,k}(n) -  \sqrt{\frac{\log_{+}\left(\frac{S}{Kn}\right)}{n}}. 
\end{align}
We will use $\gamma_n$ to express the confidence width, 
\begin{align}
\label{eq:widthg}
\gamma_{n} = \sqrt{\frac{\log_{+}\left(\frac{S}{Kn}\right)}{n}}.
\end{align}

\section{Lemmas for Theorem 1}
\subsubsection{Conditions Used in This Section}
Throughout this section, we use the same condition assumed in Theorem~\ref{thm:upper}:  we assume that there exists $\bar{d} > 0$ such that  
the support of the distribution $\mathcal{D}_{t,k}(\theta)$ is finite and included in $[0, \bar{d}]$ for all $\theta$ and that $4K \leq S$. 
In addition, we consider that prices are large in subscript order, $p_1<p_2<\dots<p_K$, for simplicity.

\subsubsection{Constractions of This Section}
Here, we provide lemmas toward the proof of Theorem 1  given in Appendix \ref{sec:proofth1}.
The main building blocks for the proof of Theorem~\ref{thm:upper} are composed of Appendices~\ref{sec:concentration}, \ref{sec:revenueanalysis} and \ref{sec:lostsalesanalysis} 
while Appendices~ from \ref{sec:def} and \ref{sec:basics} include basic definitions in random variables and basic lemmas, respectively. 
Lemmas in Appendix \ref{sec:regretdecomposed} use those in Appendices~\ref{sec:revenueanalysis} and \ref{sec:lostsalesanalysis} to prove Theorem~1.

\subsection{Definitions}
\label{sec:def}
Here, we define the property of $\sigma$-subgaussian to prove basic lemma in Appendix \ref{sec:basics}. 
We use the definition~5.2 in \citet{lattimore2020bandit}. 
\begin{dfn}[{\rm $\sigma$-subgaussian}]
  {\rm
  A random variable $X$ is $\sigma$-subgaussian if for all real number $h$, it satisfies $\mathbb{E}[\exp(h X)] \leq \exp(h^2 \sigma^2/2)$.
}  
\end{dfn}
\subsection{Basic Lemmas}
\label{sec:basics}
Here we introduce four preliminary lemmas.

%\label{sec:Basiclemmas}

\begin{lem} {(\rm Lemma 1 in \citet{hoeffding1963probability})} Let $X$ be any real-valued random variable with $\mathbb{E}[X]=0$ and $a \leq X \leq b$ almost surely for constants $a<b$. 
\label{Lem:Hoeflemm}
Then, $X$ is $\frac{b-a}{2}$-subgaussian. 
\end{lem} 
The proof of this lemma is given in exercise 5.11 in \citet{lattimore2020bandit}. 

\begin{lem} 
    \label{lem:lattimor}
    Let $X_{1}, X_{2}, \dots X_{m}$ be a sequence of independent $\sigma$-subgaussian random variables and $S_{n} = \sum_{s=1}^{n}X_{s}$. Then, for any positive number $x$.  
    We have 
    \begin{align*}
    \mathbb{P}\left(\exists\  n\in [m], \text{ s.t. }  S_n \geq x \right) &\leq \exp\left(-\frac{x^2}{2m\sigma^2} \right).
    \end{align*}
\end{lem}
This lemma is given in Theorem 9.2 of \citet{lattimore2020bandit}. 
The following lemma is a corollary of Lemma~\ref{lem:lattimor}:
\begin{lem} 
    \label{lem:hoeff}
    For any positive real value $x$ and integer $m \geq 1$, independent random variables $X_1, X_2, \dots X_{m}$ with mean $\mu = \mathbb{E}\left[X_{i}\right]$ and $0 \leq X_{i} \leq 1$  for $i =1 ,\dots, m$. 
    We have 
    \begin{align*}
    \mathbb{P}\left(\exists\  n \in  [m], \text{ s.t. }  \sum_{i=1}^{n}(\mu - X_i) \geq x \right) &\leq \exp\left(-\frac{2x^2}{m} \right), \\ 
    \mathbb{P}\left(\exists\ n \in [m], \text{ s.t. }  \sum_{i=1}^{n}( X_i - \mu) \geq x \right) &\leq \exp\left(-\frac{2x^2}{m} \right). 
    \end{align*}
\end{lem}

\begin{proof}
From the definition of $X_{i}$, $X_{i}-\mu$ is zero mean and $-\mu \leq X_{i}-\mu \leq 1-\mu$ for $i = 1, \dots, m$. 
By Lemma \ref{Lem:Hoeflemm}, $X_{i}-\mu$ for $i=1, \dots, m$ are independent $1/2$-subgaussian random variables, and $\mu - X_{i}$ 
are also independent $1/2$-subgaussian random variables.
Then, from Lemma \ref{lem:lattimor} we have obtained the desired bounds
\end{proof}

\begin{lem} ({\rm Hoeffding's Inequality \citep{hoeffding1963probability}})
\label{lem:hoeffdingineq}
Let $X_{1},\dots, X_{m}$ be independent random variables satisfying $0 \leq X_{i} \leq 1$ almost surely with mean $\mathbb{E}[X_i]=\mu$ for $i =1, \dots m$ and 
$S_{m}$ be the sum of these random variables, which is $S_{m} = \frac{1}{n}\sum_{n=1}^{m}X_{n}$. Then, for all positive real number $x$, we have 
\begin{align*}
\mathbb{P}\left(S_{m}- \mu \geq x\right) \leq \exp\left(- 2m x^2\right),\quad \mathbb{P}\left(\mu - S_{m} \geq x\right) \leq \exp\left(- 2m x^2\right). 
\end{align*}
\end{lem}

\subsection{Concentration Bounds}
\label{sec:concentration}
Here, we introduce two lemmas to bound the probabilities that 
the upper and lower confidence bounds are far apart from the true mean demand.  
These probability bounds are utilized in the lemmas in Appendix \ref{sec:revanalysis}, 
which bound expectation values. 

In the course of the proof of the following lemma, we use the argument in \citet{audibert2010regret} (see p2680-2681). 
  \begin{lem}
  \label{lem:eva*UCB*}
For any $s \in [S]$, $k \in[K]$ and $t\in [T]$, when $\delta_{0} = 2\sqrt{\frac{1}{KS}}$ and $u \in [\delta_{0}, 1]$, then we have 
\begin{align}
  \label{eq:lem5upper}
\mathbb{P}\left( \left. \left( \hat{\lambda}_{t,k}(\theta) - \tilde{U}_{t,k;s}  \right) \geq u \right| \theta\right)  \leq \frac{4}{SKu^2}\log\left(\sqrt{SK}u\right) + \frac{1}{KSu^2-1},\\ 
\mathbb{P}\left( \left. \left( \tilde{L}_{t,k;s} - \hat{\lambda}_{t,k}(\theta)  \right) \geq u \right| \theta \right)  \leq \frac{4}{SKu^2}\log\left(\sqrt{SK}u\right) + \frac{1}{KSu^2-1}.
\end{align}
\end{lem} 

\begin{proof}
Throughout the course of this proof, we evaluate the left-hand sides of the above inequalities conditionally on $\theta$ 
and then we will omit all the dependency on $\theta$ unless necessary. 
Using the monotonicity of a cumulative distribution function, we have 
\begin{align*}
  \mathbb{P}\left(\left( \hat{\lambda}_{t,k} - \tilde{U}_{t,k;s}\right) \geq u \right) \leq \mathbb{P}\left(\left(\hat{\lambda}_{t,k}- \min _{1\leq s\leq S}\tilde{U}_{t,k}(s)\right) \geq u \right).
\end{align*}
Let $f(u) = \lfloor \frac{2}{u^2}\log\left(u\sqrt{SK}\right) \rfloor $ where $\lfloor x \rfloor$ is the greatest integer less than or equal to $x$. 
This function $f(u)$ is non-increasing in $u \in [\delta_0, 1]$ and then is bounded as
\begin{align}
\label{eq:fubound}
f(u) \leq \left\lfloor \frac{KS\log\left(2\right)}{2} \right\rfloor \leq KS.  \quad (\lfloor x \rfloor \text{ is non-decreasing in } x)
\end{align}
Following the result in \citet{audibert2010regret}, which appears in the third step of the proof of Theorem 24, we have 
\begin{align}
  \label{eq:peeling}
  \mathbb{P}\left(\left(\hat{\lambda}_{t,k}- \min _{1\leq s \leq S}\tilde{U}_{t,k}(s)\right) \geq u \right) &= 
  \mathbb{P}\left(\exists s_{0}  \in \{1, 2, \dots,S\}, \text{ s.t. }  \hat{\lambda}_{t,k} - \tilde{U}_{t,k}(s_0) \geq u  \right) \nonumber \\
  &\leq \mathbb{P}\left(\exists s_0 \in  \{ 1,2, \dots, f(u)\}, \text{ s.t. }  \hat{\lambda}_{t,k} - \tilde{U}_{t,k}(s_0) \geq u  \right) \nonumber \\
  &\quad + \mathbb{P}\left(\exists s_0  \geq  f(u) + 1, \text{ s.t. }  \hat{\lambda}_{t,k} - \tilde{U}_{t,k}(s_0) \geq u  \right). 
\end{align}
To evaluate the first term, we divide the interval $[0,f(u)]$ into infinite pieces, each of which is in the form $[f(u)/2^{m+1}, f(u)/2^{m}]$ for an non-negative integer $m$.
When $s_0\leq f(u)$, we have $\log_{+}\left(SK/s_0\right) = \log\left(SK/s_{0}\right)$, and then obtain
\begin{align*}
&\mathbb{P}\left( \exists s_0 \in \{ 1,2, \dots,  f(u)\}, \text{ s.t. }  \hat{\lambda}_{t,k} - \tilde{U}_{t,k}(s_0) \geq u  \right) 
\nonumber \\
&\leq \sum_{m=0}^{\infty} \mathbb{P}\left(\exists s_0 \in \left[\frac{f(u)}{2^{m+1}}, \frac{f(u)}{2^{m}}\right], \text{ s.t. }  \hat{\lambda}_{t,k} - \tilde{U}_{t,k}(s_0) \geq u  \right) 
\nonumber \\
&=  \sum_{m=0}^{\infty}\mathbb{P}\left(\exists s_0 \in  \left[\frac{f(u)}{2^{m+1}}, \frac{f(u)}{2^{m}}\right], \text{ s.t. }  \frac{1}{s_0}\sum_{u=1}^{s_0}\left(\hat{\lambda}_{t,k}-\hat{D}_{t;u}\right) \geq \sqrt{\frac{\log\left(\frac{SK}{s_0}\right)}{s_0}}  \right) &&(u \geq 0  ) 
\nonumber  \\
&\leq \sum_{m=0}^{\infty}\mathbb{P}\left(\exists s_0 \in \left[ \frac{f(u)}{2^{m+1}},  \frac{f(u)}{2^{m}} \right], \text{ s.t. }  \sum_{u=1}^{s_0}\left(\hat{\lambda}_{t,k}-\hat{D}_{t;u}\right) \geq \sqrt{\frac{f(u)}{2^{m+1}}\log\left(\frac{2^{m}SK}{f(u)}\right)}  \right)\quad &&(\text{by the monotonicity of } \log x, \text{and} \sqrt{x}  )
\nonumber \\
&\leq \sum_{m=0}^{\infty}\exp\left(-2\frac{\frac{f(u)}{2^{m+1}}\log\left(\frac{2^m SK}{f(u)}\right)}{\frac{f(u)}{2^m}} \right) &&  (\text{by  Lemma \ref{lem:hoeff}}) 
\nonumber \\
&= \frac{f(u)}{SK}\sum_{m=0}^{\infty}\frac{1}{2^m} \leq \frac{4}{SKu^2}\log\left(u\sqrt{SK}\right). && \left( f(u) \leq \frac{2\log\left(u\sqrt{SK}\right)}{u^2} \right)
\end{align*}

The second term in \eqref{eq:peeling} can also be evaluated by dividing the interval $[f(u),\infty)$ into infinitely many pieces 
with the grides of the form $2^m g(u) \leq s \leq 2^{m+1}g(u)$, where $g(u) = f(u)+1$, as 
\begin{align*}
&\mathbb{P}\left(\exists s_0  \geq  g(u), \text{ s.t. }  \hat{\lambda}_{t,k} - \tilde{U}_{t,k}(s_0) \geq u  \right)&& \nonumber\\
&= \sum_{m=0}^{\infty}\mathbb{P}\left(\exists s_0 \in  \left[2^{m}g(u),  2^{m+1}g(u)\right], \text{ s.t. }  \hat{\lambda}_{t,k} - \tilde{U}_{t,k}(s_0) \geq u  \right) \nonumber \\
&\leq \sum_{m=0}^{\infty}\mathbb{P}\left(\exists  s_0 \in \left[ 2^{m}g(u),  2^{m+1}g(u) \right], \text{ s.t. }  \frac{1}{s_0}\sum_{u=1}^{s_0}(\hat{\lambda}_{t,k} - \hat{D}_{t;u})\geq u + \sqrt{\frac{\log_{+}\left(\frac{SK}{s_0}\right)}{s_0}} \right)\nonumber \\
&\leq \sum_{m=0}^{\infty}\mathbb{P}\left(\exists  s_0 \in \left[2^{m}g(u),  2^{m+1}g(u)\right] \text{ s.t. }  \frac{1}{s_0}\sum_{u=1}^{s_0}(\hat{\lambda}_{t,k} - \hat{D}_{t;u})\geq  u \right) &&\left(\sqrt{\frac{\log_{+}\left(\frac{SK}{s_0}\right)}{s_0}} \geq 0 \right)\nonumber \\
&\leq \sum_{m=0}^{\infty}\mathbb{P}\left(\exists  s_0 \in \left[2^{m}g(u),  2^{m+1}g(u)\right], \text{ s.t. }  \sum_{u=1}^{s_0}(\hat{\lambda}_{t,k} - \hat{D}_{t;u})\geq 2^{m}g(u)u \right) &&\left(  s_{0} \geq 2^mg(u) \right) \nonumber  \\
&\leq \sum_{m=0}^{\infty}\exp\left(-2 \frac{\left(2^{m}g(u)u\right)^2}{2^{m+1}g(u)}\right) && \text{(By Lemma \ref{lem:hoeff})}\nonumber\\
&= \sum_{m=0}^{\infty}\exp\left(-\left(2^{m}g(u)u^2\right)\right) \\
&\leq \sum_{m=0}^{\infty}\exp\left(-(m+1)g(u)u^2 \right) &&\left(m+1 \leq 2^m, m \in \mathbb{N} \cup \{0\}\right) \nonumber \\
&= \frac{1}{\exp\left(g(u)u^2\right) - 1} \leq \frac{1}{KSu^2 - 1}. &&\left( 2\log\left(u\sqrt{SK}\right) \leq g(u)u^2. \right)
\end{align*}
Consequently, we have shown \eqref{eq:lem5upper}.
Next, we consider the probability that the lower confidence $\tilde{L}_{t,k;s}$ overestimate $\hat{\lambda}_{t,k}(\theta)$ by $u$. 
For any $k \in [K]$, we have
\begin{align*} 
  \mathbb{P}\left(  \tilde{L}_{t,k;s} - \hat{\lambda}_{t,k}(\theta) \geq u \right) \leq \mathbb{P}\left(\left(\max_{1\leq s \leq S} \tilde{L}_{t,k;s} - \hat{\lambda}_{t,k} \right) \geq u \right).
\end{align*}
Following the same procedure as in the evaluation of  $\mathbb{P}\left(\hat{\lambda}_{t,k}- \min_{1\leq s \leq S} \tilde{U}_{t,k}(s) \geq u\right)$, we obtain 
\begin{align*}
  \mathbb{P}\left(\max_{1\leq s \leq S}\tilde{L}_{t,k;s} - \hat{\lambda}_{t,k}(\theta)  \geq u \right)  &\leq \mathbb{P}\left(\exists  s_0 \in  \left[1, f(u)\right], \text{ s.t. }  -\hat{\lambda}_{t,k} +  \tilde{L}_{t,k}(s_0) \geq u  \right) \nonumber \\
  &\quad + \mathbb{P}\left(\exists s_0 \geq f(u) + 1  , \text{ s.t. }  -\hat{\lambda}_{t,k} + \tilde{L}_{t,k}(s_0) \geq u  \right),
\end{align*}
and then have
\begin{align*}
\mathbb{P}\left(\exists s_0 \in \left[1, f(u)\right], \text{ s.t. }  -\hat{\lambda}_{t,k} + \tilde{L}_{t,k}(s_0) \geq u  \right) 
&\leq  \sum_{m=0}^{\infty} \mathbb{P}\left(\exists s_0 \in \left[\frac{f(u)}{2^{m+1}}, \frac{f(u)}{2^{m}}\right], \text{ s.t. }  \frac{1}{s_0}\sum_{u=1}^{s_0}\left(\hat{D}_{t;u} - \hat{\lambda}_{t,k}\right) \geq \sqrt{\frac{\log\left(\frac{SK}{s_0}\right)}{s_0}}  \right),
\\
\mathbb{P}\left(\exists  s_0 \geq  f(u) + 1 , \text{ s.t. }  -\hat{\lambda}_{t,k} + \tilde{L}_{t,k}(s_0) \geq u  \right) 
&\leq \sum_{m=0}^{\infty}\mathbb{P}\left(\exists s_0 \in \left[2^{m}g(u),  2^{m+1}g(u)\right], \text{ s.t. }  \sum_{u=1}^{s_0}(\hat{D}_{t;u}- \hat{\lambda}_{t,k})\geq 2^{m}g(u)u \right). 
\end{align*}
Using Lemma \ref{lem:hoeff} to the right-hand sides of the first and second inequality,  
we find that $\mathbb{P}\left(  \tilde{L}_{t,k;s} - \hat{\lambda}_{t,k}(\theta) \geq u \right)$ has the same upper bound as $\mathbb{P}\left( \left( \hat{\lambda}_{t,k} - \tilde{U}_{t,k;s}\right) \geq u \right) $.
Therefore, we have proven the desired upper bound. 
\end{proof}

The proof of Lemma \ref{lem:UCBPi1} uses the argument in the step~3 of the proof of Theorem~1 in \citet{bubeck2013prior}.
\begin{lem}
\label{lem:UCBPi1}
For any $t \in [T]$,  when $\delta = 2\sqrt{\frac{K}{S}}$, $u \geq \delta$, and $c = 1 -\frac{1}{\sqrt{3}}$, then we have 
\begin{align*}
\sum_{s=1}^{S}\mathbb{P}\left. \left( \left(\hat{U}_{t,k(t;s);s} - \hat{\lambda}_{t,k(t;s)}(\theta)\right) \geq u \right| \theta\right) 
&\leq K \left(\frac{3\log\left( \frac{S u^2}{K} \right)}{u^2} + \frac{\exp(-12c^2\log2)}{1-\exp(-2c^2u^2)}\mathbf{1}\left( u \leq 1/c \right)\right), 
\\
\sum_{s=1}^{S}\mathbb{P}\left. \left( \left(\hat{\lambda}_{t,k(t;s)}(\theta) - \hat{L}_{t,k(t;s);s}\right) \geq u \right| \theta\right) 
&\leq K \left(\frac{3\log\left( \frac{S u^2}{K} \right)}{u^2} + \frac{\exp(-12c^2\log2)}{1-\exp(-2c^2u^2)}\mathbf{1}\left( u \leq 1/c \right) \right).
\end{align*}
\end{lem}
\begin{proof}
During the course of this proof, we evaluate the left-hand sides of the above inequalities conditionally on $\theta$. 
We will exclude any dependency on $\theta$ unless it is crucial for clarity.
First, we evaluate the following indicator function:
\begin{align}
  \sum_{s=1}^{S}\mathbf{1}\left(\hat{U}_{t,k(t;s);s} - \hat{\lambda}_{t,k(t;s)} \geq u  \right) 
  & = \sum_{s=1}^{S}\sum_{k=1}^{K} \mathbf{1}\left(k(t;s)=k, \hat{U}_{t,k;s} - \hat{\lambda}_{t,k} \geq u  \right) \nonumber 
  \\
  &= \sum_{s=1}^{S}\sum_{k=1}^{K} \sum_{n=1}^{S}\mathbf{1}\left(k(t;s)=k, N_{t;k}(s-1)= n, \hat{U}_{t,k}(n) - \hat{\lambda}_{t,k} \geq u  \right). \nonumber
\end{align} 
For any integers $s, s'$ satisfying $1\leq s'<s \leq S$ and fixed number $n$, 
the events $\{ k(t;s)=k, N_{t;k}(s-1)= n \}$ and $\{ K(t;s')=k, N_{t;k}(s'-1)= n \}$ are mutually exclusive. This implies that 
\begin{align}
\label{eq:Union}
\sum_{s=1}^{S}\mathbf{1}\left(k(t;s)=k, N_{t;k}(s-1)=n\right) = \mathbf{1}\left(\bigcup_{s=1}^{S}\{k(t;s)=k, N_{t;k}(s-1)=n\}\right).
\end{align}
Using this equality, we have   
\begin{align*} 
  \sum_{s=1}^{S} \mathbf{1}\left( \hat{U}_{t,k(t;s);s} - \hat{\lambda}_{t,k(t;s)} \geq u  \right) =&\sum_{s=1}^{S}\sum_{k=1}^{K} \sum_{n=1}^{S}\mathbf{1}\left(k(t;s)=k, N_{t;k}(s-1)= n, \hat{U}_{t,k}(n) - \hat{\lambda}_{t,k}(\theta) \geq u  \right) 
  \nonumber \\
  =&\sum_{k=1}^{K} \sum_{n=1}^{S}\mathbf{1}\left( \bigcup_{s=1}^{S}\{k(t;s)=k, N_{t;k}(s-1)= n\},\hat D_{t,k}(n) + \gamma_{n}- \hat{\lambda}_{t,k}(\theta) \geq u \right) \text{(by \eqref{eq:Union})}
  \nonumber \\
  \leq& \sum_{k=1}^{K} \sum_{n=1}^{S}\mathbf{1}\left(\hat D_{t,k}(n) + \gamma_{n}- \hat{\lambda}_{t,k}(\theta) \geq u \right).
\end{align*}
We take the expectation value of the both side in the above inequality 
and thus obtain the following inequality: 
\begin{align}
  \label{eq:UCBPi1union}
  \sum_{n=1}^{S} \mathbb{P}\left(\hat{U}_{t,k(t;s);s} - \hat{\lambda}_{t,k(t;s)} \geq u  \right) \leq \sum_{k=1}^{K}\sum_{n=1}^{S}\mathbb{P}\left(\hat D_{t,k}(n) + \gamma_{n} - \hat{\lambda}_{t,k} \geq u \right).
\end{align}
We then divide the summation in the right-hand side into two parts as 
\begin{align}
  \label{eq:UCBPi1_1}
  \sum_{n=1}^{S}\mathbb{P}\left(\hat D_{t,k}(n) + \gamma_n - \hat{\lambda}_{t,k} \geq u \right)  \leq 
  &  \sum_{n=1}^{n(u)}\mathbb{P}\left(\hat D_{t,k}(n) + \gamma_{n} - \hat{\lambda}_{t,k} \geq u \right) + \sum_{n=n(u)+1} ^{\infty} \mathbb{P}\left(\hat D_{t,k}(n) + \gamma_{n} - \hat{\lambda}_{t,k}\geq u \right),
\end{align}
where $n(u) = \lfloor 3\log\left( \left(Su^2\right)/{K}\right)/u^2 \rfloor$. 
The first term is trivially bounded as 
\begin{align}
\label{eq:UCBPi1st}
\sum_{n=1}^{n(u)} \mathbb{P}\left(\hat D_{t,k}(n) + \gamma_n - \hat{\lambda}_{t,k} \geq u \right) \leq n(u) \leq h(u),
\end{align}
where $h(u) = 3\log\left( \left(Su^2\right)/{K}\right)/u^2$. 
To evaluate the second term of \eqref{eq:UCBPi1_1}, we provide an inequality of $\gamma_n$ for $n \geq n(u) + 1$.  
It is non-increasing since $\gamma_n$ is decreasing when $n \leq \lfloor S/K \rfloor $ and constantly 0 when $n \geq \lceil S/K \rceil $. %since $\log_{+}\left(\frac{S}{Kn}\right)=0$. 
When $n(u)+1 \leq n \leq \lfloor S/K\rfloor $, then we have $\log_{+}\left(\frac{S}{Kn}\right)=\log\left(\frac{S}{Kn}\right) $ and 
\begin{align*}
\gamma_{n} \leq \gamma_{n(u) + 1}  &\leq   \sqrt{\frac{\log\left(\frac{S}{Kh(u)}\right)}{h(u)}} &&  \left(h(u) \leq n(u) + 1\right)
\\ 
&= u\sqrt{\frac{\log\left(\frac{Su^2}{Ku^2h(u)}\right)}{u^2h(u)}} 
\\
&= u \sqrt{\frac{1}{3} - \frac{\log (u^2h(u))}{u^2h(u)}} && \left(u^2 h(u) = 3\log\left( \frac{Su^2}{K}\right) \right)
\\
&\leq u \sqrt{\frac{1}{3}}. 
\end{align*}
When $n \geq \lceil S/K\rceil $, $\gamma_{n}=0 \leq u \sqrt{\frac{1}{3}}$.  Consequently, for any $n \geq n(u) +1$,  $\gamma_{n} \leq u\sqrt{\frac{1}{3}} $. 
Using this inequality for $\gamma_n$ to the second term in \eqref{eq:UCBPi1_1}, we have 
  \begin{align*}
  \sum_{n=n(u)+1} ^{\infty} \mathbb{P}\left(\hat D_{t,k}(n) + \gamma_n - \hat{\lambda}_{t,k} \geq u \right) \leq 
  \sum_{n=n(u)+1} ^{\infty} \mathbb{P}\left(\hat D_{t,k}(n) - \hat{\lambda}_{t,k} \geq cu \right).
\end{align*}
We then apply Hoeffding's inequality (Lemma \ref{lem:hoeffdingineq})  to the right-hand side to have
\begin{align}
  \label{eq:UCBPi3}
  \sum_{n = n(u)+1} ^{\infty} \mathbb{P}\left(\hat D_{t,k}(n) - \hat{\lambda}_{t,k;s} \geq cu \right) \leq& \sum_{n = n(u)+1}^{\infty} \exp\left(- 2n c^2 u^2\right) \mathbf{1}\left( u \leq \frac{1}{c} \right)  &&\text{(By Lemma \ref{lem:hoeffdingineq})}\nonumber \\
  \leq & \frac{\exp\left(-2 \left(n(u) + 1\right) c^2u^2\right)}{1 - \exp\left(-2 c^2u^2\right)}\mathbf{1}\left( u \leq \frac{1}{c} \right) \nonumber \\
  \leq & \frac{\exp\left(-6\log \left(\frac{Su^2}{K}\right) c^2 \right)}{1 - \exp\left(-2 c^2u^2\right)} \mathbf{1}\left( u \leq \frac{1}{c} \right)  && ( h(u)\leq n(u) +1 ) \nonumber \\
  \leq & \frac{\exp\left(-12\log \left(2\right) c^2 \right)}{1 - \exp\left(-2 c^2u^2\right)} \mathbf{1}\left( u \leq \frac{1}{c} \right). && (u \geq \delta = 2\sqrt{\frac{K}{S}})
\end{align} 

Combining this with \eqref{eq:UCBPi1st}, we obtain 
\begin{align*}
  \sum_{s=1}^{S}\mathbb{P}\left( \hat{U}_{t,k(t;s);s} - \hat{\lambda}_{t,k(t;s)} \geq u \right) \leq K\left(\frac{3\log\left( \frac{S u^2}{K} \right)}{u^2} + \frac{\exp(-12c^2\log2)}{1-\exp(-2c^2u^2)}\mathbf{1}\left( u \leq 1/c \right) \right). 
\end{align*}
Next, we evaluate $\sum_{s=1}^{S}\mathbb{P}\left(\hat{\lambda}_{t,k(t;s)} - \hat{L}_{t,k(t;s)} \geq u \right)$ by following the same procedure we argued so far. 
In fact, we have the corresponding part to \eqref{eq:UCBPi1union} 
\begin{align}
  \sum_{n=1}^{S} \mathbb{P}\left(\hat{\lambda}_{t,k(t;s)} - \hat{L}_{t,K(t;s)} \geq u  \right) \leq \sum_{k=1}^{K}\sum_{n=1}^{S}\mathbb{P}\left( \hat{\lambda}_{t,k}  - \hat{D}_{t,k}(n) + \gamma_{n} \geq u \right), 
\end{align}
because the argument during the evaluation of \eqref{eq:UCBPi1union} is independent of the details of the event. %that the probability function measures.
Subsequently, we have
\begin{align*}
\frac{1}{K}\sum_{n=1}^{S}\mathbb{P}\left( \hat{\lambda}_{t,k}  - \hat{D}_{t,k}(n) + \gamma_{n} \geq u \right)  & \leq h(u) + \frac{1}{K}\sum_{k=1}^{K}\sum_{n=n(u) +1}^{\infty}\mathbb{P}\left( \hat{\lambda}_{t,k}  - \hat{D}_{t,k}(n) \geq cu \right) \\
&\leq h(u) +\frac{\exp(-12c^2\log2)}{1-\exp(-2c^2u^2)}\mathbf{1}\left( u \leq 1/c \right). && \text{(by the same argument in \eqref{eq:UCBPi3})}
\end{align*}
We finally have proved lemma \ref{lem:UCBPi1}. 
\end{proof}

\subsection{Revenue Analysis after Regret Decomposition}
\label{sec:revenueanalysis}
The lemmas in this section are main building blocks in the proof of the Theorem 1. 
These lemmas ensure upper bounds of the terms emerging after the regret decomposition with upper confidence bounds in the proof of Lemmas \ref{lem:revenuediff}. 
These lemmas are also used in the analysis of the lost sales part in Appendix \ref{sec:lostsalesanalysis}. 
The proof of Lemma \ref{lem:evarev1} and \ref{lem:Bubeckstep3} are based on the results in the step~2 and step~3 in the proof of Theorem~1 in \citet{bubeck2013prior}, respectively.

\label{sec:revanalysis}
\begin{lem}
    \label{lem:evarev1}
For any $t \in [T]$,  we have  
\begin{align*}
\sum_{k=1}^{K}\sum_{s=1}^{S}\mathbb{E}^{\pi}\left[\left(\hat{\lambda}_{t,k}(\theta) - U_{t,k;s} \right)^{+} \bigg|\theta \right] &\leq 6\sqrt{SK},\\ 
\sum_{k=1}^{K}\sum_{s=1}^{S}\mathbb{E}^{\pi}\left[\left(L_{t,k;s} - \hat{\lambda}_{t,k}(\theta) \right)^{+} \bigg|\theta \right] &\leq 6\sqrt{SK}.
\end{align*}
\end{lem}
\begin{proof}
Let $\delta_0 = 2\sqrt{\frac{1}{SK}}$, and we have
\begin{align}
\label{eq:r*-u*}
\mathbb{E}\left[\left(\hat{\lambda}_{t,k}(\theta) - U_{t,k;s}\right)^{+}\bigg| \theta \right] 
&= \left(
\int_{0}^{1} \mathbb{P}\left( \left. \left(\hat{\lambda}_{t,k}(\theta) - U_{t,k;s}\right)^{+} \geq u  \right| \theta\right)du \right),\quad (\text{by partial integral}) 
\nonumber \\
&= \left(\int_{0}^{\delta_0} \mathbb{P}\left(\left. \left(\hat{\lambda}_{t,k}(\theta) - U_{t,k;s}\right)^{+} \geq u \right| \theta \right)du + 
\int_{\delta_0}^{1} \mathbb{P} \left(\left. \left(\hat{\lambda}_{t,k}(\theta) - U_{t,k;s}\right)^{+} \geq u \right| \theta \right) du
\right) 
\nonumber \\
&\leq \left( \delta_0 + 
\int_{\delta_0}^{1} \mathbb{P}\left(\left. \left(\hat{\lambda}_{t,k}(\theta) - U_{t,k;s}\right)^{+} \geq u \right| \theta  \right)du\right).
\end{align}
To evaluate the rightest-hand side,  we use the relation $\left(\hat{\lambda}_{t,k}(\theta) - U_{t,k;s}\right)^{+} = \left(\hat{\lambda}_{t,k}(\theta) - \tilde{U}_{t,k;s}\right)^{+}$, 
which holds because when $\tilde{U}_{t,k;s} \geq 1$, $\left(\hat{\lambda}_{t,k}(\theta) - U_{t,k;s}\right)^{+} = \left(\hat{\lambda}_{t,k}(\theta) - \tilde{U}_{t,k;s}\right)^{+} =0$ or otherwise, $U_{t,k;s} = \tilde{U}_{t,k;s}$. 
We then have $ \mathbb{P}\left( \left. \left(\hat{\lambda}_{t,k}(\theta) - \tilde{U}_{t,k;s}  \right)^{+} \geq u \right| \theta \right) =
 \mathbb{P}\left(\left. \ \hat{\lambda}_{t,k}(\theta) - U_{t,k;s} \geq u \right| \theta \right)$ for any positive real number $u$. 
This enables us to bound the integral of \eqref{eq:r*-u*} by Lemma \ref{lem:eva*UCB*} as
\begin{align*}
\int_{\delta_0}^{1} \mathbb{P}\left( \left. \left(\lambda_{t,k}(\theta) - \tilde{U}_{t,k;s}\right)^{+} \geq u  \right| \theta \right) du 
\leq \int_{\delta_0}^{1} \frac{4}{SKu^2}\log\left(\sqrt{SK}u\right)du + \int_{\delta_0}^{1}\frac{1}{KSu^2-1}du.
\end{align*}
The integrals in the right-hand side are evaluated as 
\begin{align*}
\int_{\delta_{0}}^{1}\frac{4\log\left(\sqrt{SK}u\right)}{SKu^2}du &= 
\left[\frac{-4\log\left(\sqrt{SK}u\right)}{SKu}\right]^{u=1}_{u=\delta_{0}} - \left[\frac{4}{SKu}\right]^{u=1}_{u=\delta_{0}}\nonumber \\
&\leq \frac{1}{KS\delta_{0}}\log\left(\text{e}\sqrt{SK}\delta_{0}\right) = \frac{2(1+\log2)}{\sqrt{SK}}, \\
\int_{\delta_{0}}^{1}\frac{du}{SKu^2-1} &= 
\left[\frac{1}{2\sqrt{SK}}\log\left(\frac{\sqrt{SK}u-1}{\sqrt{SK}u+1}\right)\right]^{u=1}_{u=\delta_{0}} \leq \frac{\log 3 }{2\sqrt{SK}}.
\end{align*}
We then have proved that 
\begin{align*}
\sum_{k=1}^{K}\sum_{s=1}^{S}\mathbb{E}\left[\left(\hat{\lambda}_{t,k}(\theta) - U_{t,k;s} \right)^{+} \bigg|\theta \right] \leq \left(2+2(1+\log2)+\frac{\log3}{2}\right)\sqrt{KS}
\leq  6 \sqrt{KS}.
\end{align*}
Following the same argument so far, we also have 
\begin{align*}
  \sum_{s=1}^{S}\sum_{k=1}^{K}
      \mathbb{E}\left[\left(L_{t,k;s} - \hat{\lambda}_{t,k}(\theta) \right)^{+}\bigg| \theta \right] & \leq 6\sqrt{KS},
\end{align*}
and thus Lemma~\ref{lem:evarev1}.
\end{proof}

\begin{lem}
\label{lem:Bubeckstep3}
For any $t \in [T]$, we have 
\begin{align*}
  \sum_{s=1}^{S}\mathbb{E}^{\pi}\left[
  \left(\hat{U}_{t,k(t;s);s} - \hat{\lambda}_{t;k(t;s)}(\theta) \right)^{+} \bigg| \theta \right] &\leq 8\sqrt{SK}, \\
  \sum_{s=1}^{S}\mathbb{E}^{\pi}\left[
      \left(\hat{\lambda}_{t;k(t;s)}(\theta) - \hat{L}_{t,k(t;s);s} \right)^{+} \bigg| \theta \right]&\leq 8\sqrt{SK}.
\end{align*}
\end{lem}
\begin{proof} 
Following the similar procedure in the evaluation of \eqref{eq:r*-u*} of Lemma \ref{lem:evarev1}, 
we have  
\begin{align*}
\label{eq:Bubeckstep3}
 \mathbb{E}^{\pi}\left[
  \left(\hat{U}_{t,K(t;s);s} - \hat{\lambda}_{t;k(t;s)}(\theta)\right)^+  \bigg| \theta \right] 
 & \leq \left(\delta + \int_{\delta}^{\infty} \mathbb{P}\left( \hat{U}_{t,k(t;s);s} - \hat{\lambda}_{t,k(t;s)}(\theta) \geq u \right)du \right), 
\end{align*}
where $\delta = 2\sqrt{\frac{K}{S}}$. Using Lemma \ref{lem:UCBPi1}, we have 
\begin{align*}
% \label{eq:UCBPi1}
\sum_{s=1}^{S}\mathbb{E}^{\pi}\left[
  \left(\hat{U}_{t,k(t;s);s} - \hat{\lambda}_{t;k(t;s)}(\theta)\right) \bigg| \theta \right]
 \leq S\delta+ K\left(\int_{\delta}^{\infty}\frac{3\log\left( \frac{S u^2}{K} \right)}{u^2}du + 
 \int_{\delta}^{\infty}\frac{\exp(-12c^2\log2)}{1-\exp(-2c^2u^2)}\mathbf{1}\left( u \leq 1/c \right)du\right).
\end{align*}
The first integral is evaluated as 
\begin{align*}
\int_{\delta_{0}}^{\infty}\frac{3\log\left(\frac{Su^2}{K}\right)}{u^2}du =- \left[\frac{6\log\left(u\sqrt{\frac{S}{K}}\right)}{u}\right]^{u=\infty}_{u=\delta} + \int_{\delta}^{\infty} \frac{6}{u^2}du = 
  3\left(1+\log2\right)\sqrt{\frac{S}{K}} \leq 5.1\sqrt{\frac{S}{K}}, 
\end{align*}
and the second integral is bounded as
\begin{align*}
\int_{\delta_{0}}^{\frac{1}{c}}\frac{\exp\left(-12c^2\log2\right)}{1-\exp\left(-2c^2u^2\right)} du
&\leq \int_{\delta_0}^{\frac{1}{2c}}   \frac{\exp\left(-12c^2\log2\right)}{1-\exp\left(-2c^2u^2\right)}du + \int_{\frac{1}{2c}}^{\frac{1}{c}}\frac{\exp\left(-12c^2\log2\right)}{1-\exp\left(-2c^2u^2\right)}du 
 \\
&\leq  \int_{\delta_{0}}^{\frac{1}{2c}}\frac{\exp\left(-12c^2\log2\right)}{2c^2u^2 - 2c^4u^4}du + \int_{\frac{1}{2c}}^{\frac{1}{c}} \frac{\exp\left(-12c^2\log2\right)}{\left(1-\exp(-\frac{1}{2})\right)}du &&(1-\e^{-x}\geq x-x^2/2,\quad x\geq0) 
\\
&\leq  \int_{\delta_{0}}^{\frac{1}{2c}}\frac{2\exp\left(-12c^2\log2\right)}{3c^2u^2}du +  \frac{\exp\left(-12c^2\log2\right)}{2c \left(1-\frac{1}{\sqrt{e}}\right)} && (-2c^2u^4 \geq -2c^2u^2(1/2c)^2, u \geq \frac{1}{2c}) 
\\
&\leq  \exp\left(-12c^2\log2\right) \left( \frac{2}{3c^2\delta_0} - \frac{4}{3c} \right) +  \frac{\exp\left(-12c^2\log2\right)}{2c \left(1-\frac{1}{\sqrt{\e}}\right)} 
 \\
&\leq  \exp\left(-12c^2\log2\right) \left( \left(1 + \frac{\sqrt{3}}{2}\right)\sqrt{\frac{S}{K}} + \frac{1}{2c}\left(\frac{\sqrt{\e}}{\sqrt{\e}-1} - \frac{8}{3} \right) \right) 
 \\
&\leq \exp\left(-12c^2\log2\right) \left(1 + \frac{\sqrt{3}}{2}\right) \sqrt{\frac{S}{K}} &&\left( \frac{\sqrt{\e}}{\sqrt{\e}-1} - \frac{8}{3} \leq 0\right) 
\\ 
&\leq  \exp\left( - \frac{8}{4}\log 2\right) \left(1 + \frac{\sqrt{3}}{2}\right) \sqrt{\frac{S}{K}} \leq \frac{1}{2}\sqrt{\frac{S}{K}}.
\end{align*}
Consequently, we have 
\begin{align*}
\sum_{s=1}^{S}\sum_{k=1}^{K}\mathbb{E}^{\pi}\left[
\left(\hat{U}_{t,k(t;s);s} - \hat{\lambda}_{t;k(t;s)}(\theta) \right)^{+} \bigg| \theta \right] \leq \left(2 + 5.1\ + \frac{1}{2}\right)\sqrt{SK}\leq 8\sqrt{SK}.
\end{align*}
Using the same argument so far, we also obtain 
\begin{align*}
  \sum_{s=1}^{S}\sum_{k=1}^{K}\mathbb{E}^{\pi}\left[
    \left(\hat{\lambda}_{t;k(t;s)}(\theta) - \hat{L}_{t,k(t;s);s} \right)^{+}  \bigg| \theta \right] \leq 8\sqrt{SK}.
\end{align*}
and have proved Lemma \ref{lem:Bubeckstep3}.
\end{proof}

To prove the following lemma, we follows the argument presented in Proposition 2 of \citet{russo2014learning}.
\begin{lem}
  \label{lem:UCBPi2}
  For any $t \in [T]$, then we have 
\begin{align}
  \sum_{s=1}^{S}\sum_{k=1}^{K}\mathbb{E}^{\pi}\left[\min\left(\sqrt{\frac{2\log\left(K\right)}{N_{k;t}(s-1)}},1\right)\bigg| \theta\right]
  \leq \left( 1 + 2\sqrt{2} \right)\sqrt{KS\log\left(K\right)}
\end{align}
\end{lem}
\begin{proof}
We have
\begin{align}
\sum_{s=1}^{S}\sum_{k=1}^{K}\mathbb{E}^{\pi}\left[\min\left(\sqrt{\frac{2\log\left(K\right)}{N_{k;t}(s-1)}},1\right)\bigg| \theta \right]  = \mathbb{E}^{\pi}\left[\sum_{k=1}^{K}\sum_{s \in \sigma_{k;t}(S)}\min\left(\sqrt{\frac{2\log\left(K \right)}{N_{k;t}(s-1)}},1\right) \bigg| \theta\right],
\end{align}
where $\sigma_{k;t}(S)$ denotes the set of episodes in which $k$-th price is offered and is defined as $\sigma_{k;t}(S) = \{s\in [S] | P(t;s)=p_{k}\}$. 

Let $s_{1}$ be the first episode where the $k$-th price is offered, which implies that $N_{k;t}(s_1-1)=0$. 
We then have
\begin{align*}
  \mathbb{E}^{\pi}\left[\sum_{k=1}^{K}\sum_{s \in \sigma_{k;t}(S)}\min\left(\sqrt{\frac{2\log\left(K \right)}{N_{k;t}(s-1)}},1\right)\bigg| \theta \right] &= K + \mathbb{E}^{\pi}\left[\sum_{k=1}^{K}\sum_{s \in \sigma_{k;t}(S)\backslash s_1}\min\left(\sqrt{\frac{2\log\left(K \right)}{N_{k;t}(s-1)}},1\right)\bigg| \theta \right] \nonumber \\
    &\leq K + \mathbb{E}^{\pi}\left[\sum_{k=1}^{K}\sum_{s \in \sigma_{k;t}(S)\backslash s_1}\sqrt{\frac{2\log\left(K \right)}{N_{k;t}(s-1)}}\bigg| \theta \right] \quad\left( \min(x,y) \leq x  \right)
    \nonumber \\
    &\leq K + \mathbb{E}^{\pi}\left[\sum_{k=1}^{K}\sum_{s \in \sigma_{k;t}(S)}\sqrt{\frac{2\log\left(K \right)}{1 + N_{k;t}(s-1)}} \bigg| \theta\right] 
    \nonumber \\
    &= K+\mathbb{E}^{\pi}\left[\sum_{k=1}^{K}\sum_{j=0}^{N_{t;k}(S)-1}\sqrt{\frac{2\log\left(K \right)}{j + 1}} \bigg| \theta \right] \quad \left(|\sigma_{k;t}(S)|= N_{t,k}(S)\right) 
    \nonumber \\
    &\underset{(\dagger)}{\leq} K+ 2\mathbb{E}^{\pi}\left[\sum_{k=1}^{K}\sqrt{2\log\left(K \right)N_{t,k}(S)}\bigg| \theta\right] 
    \nonumber \\
    &\leq K + 2\sqrt{2KS\log\left(K\right)}\quad \left( \text{by the Cauchy-Schwarz inequality and } \sum_{k=1}^{K}N_{t,k}(S)\leq S\right) 
    \nonumber \\
    &\leq \left( 1 + 2\sqrt{2} \right) \max\left\{ \sqrt{KS\log\left(K\right)}, K\sqrt{\log \left(K\right)} \right\}
    \\
    & \leq  \left( 1 + 2\sqrt{2} \right)\sqrt{KS\log\left(K\right)}.\quad  (K\leq S)
\end{align*}
When obtaining the inequality marked by $(\dagger)$,  we used
 the following relation:  
\begin{align}
\sum_{j=0}^{N_{k;t}(S)-1}\frac{1}{\sqrt{j+1}} \leq \int_{0}^{N_{k;t}(S)}\frac{dx}{\sqrt{x}}= 2\sqrt{N_{t;k}(S)}.
\end{align}
Therefore, we have completed the proof of Lemma \ref{lem:UCBPi2}.
\end{proof}

%In the proofs of the following lemmas, we will use this $\hat{U}_{t,k;s}$ rather than $\tilde{U}_{t,k;s}$ defined in \eqref{def:UCBtilde}.
\begin{lem}
  \label{lem:UCBPi}
  For any $t \in [T]$, we have 
\begin{align*}
%\label{proof2}
\sum_{s=1}^{S}\mathbb{E}^{\pi}\left[ \left(U_{t,k(t;s);s} - \hat{\lambda}_{t,k(t;s)}(\theta)\right)^{+}  \bigg| \theta \right]
 &\leq 12\sqrt{SK\log(K)}, \\
 \sum_{s=1}^{S}\mathbb{E}^{\pi} \left[\left. \left( \hat{\lambda}_{t,k(t;s)}(\theta) - L_{t,k(t;s);s}\right)^{+} \right| \theta \right] &\leq 12 \sqrt{SK\log(K)}.
\end{align*}
\end{lem}
\begin{proof}
  
We have 
\begin{align*}
\label{eq:UCBPi1}
\mathbb{E}^{\pi}\left[ \left(U_{t,k(t;s);s} - \hat{\lambda}_{t, k(t;s)}(\theta)\right)^{+} \bigg| \theta \right]
\leq \underset{(A)}{\underline{\mathbb{E}^{\pi}\left[
    \left(\hat{U}_{t,k(t;s);s} - \hat{\lambda}_{t;k(t;s)}(\theta)\right)^{+} \bigg| \theta \right]}}
    + \underset{(B)}{\underline{ \mathbb{E}^{\pi}\left[
    \min\left\{\sqrt{\frac{2\log(K)}{N_{k(t;s),t}(s-1)}}, 1 \right\} \bigg| \theta \right]}},
\end{align*}
where we extract the $\log(K^2)$ factor from $\tilde{U}_{t,k;s}$ and can bound $\tilde{U}_{t,k;s}$ as 
\begin{align*}
\tilde{U}_{t, k(t;s);s} &= \min\left(\hat{D}_{t, k(t;s);s} + \sqrt{\frac{\log_{+}\left(\frac{S}{KN_{k(t;s);t}(s-1)}\right)+ \log_{+}\left(K^2\right)}{N_{k(t;s),t}(s-1)}},1 \right)\nonumber 
\\
&\leq \min\left(\hat{U}_{t,k(t;s);s}+ \sqrt{\frac{2\log\left(K\right)}{N_{k(t;s);t}(s-1)}}, 1\right)\quad &&( \sqrt{x}+\sqrt{y} \geq \sqrt{x+y}) \nonumber 
\\
&\leq \hat{U}_{k(t;s),t;s}+ \min\left(\sqrt{\frac{2\log\left(K\right)}{N_{k(t;s);t}(s-1)}},1\right). \quad &&( \min\left(x+y,z\right) \leq x + \min\left(y,z\right))
\end{align*}    
The under line part $(A)$ has an upper bound, which is given by Lemma \ref{lem:Bubeckstep3}, 
\begin{align*}
\sum_{s=1}^{S}\mathbb{E}^{\pi}\left[ 
\left(\hat{U}_{t,k(t;s);s}-\hat{\lambda}_{t,k(t;s)}(\theta) \right)^{+}  \bigg|\theta \right] \leq 8\sqrt{SK},
\end{align*}
and the underline part $(B)$ is also bounded above using Lemma \ref{lem:UCBPi2}: 
\begin{align*}
    \sum_{s=1}^{S}\mathbb{E}^{\pi}\left[\min\left(\sqrt{\frac{2\log\left(K\right)}{N_{K(t;s);t}(s-1)}},1\right) \bigg| \theta\right] 
    \leq \sum_{s=1}^{S}\sum_{k=1}^{K}\mathbb{E}^{\pi}\left[\min\left(\sqrt{\frac{2\log\left(K\right)}{N_{k;t}(s-1)}},1\right) \bigg| \theta\right] 
    \leq \left( 1 + 2\sqrt{2} \right)\sqrt{KS\log\left(K\right)}.
\end{align*}
Thus, we have
\begin{align*}
  \sum_{k=1}^{K}\sum_{s=1}^{S}\mathbb{E}^{\pi}\left[ \left(U_{t,K(t;s);s} - \hat{\lambda}_{t,K(t;s)}(\theta)\right)  \bigg| \theta \right]  \leq \left( 9 + 2\sqrt{2} \right)\sqrt{KS\log{K}}\leq 12 \sqrt{KS\log K}.
\end{align*}
The same argument is applicable to the evaluation of $\sum_{s=1}^{S}\mathbb{E}^{\pi} \left[\left( \hat{\lambda}_{t,k(t;s)}(\theta) - L_{t,k(t;s);s}\right)^{+} \right]$.
Therefore, we have proved Lemma~\ref{lem:UCBPi}.
\end{proof}

\begin{lem}
  \label{lem:lostUL}
For any $t \in [T]$, we then have 
\begin{align}
\sum_{s=1}^{S}\mathbb{E}^{\pi} \left[\left(U_{t,k(t;s);s} - L_{t,k(t;s);s}\right) \bigg| \theta  \right] \leq 24\sqrt{SK\log(K)}.
\end{align}
\end{lem}
\begin{proof}
The left-hand side is evaluated as
\begin{align*}
  \sum_{s=1}^{S}\mathbb{E}^{\pi} \left[\left(U_{t,k(t;s);s} - L_{t,k(t;s);s}\right) \bigg| \theta  \right] 
  &=\sum_{s=1}^{S}\mathbb{E}^{\pi} \left[\left(U_{t,k(t;s);s} - \hat{\lambda}_{t,k(t;s)}(\theta) + \hat{\lambda}_{t,k(t;s)}(\theta) - L_{t,K(t;s);s}\right) \bigg| \theta \right] \nonumber \\
  &\leq \sum_{s=1}^{S} \left(\underset{(a)}{\underline{\mathbb{E}^{\pi}\left[\left(U_{t,k(t;s);s} - \hat{\lambda}_{t,k(t;s)}(\theta)\right)^{+} \bigg| \theta \right]}} 
  + \underset{(b)}{\underline{\mathbb{E}^{\pi} \left[\left(\hat{\lambda}_{t,k(t;s)}(\theta) - L_{t,k(t;s);s}\right)^{+} \bigg| \theta \right]}}\right).
\end{align*}
Both the underlined part $(a)$ and $(b)$ are bounded above by lemma \ref{lem:UCBPi}. 
Therefore, we have proven Lemma \ref{lem:lostUL}.
\end{proof}

\subsection{Lost Sales Analysis}
\label{sec:lostsalesanalysis}
\begin{lem}
  \label{lem:loststatic}
\begin{align*}
  \mathbb{E}^{\pi}\left[\sum_{s=1}^{S}\left(\sum_{t=1}^{T}\left(D_{t;s}- \sum_{k=1}^{K}\lambda_{t,k}(\theta)x_{t,k}(\theta_{s})\right)\right)^{+} \bigg| \theta \right] \leq S\bar{d}\sqrt{T}.
\end{align*}
\end{lem}
\begin{proof}
To evaluate the left-hand side, we have 
\begin{align*}
\mathbb{E}^{\pi}\left[\left(\sum_{t=1}^{T}\left(D_{t,s}-\sum_{k=1}^{K}\lambda_{t,k}(\theta)x_{t,k}(\theta_{s}) \right)\right)^{+}\bigg| \theta\right] 
&\leq \mathbb{E}^{\pi}\left[\left|\sum_{t=1}^{T}\left( D_{t,s} - \sum_{k=1}^{K}\lambda_{t,k}(\theta)x_{t,k}\left(\theta_{s}\right)\right)\right| \bigg| \theta \right] \nonumber \\
&\leq \left(\mathbb{E}^{\pi}\left[\left(\sum_{t=1}^{T}\left(D_{t,s}-\sum_{k=1}^{K}\lambda_{t,k}(\theta)x_{t,k}(\theta_{s})\right)\right)^2 \bigg| \theta \right] \right)^\frac{1}{2},
\end{align*}
where we use the relations $x^{+}\leq |x|$ and the Cauchy-Schwarz inequality $\mathbb{E}[|X|] \leq \left(\mathbb{E}[X^2]\right)^{\frac{1}{2}}$.
We also have 
\begin{align}
  \label{eq:loststaticint}
&\mathbb{E}^{\pi}\left[\left(\sum_{t=1}^{T}\left(D_{t,s}-\sum_{k=1}^{K}\lambda_{t,k}(\theta)x_{t,k}\left(\theta_{s}\right)\right)\right)^2 \bigg| \theta \right] \nonumber \\
  =& \mathbb{E}^{\pi}\left[\sum_{t=1}^{T}\left(D_{t,s}-\sum_{k=1}^{K}\lambda_{t,k}(\theta)x_{t,k}\left(\theta_{s}\right)\right)\sum_{u=1}^{T}\left(D_{u,s}-\sum_{k=1}^{K}\lambda_{u,k}(\theta)x_{u,k}\left(\theta_{s}\right)\right) \bigg| \theta \right] \nonumber \\
=& \mathbb{E}^{\pi}\left[\sum_{t=1}^{T}\left(D_{t,s}-\sum_{k=1}^{K}\lambda_{t,k}(\theta)x_{t,k}\left(\theta_{s}\right)\right)^2 \bigg| \theta\right] \nonumber \\
&+ 2\mathbb{E}^{\pi}\left[\sum_{1\leq t < u \leq T}^{T}\left(D_{t,s}-\sum_{k=1}^{K}\lambda_{t,k}(\theta)x_{t,k}\left(\theta_{s}\right)\right)\left(D_{u,s}-\sum_{k=1}^{K}\lambda_{u,k}(\theta)x_{u,k}\left(\theta_{s}\right)\right)\bigg| \theta  \right].
\end{align}
The second term vanishes because $D_{t;s}$ and $D_{u;s}$ are conditionally independent on $\theta_{s}$ and $H_{0}^{s}$. 
%In fact, \algone{} decides pricing over the episode at the beginning of the episode. 
To be specified, $D_{t;s}$ and $D_{u;s}$ have the conditional expectations $\sum_{k=1}^{K}\lambda_{t,k}(\theta)x_{t,k}\left(\theta_{s}\right)$
and $\sum_{k=1}^{K}\lambda_{u,k}(\theta)x_{u,k}\left(\theta_{s}\right)$, respectively. Then, we have  
\begin{align*}
  &\mathbb{E}^{\pi}\left[\left(D_{t,s}-\sum_{k=1}^{K}\lambda_{t,k}(\theta)x_{t,k}(\theta_{s})\right)\left(D_{u,s}-\sum_{k=1}^{K}\lambda_{u,k}(\theta)x_{u,k}\left(\theta_{s}\right)\right) \bigg| \theta  \right] \nonumber \\ 
  &= \mathbb{E}^{\pi}\left[\mathbb{E}^{\pi}\left[\left(D_{t,s}-\sum_{k=1}^{K}\lambda_{t,k}(\theta)x_{t,k}\left(\theta_{s}\right)\right) \bigg| H_{0}^{s}, \theta_{s}, \theta\right]
  \mathbb{E}^{\pi}\left[\left(D_{u,s}-\sum_{k=1}^{K}\lambda_{u,k}(\theta)x_{u,k}\left(\theta_{s}\right)\right)\bigg| H_{0}^{s}, \theta_{s},  \theta  \right] \bigg| \theta\right] \nonumber \\
  &=0.
\end{align*}
From the boundedness of demand,  we can bound the first term in \eqref{eq:loststaticint} as 
\begin{align*}
&\mathbb{E}^{\pi}\left[\sum_{t=1}^{T}\left(D_{t,s}-\sum_{k=1}^{K}\lambda_{t,k}(\theta)x_{t,k}\left(\theta_{s}\right)\right)^2 \bigg| \theta \right] \leq \bar{d}^2T. 
\end{align*}
Therefore, we obtain the desired upper bound.
\end{proof}

\begin{lem}
  \label{lem:lostsales2}
\begin{align*}
%\label{eq:lostsales2}
\mathbb{E}_{\theta}\left[\mathbb{E}^{\pi} \left[\sum_{s=1}^{S}\left(\sum_{t=1}^{T}\sum_{k=1}^{K}\lambda_{t,k}(\theta)x_{t,k}(\theta_{s}) - n_{0}\right)^{+} \bigg| \theta\right] \right] 
\leq 36\bar{d}T\sqrt{SK\log(K)}.
\end{align*}
\end{lem}
\begin{proof}
From the inventory constraint $\sum_{t=1}^{T}\sum_{k=1}^{K}\lambda_{t,k}(\theta_{s})x_{t,k}\left(\theta_{s}\right) \leq n_{0}$ in $ {\rm LP}(\lambda(\theta_{s}), 1, n_{0}$), 
we have
\begin{align*}
\mathbb{E}_{\theta}\left[\mathbb{E}^{\pi} \left[\left(\sum_{t=1}^{T}\sum_{k=1}^{K}\lambda_{t,k}(\theta)x_{t,k}(\theta_{s}) - n_0\right)^{+} \bigg| \theta\right] \right] \leq 
\mathbb{E}_{\theta}\left[\mathbb{E}^{\pi} \left[\sum_{t=1}^{T}\sum_{k=1}^{K}\left(\lambda_{t,k}(\theta) - \lambda_{t,k}(\theta_{s}) \right)^{+}x_{t,k}(\theta_{s}) \bigg| \theta\right] \right], 
\end{align*} 
where we use $(x+y)^{+} \leq x^{+} + y^{+}$. By using upper and lower confidence bounds defined in \eqref{def:UCB} and \eqref{def:LCB}, the right-hand side is decomposed as  
\begin{align*}
& \mathbb{E}_{\theta}\left[\mathbb{E}^{\pi} \left[\left(\lambda_{t,k}(\theta) - \lambda_{t,k}(\theta_{s}) \right)^{+}x_{t,k}\left(\theta_{s}\right) \bigg| \theta\right] \right]\nonumber \\
\leq&  \bar{d} \mathbb{E}_{\theta}\left[\mathbb{E}^{\pi} \left[\left(\hat{\lambda}_{t,k}(\theta) - U_{t,k;s} + U_{t,k;s} - L_{t,k;s} + L_{t,k;s}  - \hat{\lambda}_{t,k}(\theta_{s}) \right)^{+}x_{t,k}\left(\theta_{s}\right) \bigg| \theta\right] \right] \nonumber \\
\leq&  \bar{d}\mathbb{E}_{\theta}\left[\underset{(1)}{\underline{\mathbb{E}^{\pi} \left[\left(\hat{\lambda}_{t,k}(\theta)- U_{t,k;s}\right)^+ \bigg| \theta \right]}}+ 
\underset{(2)}{\underline{\mathbb{E}^{\pi} \left[\left(U_{t,k;s} - L_{t,k;s}\right)x_{t,k}\left(\theta_{s}\right)\bigg| \theta \right]}} +
\underset{(3)}{\underline{\mathbb{E}^{\pi} \left[\left(L_{t,k;s}  - \hat{\lambda}_{t,k}(\theta_{s})\right)^+ \bigg| \theta \right]} } \right].
\end{align*}
The underlined part (1) is bounded by $6/\sqrt{{SK}}$ from lemma \ref{lem:evarev1}.

Recall that $k(t;s)=k$ is chosen with probability $x_{t,k}(\theta_{s})$ given $\theta_{s}$, which implies that 
$x_{t,k}(\theta_s) = \mathbb{E}^{\pi}\left[\left. \mathbf{1}\left(k(t;s) =k\right) \right| \theta_s\right]$. 
In addition, as explained in Appendix \ref{sec:UCBdef}, $U_{t,k;s}$ and $L_{t,k;s}$ are constant conditionally on $H_{0}^{s}$. 
Then, the underlined part $(2)$ is bounded above as
\begin{align*}
\sum_{s=1}^{S}\sum_{k=1}^{K}\mathbb{E}^{\pi} \left[\left(U_{t,k;s} - L_{t,k;s}\right)x_{t,k}\left(\theta_{s}\right)\bigg| \theta \right] 
&= \sum_{s=1}^{S}\sum_{k=1}^{K}\mathbb{E}^{\pi}\left[ \mathbb{E}^{\pi} \left.\left[ \left. \mathbf{1}\left(k(t;s)=k\right)\left(U_{t,k;s}(H_{0}^{s}) - L_{t,k;s}(H_{0}^{s})\right)  \right| H_{0}^{s}, \theta_s,  \theta \right] \right| \theta\right] 
\\
&= \sum_{s=1}^{S}\mathbb{E}^{\pi}\left[ U_{t,k(t;s);s} - L_{t,k(t;s);s} \bigg|\theta \right] 
\\
&\leq 24\sqrt{SK\log(K)}. \quad \text{(by Lemma \ref{lem:lostUL})}
\end{align*}
Finally, to evaluate the underlined part (3), we follow the analysis in \citet{ferreira2018online}. 
They uses the argument in \citet{russo2014learning}, which states that 
both $\theta_{s}$ and $\theta$ given a history $H_{0}^{s}$ follow the same posterior distribution $f(\cdot| H_{0}^{s})$. 
We rearrange the underlined part (3) using the law of total expectation as  
\begin{align*}
  \mathbb{E}_{\theta}\left[\mathbb{E}^{\pi} \left[\left(L_{t,k;s}  - \hat{\lambda}_{t,k}(\theta_{s})\right)^+ \bigg| \theta \right] \right] %= &\mathbb{E}_{\theta}\left[\mathbb{E}^{\pi}_{H_{0}^{s}} \left[ \mathbb{E}^{\pi}_{\theta_{s}}\left[ \left(L_{t,k;s} - \hat{\lambda}_{t,k}(\theta_{s})\right)^+ \bigg| H_{0}^{s}, \theta \right] \bigg | \theta \right]  \right] \nonumber \\
  &= \mathbb{E}^{\pi}\left[ \left(L_{t,k;s} - \hat{\lambda}_{t,k}(\theta_{s})\right)^+ \right] \\
  %&= \mathbb{E}^{\pi}\left[ \mathbb{E}_{\theta}\left[\mathbb{E}^{\pi}\left[\left(L_{t,k;s}(H_{0}^{s}) - \hat{\lambda}_{t,k}(\theta_{s})\right)^+ \bigg| H_{0}^{s}, \theta \right] \bigg| H_{0}^{s}\right] \right] 
  %\\
  & = \mathbb{E}^{\pi}\left[ \mathbb{E}^{\pi}\left[\left(L_{t,k;s}(H_{0}^{s}) - \hat{\lambda}_{t,k}(\theta_{s})\right)^+ \bigg| H_{0}^{s}\right]  \right] 
  \\
  & = \mathbb{E}^{\pi}\left[ \mathbb{E}^{\pi}\left[\left(L_{t,k;s}(H_{0}^{s}) - \hat{\lambda}_{t,k}(\theta)\right)^+ \bigg| H_{0}^{s}\right]  \right]  \quad (\theta, \theta_s \sim f(\cdot|H^{s}_{0}))
  \\
  & = \mathbb{E}_{\theta}\left[\mathbb{E}^{\pi}\left[ \left(L_{t,k;s} - \hat{\lambda}_{t,k}(\theta)\right)^+ \bigg| \theta \right]\right] 
  \\
  &\leq 6\sqrt{\frac{1}{SK}}. \quad ( \text{by Lemma \ref{lem:evarev1}})
\end{align*}
Therefore, we have 
\begin{align*}
  \mathbb{E}^{\pi} \left[\sum_{s=1}^{S}\left(\sum_{t=1}^{T}\sum_{k=1}^{K}\lambda_{t,k}(\theta)x_{t,k}(\theta_{s})-n_{0}\right)^{+}\right] \leq& \bar{d}T \left( 6\sqrt{SK}+6\sqrt{SK} +24\sqrt{SK\log (K)} \right) \nonumber \\
\leq& 36 \bar{d}T  \sqrt{SK\log (K)}.
\end{align*}
\end{proof}

\subsection{Regret Analysis for the Decomposed Parts}
\label{sec:regretdecomposed}
\begin{lem}
  \label{lem:rev*vsoptLP}
  Let ${\rm OPT_{LP}}\left(T, \theta\right)$ denote the optimal value of ${\rm LP}(\theta, 1, n_{0})$ and 
  ${\rm Rev}^{*}\left(T, \theta\right)$ denote the optimal total expected revenue of the revenue management problem over a selling season. 
  Then, we have
    \begin{align*}
      {\rm Rev}^{*}\left(T, \theta\right) \leq {\rm OPT_{LP}}\left(T, \theta\right). 
    \end{align*}
 \end{lem}
 \begin{proof}
 
For the optimal policy $\pi^{*}$, which induces the optimal revenue ${\rm Rev}^{*}\left(T,\theta\right)$, 
 let $x^{*}_{t,k}$ denote $\mathbb{P}_{\pi^*}\left(P_{t}=p_{k}\right)$, 
 and then $1- \sum_{k=1}^{K}x^{*}_{t,k}$ represents $\mathbb{P}_{\pi^{*}}\left(P_{t}=p_{\infty}\right)$.
 This allows ${\rm Rev}^{*}\left(T,\lambda\right)$ to expand as 
 \begin{align*}
   {\rm Rev}^{*}\left(T, \theta \right)  = \sum_{t=1}^{T}\sum_{k=1}^{K}p_{k}x^{*}_{t,k}\mathbb{E}\left[\left. \tilde{D}_{t} \right|P_t = p_{k}, \theta\right].
 \end{align*}
 From the definition of $\tilde{D}_{t;s}$, the inequalities $\tilde{D}_{t}\leq D_{t}$ and $\sum_{t=1}^{T}\tilde{D}_t \leq n_0$ hold, which leads to 
 \begin{align*}
 \mathbb{E}^{\pi^{*}}\left[\sum_{t=1}^{T}\tilde{D}_t\bigg|\theta \right] 
 &%=\sum_{t=1}^{T}\sum_{k=1}^{K}x^{*}_{t,k}\mathbb{E}[\tilde{D}|P_t = p_{k}, \theta] 
 \leq n_0, \\
 \mathbb{E}\left[\tilde{D}_t\bigg| P_{t}=p_{k}, \theta \right] &\leq \mathbb{E}\left[D_t \bigg|P_t =p_{k}, \theta \right] = \lambda_{t,k}(\theta),\quad \forall t \in [T], \forall k \in [K].
 \end{align*}
 Using these inequalities, we can construct a feasible solution of ${\rm LP}(\lambda(\theta), 1,n_{0})$ as
 \begin{align*}
 x_{t,k}= \frac{x^{*}_{t,k}\mathbb{E}\left[\tilde{D}_t\bigg| P_{t}=p_{k}, \theta \right] }{\lambda_{t,k}(\theta) } \leq x^{*}_{t,k}, 
 \end{align*}
 where $\lambda_{t,k}(\theta) >0$ for $t\in [T]$ and $k \in [K]$ because the mean demand is positive except when the shut off price is offered. 
 This $\{x_{t,k}\}_{t\in[T], k\in[K]}$ satisfies all the constrains of ${\rm LP}(\lambda(\theta), 1,n_{0})$. In fact, we have 
 \begin{align*}
 x_{t,k} &\geq 0, \quad \forall k\in [K], t\in [T], \nonumber \\  
 \sum_{k=1}^{K} x_{t,k} &\leq \sum_{k=1}^{K} x^{*}_{t,k} \leq 1, \quad \forall t\in [T], \nonumber \\  
 \sum_{t=1}^{T}\sum_{k=1}^{K}x_{t,k} \lambda_{t,k}(\theta) &= \sum_{t=1}^{T}\sum_{k=1}^{K}x^{*}_{t,k}\mathbb{E}[\tilde{D}_t|P_t = p_{k}, \theta] = \mathbb{E}^{\pi}\left[\sum_{t=1}^{T}\tilde{D}_t\bigg|\theta \right] \leq n_0.
 \end{align*}
 Thus, there exists a feasible solution satisfying 
 \begin{align*} 
 \sum_{t=1}^{T}\sum_{k=1}^{K} p_{k}\lambda_{t,k}(\theta)x_{t,k} =  {\rm Rev}^{*}(T, \theta),
 \end{align*}
 which ensures that  
 \begin{align*}
   {\rm Rev}^{*}(T, \theta) \leq {\rm OPT_{LP}}(T, \theta).
 \end{align*}
 Therefore,  the proof has been completed.
 \end{proof}

\begin{lem}
  \label{lem:revenuediff}
%We then have
\begin{align}
  \mathbb{E}_{\theta} \left[S\left(\mathbb{E}^{\pi^{*}(\theta)}\left[\left. \sum_{t=1}^{T}\tilde{R}_{t} \right|\theta \right]\right) - 
  \mathbb{E}^{\pi}\left[\left. \sum_{s=1}^{S}\sum_{t=1}^{T}R_{t,s} \right| \theta \right] \right]\leq 18 p_{K}\bar{d}T\sqrt{KS\log(K)}.
\end{align}
\end{lem}
\begin{proof}

From Lemma \ref{lem:rev*vsoptLP}, we have
\begin{align*}
%\label{eq:toeva0}
&S\left(\mathbb{E}^{\pi^{*}}\left[\sum_{t=1}^{T}\tilde{R}_{t} \bigg|\theta \right]\right) - 
\mathbb{E}^{\pi}\left[\sum_{s=1}^{S}\sum_{t=1}^{T}R_{t;s} \bigg| \theta  \right] \\
&\leq  \sum_{s=1}^{S}\sum_{t=1}^{T}\left(\sum_{k=1}^{K} \left(x_{t,k}(\theta)\lambda_{t,k}(\theta)p_{k} \right) - \mathbb{E}^{\pi}\left[P_{t;s}D_{t;s} \bigg|\theta\right] \right) \\
&=\sum_{s=1}^{S}\sum_{t=1}^{T}\mathbb{E}^{\pi}\left[ \left. \left(\underset{(1)}{\underline{\sum_{k=1}^{K} \left(x_{t,k}(\theta)\lambda_{t,k}(\theta)p_{k}\right) -  \mathcal{U}^{*}_{t;s}}}
+ \underset{(2)}{\underline{\mathcal{U}^{\pi}_{t;s} - \sum_{k=1}^{K}p_{k}\lambda_{t,k}(\theta)x_{t,k}(\theta_s)} }
+ \underset{(3)}{\underline{\mathcal{U}^{*}_{t;s} - \mathcal{U}^{\pi}_{t;s}}} \right) \right| \theta\right]
\end{align*}
where we define $\mathcal{U}^{\pi}_{t;s}$ and $\mathcal{U}^{*}_{t;s}$ using the upper confidence bound \eqref{def:UCB} as 
\begin{align*}
\mathcal{U}^{\pi}_{t;s} 
= \bar{d}\sum_{k=1}^{K}p_{k}U_{t,k;s}x_{t,k}(\theta_s), \quad 
\mathcal{U}^{*}_{t;s} 
= \bar{d}\sum_{k=1}^{k} x_{t,k}(\theta) p_{k} U_{t,k;s}.
\end{align*}
The underlined part $(1)$ is bounded above by Lemma \ref{lem:evarev1} as 
\begin{align*}
\sum_{s=1}^{S}\mathbb{E}^{\pi}\left[  \left. \left(\sum_{k=1}^{K} \left(x_{t,k}(\theta)\lambda_{t,k}(\theta)p_{k}\right) -  \mathcal{U}^{*}_{t;s}\right) \right| \theta \right]&= \sum_{s=1}^{S}\sum_{k=1}^{K} \bar{d} \left(\mathbb{E}^{\pi}\left[p_kx_{t,k}(\theta) \left(\hat{\lambda}_{t,k}(\theta) - U_{t,k;s}\right)  \bigg|\theta \right]\right)
  \\ 
  &\leq \sum_{s=1}^{S}\sum_{k=1}^{K} \bar{d} \left(\mathbb{E}^{\pi}\left[p_kx_{t,k}(\theta) \left(\hat{\lambda}_{t,k}(\theta) - U_{t,k;s}\right)^{+} \bigg|\theta \right]\right) \quad  (x \leq x^{+})
 \\ 
  &\leq \sum_{s=1}^{S}\sum_{k=1}^{K} \bar{d} \left(\mathbb{E}^{\pi}\left[p_K \left(\hat{\lambda}_{t,k}(\theta) - U_{t,k;s}\right)^{+}  \bigg|\theta \right]\right) \quad \left(0\leq p_k x_{t,k}(\theta)\leq p_{K},\ \forall k \in [K]\right)
  \\ 
&\leq 6p_{K}\bar{d}\sqrt{SK}. \quad  (\text{by Lemma \ref{lem:evarev1}})
\end{align*}

Recall that $U_{t,k;s}(H_{0}^{s})$ is constant and that the $k$-th price is chosen independently with probability $x_{t,k}(\theta_{s})$ given $\theta_{s}$, which implies that 
$x_{t,k}(\theta_s) = \mathbb{E}^{\pi}\left[\left. \mathbf{1}\left(k(t;s) =k\right) \right| \theta_s\right]$. 
We can bound the underlined part (2) above as  
\begin{align*}
\sum_{s=1}^{S}\mathbb{E}^{\pi}\left[\left. \mathcal{U}^{\pi}_{t;s} - \sum_{k=1}^{K}p_{k}\lambda_{t,k}(\theta)x_{t,k}(\theta_s) \right| \theta\right]
&= \sum_{s=1}^{S}\sum_{k=1}^{K}\left(\bar{d}\mathbb{E}^{\pi}\left[\left(U_{t,k;s} - \hat{\lambda}_{t,k}(\theta)\right)p_k x_{t,k}(\theta_s) \bigg|\theta \right]\right) 
\\
& \leq \sum_{s=1}^{S}\sum_{k=1}^{K}\left(\bar{d}p_{K}\mathbb{E}^{\pi}\left[\left(U_{t,k;s} - \hat{\lambda}_{t,k}(\theta)\right)^{+} x_{t,k}(\theta_s) \bigg|\theta \right]\right) 
\\
& = \sum_{s=1}^{S}\sum_{k=1}^{K}\left(\bar{d}p_{K}\mathbb{E}^{\pi}\left[\mathbb{E}^{\pi}\left[\left. \left(U_{t,k;s}(H_{0}^{s}) - \hat{\lambda}_{t,k}(\theta) \right)^{+} \mathbf{1}\left(k_{t;s} = k\right) \bigg| H_{0}^{s}, \theta_s, \theta \right]  \right| \theta \right]\right) 
\\
& = \sum_{s=1}^{S}\left(p_{K}\bar{d}\mathbb{E}^{\pi}\left[\mathbb{E}^{\pi}\left[\left.\left(U_{t,k(t;s);s}(H_{0}^{s}) - \hat{\lambda}_{t,k(t;s)}(\theta) \right)^{+}\bigg| H_{0}^{s},\theta_s,  \theta \right] \right| \theta \right]\right) 
\\
&\leq \sum_{s=1}^{S}\left(p_{K}\bar{d}\mathbb{E}^{\pi}\left[p_{K}\left(U_{t,k(t;s);s} - \hat{\lambda}_{t,k(t;s)}(\theta) \right)^{+} \bigg|\theta \right]\right) 
\\
& \leq 12p_{K}\bar{d}\sqrt{SK\log(SK)}.\quad \left(\text{by Lemma \ref{lem:UCBPi}}\right)
\end{align*}
The third underlined part (3) vanishes in the expectation over the parameter $\theta$ associated with a prior distribution $f$ because 
the $\theta_{s}$ and $\theta$ is identically and independently distributed on $f(\cdot | H_{0}^{s})$ given a history $H_{0}^{s}$. In fact, the underline part (3) is 
\begin{align*}
  %\label{eq:cancel}
  \mathbb{E}_{\theta} \left[\mathbb{E}^{\pi}\left[\left.  \left( \mathcal{U}^{*}_{t;s} - \mathcal{U}^{\pi}_{t;s} \right) \right| \theta \right] \right]  
  &= \mathbb{E}^{\pi} \left[\sum^{K}_{k=1}U_{t,k;s}  \mathbb{E}^{\pi} \left[\left(x_{t,k}(\theta) - x_{t,k}(\theta_{s}) \right) \bigg| H_{0}^{s} \right] \right]
  \\
  &=0. \quad \left(\theta, \theta_s \sim f(\cdot| H^s_{0}) \right)
\end{align*}
Finally, we have proven Lemma~\ref{lem:revenuediff}.
\end{proof}

\begin{lem}
  \label{lem:lostsales1}
\begin{align}
\label{eq:proof3}
p_{K}\mathbb{E}_{\theta}\left[\mathbb{E}^{\pi}\left[\sum_{s=1}^{S}\left(\sum_{t=1}^{T}D_{t,s} -n_{0}\right)^{+} \bigg|\theta \right]\right] 
\leq p_{K}\bar{d} \left({36T\sqrt{SK\log(K)}} + S\sqrt{T}\right).
\end{align}
\end{lem}
\begin{proof}
First, we decompose the left-hand-side as 
\begin{align*}
&\mathbb{E}^{\pi}\left[\sum_{s=1}^{S}\left(\sum_{t=1}^{T}D_{t,s}-n_{0}\right)^{+} \bigg| \theta\right] 
\\ 
&= \mathbb{E}^{\pi}\left[\sum_{s=1}^{S}\left(\sum_{t=1}^{T}D_{t,s} - \sum_{t=1}^{T}\sum_{k=1}^{K}\lambda_{t,k}(\theta)x_{t,k}(\theta_{s}) 
+ \sum_{t=1}^{T}\sum_{k=1}^{K}\lambda_{t,k}(\theta)x_{t,k}(\theta_{s}) - n_{0}\right)^{+} \bigg| \theta \right] 
\nonumber \\
&\leq \underset{(1)}{\underline{\mathbb{E}^{\pi}\left[\sum_{s=1}^{S}\left(\sum_{t=1}^{T}\left(D_{t;s}- \sum_{k=1}^{K}\lambda_{t,k}(\theta)x_{t,k}(\theta_{s})\right)\right)^{+}\bigg| \theta \right]}} 
+ \underset{(2)}{\underline{\mathbb{E}^{\pi} \left[\sum_{s=1}^{S}\left(\sum_{t=1}^{T}\sum_{k=1}^{K}\lambda_{t,k}(\theta)x_{t,k}(\theta_{s}) - n_{0}\right)^{+}\bigg| \theta \right]}}.
\end{align*}
The underlined term $(1)$ is bounded above by Lemma \ref{lem:loststatic} as 
\begin{align*}
  \mathbb{E}^{\pi}\left[\left. \sum_{s=1}^{S}\left(\sum_{t=1}^{T}\left(D_{t;s}-\sum_{k=1}^{K}\lambda_{t,k}(\theta)x_{t,k}\left(\theta_{s}\right)\right)\right)^+ \right| \theta \right] \leq S\bar{d} \sqrt{T},
\end{align*}
and the underlined term $(2)$ is bounded above using lemma \ref{lem:lostsales2} as
\begin{align*}
  \label{eq:lostsales2}
  p_K\mathbb{E}_{\theta}\left[\mathbb{E}^{\pi} \left[\left. \sum_{s=1}^{S}\left(\sum_{t=1}^{T}\sum_{k=1}^{K}\lambda_{k,t}(\theta)x_{t,k}(\theta_{s})-n_{0}\right)^{+}\right| \theta \right] \right] 
  \leq 36p_{K}\bar{d}T\sqrt{SK\log(K)}.
\end{align*}
Thus, we have proved \eqref{eq:proof3}.
\end{proof}

\section{Proof of Theorem~1} 
\label{sec:proofth1}
\begin{proof}
First, we evaluate an upper bound of ${\rm Regret}\left(T,S, \theta, \pi\right)$.
The first term of the regret \eqref{def:regret} is rewritten as 
\begin{align*}
\mathbb{E}^{\pi^{*}}\left[\left. \sum_{s=1}^{S}\sum_{t=1}^{T}\tilde{R}_{t;s} \right|\theta\right] = 
S\left(\mathbb{E}^{\pi^{*}}\left[\left.\sum_{t=1}^{T}\tilde{D}_{t}P_{t} \right|\theta\right]\right),
\end{align*}
because the optimal pricing policy $\pi^{*}(\theta)$ is independent on episodes. 
The second term of \eqref{def:regret} also can be decomposed into two terms with $R_{t;s} = P_{t;s}D_{t;s}$ as 
\begin{align}
  \mathbb{E}^{\pi}\left[\left. \sum_{s=1}^{S}\sum_{t=1}^{T}\tilde{R}_{t;s} \right|\theta \right] 
  &= \mathbb{E}^{\pi}\left[\left. \sum_{s=1}^{S}\sum_{t=1}^{T}\left(R_{t;s}-R_{t;s} + \tilde{R}_{t;s} \right) \right|\theta \right] 
  \nonumber \\ 
  &=  \mathbb{E}^{\pi}\left[\left.\sum_{s=1}^{S}\sum_{t=1}^{T}R_{t;s} \right|\theta \right] 
  - \mathbb{E}^{\pi}\left[\left. \sum_{s=1}^{S}\sum_{t=1}^{T} \left(R_{t;s} - \tilde{R}_{t;s} \right) \right|\theta \right].
\end{align}
The last term is the lost sales part and can be expressed as
\begin{align}
\label{def:totalsales}
\mathbb{E}^{\pi}\left[\left. \sum_{s=1}^{S}\sum_{t=1}^{T} \left(R_{t;s} - \tilde{R}_{t;s} \right) \right|\theta \right] 
&=
\mathbb{E}^{\pi}\left[\left. \sum_{s=1}^{S} \left(\sum_{t=\tau(s) + 1} ^{T} P_{t,s}D_{t,s} + 
P_{\tau(s),s}\left(D_{\tau(s),s} - n_{\left(\tau(s)-1\right),s}\right)\right) \right| \theta \right],
\end{align}
where $ \mathcal{T}  = \{t \in [T] \ |\ n_{t-1;s} - D_{t,s} \leq 0 \}$ and $\tau(s) = \min \mathcal{T}$. 
The first term in \eqref{def:totalsales} is total lost sales after the inventory runs out. 
The second term is the lost sale occurring at the time when the inventory runs out.

      To evaluate the first term of \eqref{def:totalsales},  
      we use the fact that the revenue $P_{t;s}D_{t;s}$ is bounded above by $p_{K}D_{t;s}$ because offering the shut-off price $p_{\infty}$ yields no revenue with probability one.  
      In fact, we have 
      \begin{align}
        \label{eq:lost1stbound}
        \mathbb{E}^{\pi}\left[\left. \sum_{t=\tau(s)+1}^{T}D_{t;s}P_{t;s}\right| \theta\right] & = \mathbb{E}^{\pi}\left[\left. \mathbb{E}^{\pi}\left[\left. \sum_{t=\tau(s) + 1}^{T} P_{t;s}D_{t;s}    \right| \theta_{s}, H_{0}^{s}, \theta\right]\right| \theta \right]\nonumber 
        \\
        &= \mathbb{E}^{\pi}\left[\left.  \sum_{w=1}^{T}\mathbb{P}\left(\left. \tau(s)=w \right|\theta_s,  H_{0}^{s}, \theta \right)\sum_{t=w + 1}^{T} \sum_{k=1}^{K}p_{k}\lambda_{t;s}(\theta)x_{t,k}(\theta_s)  \right| \theta \right] \nonumber 
        \\
        & \leq p_{K} \mathbb{E}^{\pi}\left[\left.  \sum_{w=1}^{T}\mathbb{P}\left(\left. \tau(s)=w \right|\theta_s,  H_{0}^{s}, \theta \right) \sum_{t=w + 1}^{T}\sum_{k=1}^{K} \lambda_{t;s}(\theta)x_{t,k}(\theta_s)  \right| \theta \right]\nonumber  \quad (p_{k} \leq p_{K}, \forall k \in [K]).
        \\
        & =p_{K} \mathbb{E}^{\pi}\left[\left. \sum_{t=\tau(s)+1}^{T}D_{t;s} \right| \theta \right], 
      \end{align}
      where $\tau(s) = T+1$ means that the inventory remains over the selling season.  
      
      From the definition of $\tau(s)$, we have $D_{\tau(s),s} \geq  n_{\left(\tau(s)-1\right),s} > 0 $, which implies that 
      the shutoff price $p_{\infty}$ is not offered at $t=\tau(s)$. We then have 
      \begin{align}
        \label{eq:lost2nd}
      \mathbb{E}^{\pi}\left[P_{\tau(s),s}\left(D_{\tau(s),s} - n_{\left(\tau(s)-1\right),s}\right)| \theta \right] \leq p_{K}\mathbb{E}^{\pi}\left[\left(D_{\tau(s),s} - n_{\left(\tau(s)-1\right),s}\right)| \theta \right],\quad \left(P_{\tau(s),s} \leq p_{K} \right), 
      \end{align}
      The lost sales part is then bounded by the upper bonds we have shown in \eqref{eq:lost1stbound} and \eqref{eq:lost2nd} as 
      \begin{align}
      \label{eq:upperlostsales}
      \mathbb{E}^{\pi}\left[\sum_{s=1}^{S}\sum_{t=1}^{T} \left( R(t,s) - \tilde{R}(t,s) \right)\bigg| \theta \right] \leq&  
      p_{K}\mathbb{E}^{\pi}\left[\sum_{s=1}^{S} \left(\sum_{t=\tau(s) + 1} ^{T} D_{t,s} + 
      \left(D_{\tau(s),s} - n_{\left(\tau(s)-1\right),s}\right)  \right) \bigg| \theta \right], \nonumber \\
      =& p_{K}\mathbb{E}^{\pi}\left[\sum_{s=1}^{S} \left(\sum_{t=1}^{T} D_{t,s} - n_0 \right)^{+}
       \bigg| \theta \right]\quad (\text{by } n_{t,s}=n_{t-1,s} - D_{t;s}) 
      %\leq & p_{K}\mathbb{E}^{\pi}\left[\sum_{s=1}^{S} \left(\sum_{t=1}^{T} D_{t,s} - n\left( 0 \right) \right)^{+}\bigg| \theta \right]
      \end{align}
      We then bound the regret as 
      \begin{align}
      \label{eq:Regret}
      \mathrm{Regret}\left(T,S,\theta, \pi\right) 
      &= S \left(\mathbb{E}^{\pi^*}\left[\sum_{t=1}^{T}\tilde{R}_{t;s}\bigg| \theta \right]\right)- \mathbb{E}^{\pi}\left[\sum_{t=1}^{T}\sum_{s=1}^{S}\left(\tilde{R}_{t;s} - R_{t;s} + R_{t;s}\right)\bigg|\theta\right]\nonumber \\
      &= S \left(\mathbb{E}^{\pi^*}\left[\sum_{t=1}^{T}\tilde{R}_{t;s}\bigg| \theta\right]\right)- \mathbb{E}^{\pi}\left[\sum_{t=1}^{T}\sum_{s=1}^{S}R_{t;s}\bigg|\theta\right] +
       \mathbb{E}^{\pi}\left[\sum_{t=1}^{T}\sum_{s=1}^{S}\left(R_{t;s} - \tilde{R}_{t;s}\right)\bigg| \theta\right]\nonumber \\
      &= S \left(\mathbb{E}^{\pi^*}\left[\sum_{t=1}^{T}\tilde{R}_{t;s}\bigg| \theta\right]\right) - \mathbb{E}^{\pi}\left[\sum_{t=1}^{T}\sum_{s=1}^{S}R_{t;s}\bigg|\theta\right] \nonumber \\
      &\quad + \mathbb{E}^{\pi}\left[\sum_{s=1}^{S} \left(\sum_{t=\tau(s) + 1} ^{T} P_{t,s}D_{t,s} + 
      P_{\tau(s),s}\left(D_{\tau(s),s} - n_{\left(\tau(s)-1\right),s}\right)\right) \bigg| \theta \right]\quad  (\text{by } \eqref{def:totalsales})\nonumber \\
      &\leq \underset{(a)}{\underline{S\left(\mathbb{E}^{\pi^{*}}\left[\sum_{t=1}^{T}\tilde{R}(t) \bigg| \theta \right]\right) - 
      \mathbb{E}^{\pi}\left[\sum_{s=1}^{S}\sum_{t=1}^{T}R(t,s) \bigg| \theta \right]}} + \underset{(b)}{\underline{p_K\mathbb{E}^{\pi}\left[\sum_{s=1}^{S} \left(\sum_{t=1}^{T}D_{t,s} - n_ 0 \right)^{+} \bigg| \theta \right]}},
      \quad (\text{by }  \eqref{eq:upperlostsales}) \nonumber 
      \end{align}
      We refer the underlined part $(a)$ to the revenue-difference part and the other part $(b)$ to the total lost sales part.
      The expectation of the revenue-difference part over $\theta$ is bounded above using Lemma \ref{lem:revenuediff}:  
      \begin{align}
        \mathbb{E}_{\theta} \left[S\left(\mathbb{E}^{\pi^{*}}\left[\left. \sum_{t=1}^{T}\tilde{R}(t) \right|\theta \right]\right) - 
        \mathbb{E}^{\pi}\left[\left. \sum_{s=1}^{S}\sum_{t=1}^{T}R(t,s) \right| \theta \right] \right]\leq 18 p_{K}\bar{d}T\sqrt{KS\log(K)},
      \end{align}
       The expectation of underlined part $(b)$ is also bounded above using Lemma \ref{lem:lostsales1}: 
      \begin{align}
        p_{K}\mathbb{E}_{\theta}\left[\mathbb{E}^{\pi}\left[\sum_{s=1}^{S}\left(\sum_{t=1}^{T}D_{t,s} -n(0)\right)^{+} \bigg|\theta \right]\right] 
        \leq p_{K}\bar{d} \left(36T\sqrt{SK\log(K)} + S\sqrt{T}\right).
        \end{align}
      Therefore, we have proved Theorem~1. 
      \end{proof}

\section{Lemmas for Theorem 2}
First, we define a revenue management problem in static demand without inventory constraints, 
and use it in Lemma~\ref{lem:minmaxlower}. This problem is a type of conventional multi-armed bandit problems. 
\begin{dfn}[RM problem in static demand without inventory constraints]
    \label{def:RMsta}
    {\rm
    For each $s \in [S]$, an algorithm $\pi$ offers a price $P_{s} \in \mathcal{P}$. 
    Then, the algorithm observes demand $D_{s}\geq 0$, which is independent of the past prices and demands. 
    The distribution under the offered price $p_k$ is denoted by $\mathcal{D}_{k}(\theta_0)$, 
    where $\theta_0 \in \Theta_0$ is unknown to the algorithm.
    The algorithm obtains revenue $D_s P_s$.
}
\end{dfn}
We define the \pregret{} of this revenue management as
\begin{dfn}
    \label{def:RMsta_reg}{\rm
    For the RM problem in definition~\ref{def:RMsta}, the problem-dependent regret of an algorithm $\pi$ under a demand parameter $\theta_0 \in \Theta_0$ is defined
    \begin{align}
        \label{eq:RMsta_reg}
        {\rm Regret_{sta}}(S, \theta_0, \pi) = S \max_{k\in [K]}\left(p_k \lambda_k(\theta_0)\right) 
        - \sum_{s=1}^{S}\mathbb{E}^{\pi}\left[\left. D_s P_s \right| \theta_0\right],
    \end{align}
    where $\lambda_{k}(\theta_0) = \mathbb{E}^{\pi}\left[ \left. D_s \right| P_{s}=p_k, \theta_0\right]$. 
    }
\end{dfn}

The Bayesian regret for this problem is also given in the same manner as in \eqref{def:Bregret}. 
\begin{dfn}[Bayes Regret]
    The Bayesian regret of an algorithm $\pi$ for a prior $f$ over $\Theta_0$ is  
    \label{def:RMsta_bayes_reg}{\rm
    \begin{align}
    {\rm BRegret_{sta}}(S, f, \pi) = \mathbb{E}_{\theta_0}\left[  {\rm Regret_{sta}}(S, \theta_0, \pi) \right],
    \end{align}
    where $\mathbb{E}_{\theta_{0}}\left[\cdot \right]$ denotes the expectation taken for $\theta_0$ following $f$. 
}    
\end{dfn}

Finally, we introduce the KL-divergence to use it in the proof of Lemma~\ref{lem:minmaxlower}. 
\begin{dfn}[KL-divergence]
    \label{dfn:RMsta}
    Let $\left(\Omega, \mathcal{F}\right)$ be a measurable space. 
    For probability measures $P$ and $Q$ on this space, 
    the KL-divergence $D(P,Q)$ is defined as  
    \begin{align}
        D(P, Q) = \begin{cases}\displaystyle
                \int_{\Omega} \log\left(\frac{dP}{dQ}\right)dP(\omega), &\quad if P \ll Q
                    \\ \displaystyle
                    \infty, &\quad otherwise,
                \end{cases}
    \end{align}
    where $\frac{dP}{dQ}$ is the Radon-Nikodim derivative of $P$ with respect to $Q$,  
    and $P\ll Q$ means that $Q(A)=0 \Rightarrow P(A)=0$ for all $A\in\mathcal{F}$. 
\end{dfn}

\subsection{Lemmas}
The toolkits used here are general ones in analysis of multi-armed bandit problems, 
and readers would refer to from Section~13 to 15 of \citet{lattimore2020bandit} 
for more detailed information of them.

\begin{lem}[Bretagnolle-Huber inequality]
    \label{lem:prob_lower}
Let $P$ and $Q$ be probability measures 
on the same measurable space $\left(\Omega, \mathcal{F}\right)$. 
For any event $A \in \mathcal{F}$, we have
\begin{align}
    P(A) + Q(A^C) \geq \frac{1}{2} \exp\left(- D(P,Q)\right). 
\end{align}
\end{lem}
The proof is given in Theorem~14.2 of \citet{lattimore2020bandit}.

\begin{lem}[Lemma~15.1 of \citet{lattimore2020bandit}]
    \label{lem:KL-decompose}
Let $N_{k}(T)$ be the number of times when the $k$-th arm is chosen and $P_{k}$ and $P'_{k}$ 
be $k$-th reward distributions of bandit instances $\nu$ and $\nu'$. For some fixed algorithm $\pi$,   
the KL-divergence $D(\mathbb{P}_{\pi\nu},  \mathbb{P}_{\pi\nu'})$ is decomposed as 
\begin{align}
    D(\mathbb{P}_{\pi\nu},  \mathbb{P}_{\pi\nu'}) = \sum_{k=1}^{K}\mathbb{E}_{\nu}\left[N_{k}(T)\right]D(P_{k}, P'_{k}).
\end{align}
\end{lem}
Here, $\mathbb{P}_{\pi\nu}$ is a probability measure on the measurable space $(\Omega_T, \mathcal{F}_T )$ 
that are determined by the algorithm $\pi$  and the bandit instance $\nu$. See \cite{lattimore2020bandit} for the formal description of this probability measure and the proof of this lemma.   
The proof of the following lemma depends on that of Lemma~15.2 of \cite{lattimore2020bandit}. 
\begin{lem}
    \label{lem:minmaxlower}
Consider the RM problem defined in Definition~\ref{def:RMsta}.  
For any algorithm $\pi$, there exists a prior distribution $f_0$ such that the Bayesian regret satisfies 
\begin{align} 
{\rm BRegret_{sta}}(S, f_0, \pi) \geq  \Omega\left(\sqrt{\left(K-1\right)S} \right).
\end{align}
\end{lem}
\begin{proof}
Let $\mathcal{B}(\theta_0)$ be a set of demand distributions determined by $\theta_0=(\theta_{01}, \theta_{02}, \dots, 
\theta_{0K}) \in (0,\infty)^{K}$, where the demand distribution for $k$-th price, $\mathcal{D}_{k}(\theta_0)$, is the beta distribution ${\rm Beta}(\theta_{0k}, 1)$. 
In what follows, we introduce two demand parameters $\nu$ and $\nu' \in (0,\infty)^{K}$ to bound the regret below.

For an arbitrary number $0<\varepsilon<1$, 
we define the following parameters:
\begin{align}
         a^{-}_{k} = \frac{\eta_k\left(1-\varepsilon\right)}{2- \eta_k(1-\varepsilon)}, 
        \quad a^{+}_{k} = \frac{\eta_k\left(1+\varepsilon\right)}{2 - \eta_k(1+\varepsilon)}, 
\end{align}
where $\eta_k = \frac{p_1}{p_k} \leq 1$. 
With these parameters, we define a demand parameter $\nu = (1,a_{2}^{-}, \dots, a^{-}_{K}) \in (0, \infty)^{K}$. 
In this setting, the mean revenue is $p_1/2$ for $k=1$ and $p_{1}(1-\varepsilon)/2$ for $k=2, \dots, K$. 
Then, $p_1$ is the optimal price. We define the arm $i_{0}(\nu) \in [K]$ as
\begin{align*} 
    i_0(\nu) = \argmin_{j>1}\mathbb{E}^{\pi}\left[\left. N_{k}(S)\right| \nu\right], 
\end{align*}
where $N_{k}(S)$ is the number of times when the $k$-th price is offered until the round $S$, 
which is defined as $\sum_{s=1}^{S}\mathbf{1}\left(P_{s}=p_k\right)$. 
Since $\sum_{k=1}^{K}N_{k}(S)=S$, we have   
\begin{align} 
    \mathbb{E}^{\pi}\left[\left. N_{i_{0}(\nu)}\right| \nu\right] \leq \frac{S}{K-1}.
\end{align}
For the parameter $\nu$, the regret for any algorithm $\pi$ is bounded below as 
\begin{align}
    {\rm Regret_{sta}}(S, \nu, \pi) &= \sum_{k>1}^{K}\frac{p_1\varepsilon }{2}\expv^{\pi}\left[ \left. N_{k}(S) \right| \nu \right]\nonumber
    \\
    &\geq \frac{p_1 \varepsilon}{2}\expv^{\pi}\left[\left. \left(\sum_{k>1}^{K}N_{k}(S)\right) 
    \left(\mathbf{1}\left(\sum_{k>1}^{K}N_{k}(S) \geq \frac{S}{2}\right) + 
    \mathbf{1}\left(\sum_{k>1}^{K}N_{k}(S) < \frac{S}{2}\right)\right) \right| \nu \right]\nonumber
    \\
    &\geq \frac{p_1 \varepsilon}{2}\expv^{\pi}\left[\left. \left(\sum_{k>1}^{K}N_{k}(S) \right) \mathbf{1}\left(\sum_{k>1}^{K}N_{k}(S) \geq \frac{S}{2} \right)
    \right| \nu \right]\nonumber 
    \\
    & \geq \frac{p_1S\varepsilon }{4}\expv^{\pi}\left[\left. \left( \mathbf{1}\left(N_{1}(S) \leq \frac{S}{2}\right)\right) \right| \nu
    \right]\quad  \left( S - \sum_{k>1}^{K}N_{k}(S)=N_{1}(S)\right)\nonumber 
    \\
    \label{eq:nulower}
    &= \frac{p_1S \varepsilon }{4} \mathbb{P}_{\pi\nu} \left(N_{1}(S) \leq \frac{S}{2}\right).
\end{align}

Next, we introduce another demand parameter $\nu' = (1, \dots,a_{i_{0}(\nu)}^{+}, \dots, a_{K}^{-})$, 
where the $i_{0}(\nu)$-th parameter of $\nu$ is replaced with $a^{+}_{i_{0}(\nu)}$. 
By definition, the $i_{0}(\nu)$-th price is the optimal as the mean revenue for the price is $p_1(1+\varepsilon)/2$. 
For this parameter, the regret of the algorithm $\pi$ is bounded as
\begin{align}
    {\rm Regret_{sta}}(S,\nu', \pi) &= p_1\left(\frac{\varepsilon}{2}\expv^{\pi}\left[\left. N_{1}\left(S\right)\right| \nu'\right] 
    + \varepsilon \sum_{k\neq 1, i_0(\nu)}^{K}\expv^{\pi}\left[ \left. N_{k}(S) \right| \nu'\right]\right)
    \nonumber
    \\
    &\geq  \frac{p_1\varepsilon}{2}\expv^{\pi}\left[\left. N_{1}\left(S\right) \right| \nu'\right]\quad \left(N_{k}(S) \geq 0,\ k\in[K]\right) \nonumber
    \\
    &= \frac{p_1\varepsilon}{2}\expv^{\pi}\left[ \left. N_{1}\left(S\right) \left( \mathbf{1}\left(N_{1}(S)\leq \frac{S}{2}\right) 
    + \mathbf{1}\left(N_{1}(S) > \frac{S}{2}\right)\right)\right| \nu' \right] 
    \nonumber \\
    & \geq \frac{p_1\varepsilon}{2} \expv^{\pi}\left[ \left. N_{1}\left(S\right)\mathbf{1}\left(N_{1}(S) > \frac{S}{2}\right) \right| \nu'\right]
    \nonumber \\
    & \geq \frac{p_1 S\varepsilon }{4} \expv^{\pi}\left[\left. 
        \left(\mathbf{1}\left(N_{1}(S) > \frac{S}{2}\right)\right) \right| \nu' \right] 
    \nonumber \\
    \label{eq:nu2lower}
    &= \frac{p_1 S\varepsilon}{4} \mathbb{P}_{\pi\nu'}\left( N_{1}(S) > \frac{S}{2}\right).
\end{align}
The inequalities \eqref{eq:nulower} and \eqref{eq:nu2lower} enable 
us to derive a lower bound of the summation of the regrets for the two parameters $\nu$ and $\nu'$. 
\begin{align}
    {\rm Regret_{sta}}\left(S, \nu, \pi \right) + 
    {\rm Regret_{sta}}\left(S, \nu', \pi \right) &\geq \frac{p_1 S \varepsilon}{4}\left(\mathbb{P}_{\pi\nu} \left(N_{1}(S) \leq \frac{S}{2}\right)
    + \mathbb{P}_{\pi\nu'}\left( N_{1}(S) > \frac{S}{2}\right)\right) 
    \nonumber\\
    & \geq \frac{p_1S\varepsilon }{8} \exp\left(-  D\left(\mathbb{P}_{\pi\nu}, \mathbb{P}_{\pi\nu'}\right)\right)
    \quad (\text{by Lemma \ref{lem:prob_lower}}) 
    \nonumber \\
    &= \frac{p_1S\varepsilon}{8} \exp\left(-   \expv[\left. N_{i_0(\nu)}\right|\nu ] D\left(P_{i_0(\nu)}, P'_{i_0(\nu)}\right)\right)
    \quad \text{(by Lemma \ref{lem:KL-decompose}) }
    \nonumber \\
    &\geq \frac{p_1S\varepsilon}{8} \exp\left(-  \frac{S}{\left(K-1\right)}D\left(P_{i_0(\nu)}, P'_{i_0(\nu)}\right)\right)
    \quad \left(\expv^{\pi}[\left. N_{i_0(\nu)} \right| \nu] \leq \frac{S}{K-1}\right) 
    \nonumber \\ 
    \label{eq:regcomb}
    &\geq \frac{ p_1 S\varepsilon}{8}\left(1 - \frac{S}{\left(K-1\right)}D\left(P_{i_0(\nu)}, P'_{i_0(\nu)}\right)\right)\quad \left(\e^{-x} \geq 1-x\right).
\end{align}
The KL-divergence $D\left(P_{i_0(\nu)}, P'_{i_0(\nu)}\right)$ is computed as 
\begin{align}
    \label{eq:KL-div}
    D\left(P_{i_0(\nu)}, P'_{i_0(\nu)}\right) &= 
    \frac{4\gamma}{\eta_{0}} \frac{\varepsilon}{(1 - \gamma \varepsilon)(1 - \varepsilon)} + \log \left(\frac{1 - \varepsilon}{1 + \varepsilon} \frac{1 - \gamma \varepsilon}{1 + \gamma \varepsilon}\right) 
\end{align}
where $\gamma = \frac{\eta_{0}}{2 - \eta_{0}} \leq 1$, and $\eta_{0}$ denotes $\eta_{i_{0}(\nu)}$ for notational simplicity.
The KL-divergence has the power series
\begin{align}
    D\left(P_{i_0(\nu)}, P'_{i_0(\nu)}\right) = 8\left(\frac{\varepsilon}{2 - \eta_0}\right)^2 + \mathcal{O}\left(\varepsilon^3\right).
\end{align}
This immediately means that setting $\varepsilon$ to, for example,  
\begin{align}
     \varepsilon_{0} = \frac{2-\eta_0}{2}\sqrt{\frac{\left(K-1\right)}{4S}} <\frac{1}{2}, 
\end{align}
leads to an inequality 
\begin{align}
    \label{eq:minimaxreg_lem_lower}
    {\rm Regret_{sta}}\left(S, \nu, \pi\right) + {\rm Regret_{sta}}\left(S, \nu', \pi\right) \geq \frac{p_1}{32}\sqrt{S(K-1)}.
\end{align}

Then, we consider the parameter space $\Theta_0 = \{\nu, \nu'\}$ and a prior distribution $f_0$ that is uniformly distributed on $\Theta_0$. 
The equation \eqref{eq:minimaxreg_lem_lower} enables us to derive the Bayesian regret lower bound
\begin{align} 
{\rm Bregret_{sta}}(S, f_0, \pi) = \frac{1}{2}\left({\rm Regret_{sta}}\left(S, \nu, \pi\right) 
+ {\rm Regret_{sta}}\left(S, \nu', \pi\right)\right) \geq \Omega(\sqrt{S(K-1)}).
\end{align}
\end{proof}

\subsection{Proof of Theorem~2}
\label{sec:proofofthm2}
\begin{proof} 
To simplify the analysis, we set $\bar{d}$ to 1\footnote{
To recover the dependency on $\bar{d}$, 
we use $m= \lceil \frac{n_{0}}{\bar{d}} \rceil$ instead of $n_0$ in this proof.   
Then, $\bar{d}$ only appears as a overall factor of the Bayesian regret. 
}. 
Let $n_0$ be an integer satisfying that $1 \leq n_{0} < T$.  
We choose a subset $\mathcal{T}_{0} = [n_0]$ of $[T]$ 
and define a demand space as $\Theta = (0,\infty)^{n_0K}$. 
We consider the demand model where for a demand parameter $\theta = \{\theta_{t,k}\}_{t\in\mathcal{T}_0, k \in [K]} \in \Theta$, 
the demand distribution $\mathcal{D}_{t,k}(\theta) = {\rm Beta}(\theta_{tk}, 1)$ for $k \in [K]$ and $t \in \mathcal{T}_0$, 
and for $t \notin \mathcal{T}_0$, the demand $D_{t;s} =0$ with probability one. 
Under this demand model, the initial inventory never runs out 
since $\sum_{t\in T}\tilde{D}_{t;s} = \sum_{t\in \mathcal{T}_0} \tilde{D}_{t;s}  \leq n_0$. 
Then, the \pregret{} is decomposed as

\begin{align}
    {\rm Regret}\left(S, T, \theta, \pi\right) &= S\rev{T}{n_0}{\theta} - \expvp{\sum_{s=1}^{S}\sum_{t=1}^{T}\tilde{D}_{t,s}P_{t,s}}{\theta}
    \nonumber \\
    &= \sum_{t \in \mathcal{T}_0}\sum_{s=1}^{S} \left(  \mathbb{E}^{\pi^{*}(\theta)}\left[\left. D_{t;s}P_{t;s} \right| \theta \right] 
    - \expvp{D_{t,s}P_{t,s}}{\theta} \right)
    \quad \left(\tilde{D}_{t,s}=D_{t,s}\text{ a.e.}\right)
    \nonumber \\
    &= \sum_{t\in \mathcal{T}_{0}}\left(
        S\max_{k \in [K]}\left(p_{k}\lambda_{t,k}(\theta_t)\right) - \sum_{s=1}^{S}\expv^{\pi}\left[ \left. D_{t;s}P_{t;s} \right| \theta\right]
        \right)
    \nonumber \\
    &= \sum_{t\in \mathcal{T}_{0}} {\rm Regret_{sta}}[S,\theta_t, \pi] \quad \left( \text{By \eqref{eq:RMsta_reg}} \right). 
\end{align}
By Lemma~\ref{lem:minmaxlower}, there exists a prior distribution $f_{0t}$ such that
${\rm BRegret_{sta}}[S,f_{0t}, \pi] \geq \Omega(\sqrt{S(K-1)})$ for $t \in \mathcal{T}_0$. 
From the prior distributions, we construct a prior distribution $f_0 = \prod_{t \in \mathcal{T}_0}f_{0t}$.   
The Bayesian regret associated with $f_{0}$ is bounded below as
\begin{align}
    {\rm BRegret}\left(S, T, f_0, \pi\right) 
    &= \sum_{t \in \mathcal{T}_0} \mathbb{E}_{\theta_t \sim f_{0t}}\left[{\rm Regret_{sta}}\left(S,\theta_t, \pi\right)\right]
    \nonumber \\
    & \geq \sum_{t \in \mathcal{T}_0} \Omega(\sqrt{S(K-1)})\quad ( \text{ By Lemma~\ref{lem:minmaxlower}})
    \nonumber \\
    &= n_{0}\Omega(\sqrt{S(K-1)})
\end{align}
The same argument can be applied to the case when $n_0 \geq T$ 
because inventory never runs out with certainty. 
Thus, for this case, we have a lower bound of $\Omega\left(T\sqrt{S(K-1)}\right)$. 
Therefore, we have derived the lower bound $\Omega\left(\min(T,n_0)\sqrt{S(K-1)}\right)$.
\end{proof}

\section{On Extension to the Multiple Items Case}
\label{sec:multiple}

In this appendix, we show how to extend our problem to the case where a seller deals in multiple items with limited resources.  
We focus on the model that is often called the {\it network revenue management} proposed in \citet{gallego1997multiproduct}. 
The multiple-item extension of our problem can result in a slight modification of the design of our algorithms. 
This extension also slightly modifies Theorem~1.  
In what follows, we confirm this statement by starting with the extension of our problem setting. 
The following argument is based on \citet{ferreira2018online}.

We consider that a seller deals in $N$ items indexed by $i \in [N]$ in our problem. 
Each of the items consumes $M$ resources, indexed by $j \in [M]$. 
One unit of the $i$-th item consumes $a_{ij}$ units of $j$-th resource, where $\{a_{ij}\}_{i \in [N], j\in[M]}$ is fixed and known to the seller. 
The remaining unit of $j$-th resource at the end of time period $t$ is denoted by $n_{j,t;s}$ and $n_{j,0;s} = n_{j,0}$, 
where $n_{j,0}$ is initial amount of $j$-th resource. 
The initial inventory of the $j$-th resource is fixed in $n_{j,0}$ in every episode. 
The set of initial inventories $\bm{n}$ is expressed as the $M$-dimensional vector $\bm{n} = \left(n_{1,0}, \dots, n_{M,0}\right)$.

According to these changes, the settings for prices and demand also change. 
For any time period $t$ in any $s$-th selling season,  
a sequence of offered prices $\bm{P}_{t;s}$ is chosen from the set of $N$-dimensional vectors, $\mathcal{P}\cup \{p _{\infty}\}$ with a size of  $K+1$.
For any $k \in [K]$, the $k$-th price of $\mathcal{P}$ is expressed as $\bm{p}_{k} = (p_{1k}, \dots, p_{Nk})$, where $p_{ik}$ for $i \in [N]$ is the price for $i$-th item.  
Under the shut-off price $p_{\infty}$, demand for every item is zero, and no revenue is obtained with probability one. 
Then, the demand $\bm{D}_{t;s}$ is also modified to take $N$ dimensional vector values. 
The demand $\bm{D}_{t;s}$ is expressed as $\bm{D}_{t;s}= (D_{1,t;s}, \dots, D_{N,t;s})$ using demand for $i$-th item $D_{i,t;s}$ for $i \in [N]$. 
The demand distribution $\mathcal{D}_{t,k}(\theta)$ is a joint distribution defined on $\mathbb{R}^{N}_{+}$. 
Then, the mean demand function is also defined for each item given a demand parameter $\theta \in \Theta$ as 
\begin{align*}
    \lambda_{i,t,k}(\theta) = \mathbb{E}\left[D_{i,t;s}\bigg| P_{t;s}=p_{k}, \theta\right], \ i \in [N].
\end{align*}

After observing demand, the seller consumes the resources to achieve revenue. 
If enough amounts of each resource are available, the seller obtains revenue $\bm{P}_{t;s}\cdot \bm{D}_{t;s}$,
where '$\cdot$' denotes the standard inner product in $\mathbb{R}^{N}$. 
In contrast, if some resources are insufficient, the revenue obtained depends on which products 
the seller allocates the insufficient resources to. 
Without depending on how demand is specifically satisfied, we define the satisfied demand $\tilde{D}_{i,t;s}$ for $i \in [N]$, 
which satisfies the following conditions as stated in \citet{ferreira2018online} 
\begin{align}
\label{eq:demandcond}
0\leq \tilde{D}_{i,t;s}\leq D_{i,t;s} \quad \forall i \in [N], \quad  
 n_{j,t;s} = n_{j,t-1;s} - \sum_{i=1}^{N}a_{ij}\tilde{D}_{i,t;s} \geq 0 \quad \forall j \in [M], \quad n_{j,t;s} = 0\quad \exists j \in [M].
\end{align}
We assume that our algorithm chooses a specific demand satisfaction rule so that the expected revenue is well-defined.  
Let $\tilde{D}_{t;s}$ denote the sequence of the satisfied demand $\tilde{D}_{i,t;s}$ for $i \in [N]$, 
which is expressed as $\tilde{\bm{D}}_{t;s} = (\tilde{D}_{1, t;s}, \dots, \tilde{D}_{N,t;s})$. 
Consequently, the seller actually obtains revenue $\bm{P}_{t;s} \cdot \bm{\tilde{D}}_{t;s}$.

From the argument so far, the only necessary modifications in our algorithms are to use the following linear optimization problem 
$ {\rm LP}_{\text{multi}}(\theta, t, \bm{n})$ over $\{ x_{\tau, k} \}_{t\leq \tau \leq T, k \in [K]} \in [0,1]^{(T-t+1)K}$:   
\begin{align}
    \text{maximize:\,} &\sum_{\tau =t}^{T}\sum_{k=1}^{K}x_{\tau,k}\left(\sum_{i=1}^{N}\lambda_{i,\tau,k}(\theta)p_{ik}\right)  \nonumber\\
     \text{subject to:\,} &\sum_{\tau=t}^{T}\sum_{k=1}^{K}x_{\tau,k} \left(\sum_{i=1}^{N}a_{ij}\lambda_{i,\tau, k}(\theta) \right)\leq n_{j}, \quad  \forall j \in [M], \nonumber \\
     &\sum_{k=1}^{K} x_{\tau,k} \leq 1, \quad \forall \tau \in \{t,t+1, \dots, T\}. 
 \label{def:LPmodified} 
 \end{align}
Then, the theoretical performance of \algone{} in this multi-item setting is given in the following theorem.

\begin{thm*}[The multiple-item generalization of Theorem 1]
Assume that there exist $\bar{d}_{i} >0$ for all $i \in [N]$ such that the support 
of the demand distribution $\mathcal{D}_{t,k}(\theta)$ is finite and included in the product of intervals $\prod_{i=1}^{N}[0, \bar{d}_{i}]$ for any $\theta \in \Theta$. 
Then, for $K \leq S$, the Bayesian regret \eqref{def:Bregret} of \algone{} satisfies
\begin{align}
\label{eq:modification}
{\rm Bregret}(T, S, \pi) \leq (18r'_{{\rm m}} + 36r_{{\rm pm}})T\sqrt{SK\log(K)} + r_{{\rm pm}}S\sqrt{T},
\end{align}
where 
\begin{align*}
r'_{{\rm m}} = \max_{k \in [K]}\sum_{i=1}^{N}p_{ik}\bar{d}_{i},\quad  p^{j}_{\max} = \max_{ \{i \in [N], k\in[K] | a_{ij} \neq 0 \} } \frac{p_{ik}}{a_{ij}}, \quad 
r_{{\rm pm}} = \sum_{i=1}^{N}\sum_{j=1}^{M} p_{\max}^{j} a_{ij} \bar{d}_{i}.
\end{align*}
\end{thm*}
The term $r'_{\text{m}}$ denotes the possible maximum revenue achieved in one period, and 
$p^{j}_{\max} $ denotes the possible maximum revenue per unit consumption of the $j$-th resource. 
Indeed, this is a generalization of the results of Theorem 1 since $p^{j}_{\max} = p_{K}$ and  $r'_{{\rm m}}=p_{K}\bar{d}$  when $N=1, M=1, a_{11}=1$.

\begin{proof}[Proof sketch]
The proof is almost parallel to that of Theorem 1 by defining $R_{t;s} = \bm{P}_{t;s} \cdot \bm{D}_{t;s}$ and $\tilde{R}_{t;s}= \bm{P}_{t;s} \cdot \tilde{\bm{D}}_{t;s}$.
However, the evaluation of the lost sales part is slightly different from that in Section \ref{sec:proofth1}. 
This evaluation can be modified by the conditions \eqref{eq:demandcond}.  
The conditions enable us to bound the lost sales part as 
\begin{align}
    \label{eq:multilost}
    \mathbb{E}^{\pi}\left[\left.  \sum_{s=1}^{S}\sum_{t=1}^{T} \left(R_{t;s}- \tilde{R}_{t;s} \right) \right| \theta \right] \leq 
    \sum_{s=1}^{S}\sum_{j=1}^{M}p_{\text{max}}^{j}\mathbb{E}\left[\left(\sum_{t=1}^{T}\sum_{i=1}^{N}a_{ij}D_{i,t;s} - n_{j,0}\right)^{+}\bigg| \theta \right].
\end{align}
To explain the inequality intuitively, recall that
$p_{\text{m}}^{j}$ is the possible maximum revenue per unit consumption of the $j$-th resource. 
In addition, $\mathbb{E}\left[\left(\sum_{t=1}^{T}\sum_{i=1}^{N}a_{ij}D_{i,t;s} - n_{j,0}\right)^{+}\bigg| \theta \right]$ 
represents the expected lost sales arising from the $j$-th resource. 
By multiplying these quantities, we can obtain an upper bound of the expected revenue associated with lost sales from the $j$-th resource. 
Consequently, the inequality \eqref{eq:multilost} holds.

In what follows, we show how to modify the evaluation of the revenue difference part and 
the lost sales part defined in the proof of Theorem 1.

\subsubsection{Total Revenue-difference Part} 
The revenue difference part is defined as 
\begin{align}
\label{eq:multirevdif}
\mathbb{E}^{\pi^{*}(\theta)}\left[\sum_{t=1}^{T}\sum_{s=1}^{S}\tilde{R}_{t}\bigg| \theta\right] - \mathbb{E}^{\pi}\left[\sum_{t=1}^{T}\sum_{s=1}^{S}R_{t;s}\bigg| \theta\right]. 
\end{align}
To bound the first term, 
we straightforwardly modify Lemma \ref{lem:rev*vsoptLP} for the multi-product case with the conditions \eqref{eq:demandcond}, 
which yields 
\begin{align*}
    \mathbb{E}^{\pi^{*}(\theta)}\left[\sum_{t=1}^{T}\tilde{R}_{t}\bigg| \theta \right] \leq \sum_{i=1}^{N}\sum_{t=1}^{T}\sum_{k=1}^{K}x_{t,k}(\theta)\lambda_{i,t,k}(\theta)p_{ik},
\end{align*}
where the right-hand side is the optimal value of ${\rm LP}_{\text{multi}}(\theta, 1, \bm{n}_{0})$. 
Subsequently, the parallel argument in the proof of Lemma \ref{lem:revenuediff} can be applied by introducing different upper confidence bounds. 
For any $i \in [N]$, we define $U_{i,k,t;s}$, which is the upper confidence bound defined in Section \ref{sec:UCBdef} for the $i$-th product demand $D_{i,t;s}$ 
and introduce the $N$-dimensional vector
\begin{align*}
\bar{\bm{U}}_{t,k;s} = \left(\bar{d}_{1}U_{1,t,k;s}, \dots,\bar{d}_{N}U_{N,t,k;s} \right).
\end{align*}
With these quantities, we define and use
\begin{align*}
\mathcal{U}^{\pi, \text{m}}_{t;s} =  \sum_{k=1}^{K}\bm{p}_{k} \cdot \bm{U}_{t,k;s}x_{t,k}(\theta_s), \quad 
\mathcal{U}^{*, \text{m}}_{t;s} =  \sum_{k=1}^{K}x_{t,k}(\theta)\bm{p}_{k} \cdot \bar{\bm{U}}_{t,k;s},
\end{align*} 
instead of $\mathcal{U}^{\pi}_{t;s}$ and $\mathcal{U}^{*}_{t;s}$. 
Consequently, we can obtain the upper bound of \eqref{eq:multirevdif}, 
which is that in Lemma \ref{lem:revenuediff} with $p_{K}\bar{d}$ replaced with $r'_{m}$. 
That is, we have an upper bound of \eqref{eq:multirevdif}
\begin{align}
\label{eq:multirevdifupper}
18r'_{\text{m}} T\sqrt{SK\log(K)}.
\end{align}

\subsubsection{Lost Sales Part}
We apply the argument in the proof of Lemma \ref{lem:lostsales1} to the term 
\begin{align*}
\sum_{s=1}^{S}p_{m}^{j}\mathbb{E}\left[\left(\sum_{t=1}^{T}\sum_{i=1}^{N}a_{ij}D_{i,t;s} - n_{j,0}\right)^{+}\bigg| \theta \right], \quad \forall i \in [M].
\end{align*} 
Then, we have the upper bound on the right-hand side 
\begin{align*}
p_{m}^{j}\underset{(1)}{\underline{\mathbb{E}^{\pi}\left[\sum_{s=1}^{S}\left(\sum_{t=1}^{T}\sum_{i=1}^{N}a_{ij}\left(D_{i,t;s}- \sum_{k=1}^{K}\lambda_{i,k,t}(\theta)x_{t,k}(\theta_{s})\right)\right)^{+}\bigg| \theta \right]}} + 
p_{m}^{j} \underset{(2)}{\underline{\mathbb{E}^{\pi} \left[\sum_{s=1}^{S}\left(\sum_{t=1}^{T}\sum_{k=1}^{K}\sum_{i=1}^{N}a_{ij}\lambda_{i,k,t}(\theta)x_{t,k}(\theta_{s}) - n_{j,0}\right)^{+}\bigg| \theta \right]}}.
\end{align*}
For the underlined part $(1)$, we can modify Lemma \ref{lem:loststatic} straightforwardly and obtain the upper bound, whit $\bar{d}$ replaced with $\sum_{i=1}^{N}a_{ij}\bar{d}_{i}$. 
For the underlined part $(2)$, we can apply the argument in the proof of \ref{lem:lostsales2} by modifying the upper and lower confidence bound used there. 
We define $U_{j,t,k;s}$ and $L_{j,t,k;s}$ as 
\begin{align*}
U_{j,t,k;s} = \sum_{i=1}^{N}a_{ij}\bar{d}_{i}U_{i,t,k,s}, \quad  L_{j,t,k;s} = \sum_{i=1}^{N}a_{ij}\bar{d}_{i}L_{i,t,k,s},
\end{align*}
and use them instead of $U_{t,k;s}$ and $L_{t,k;s}$. 
Consequently, we have the upper bound of the underlined part $(2)$ as that in Lemma \ref{lem:lostsales2} with $\bar{d}$ replaced with $\sum_{i=1}^{N}a_{ij}\bar{d}_{i}$. 
Thus, we have the upper bound of the lost sales part as 
\begin{align}
\label{eq:multilostsalesupper}
\mathbb{E}^{\pi}\left[\sum_{s=1}^{S}\sum_{t=1}^{T} \left(R_{t;s}- \tilde{R}_{t;s} \right)| \theta \right] \leq 
\sum_{j=1}^{M}p_{m}^{j}\sum_{i=1}^{N}a_{ij}\bar{d}_{i}\left(36T\sqrt{SK\log(K)} + S\sqrt{T}\right). 
\end{align}
Therefore, by combining \eqref{eq:multirevdif} with \eqref{eq:multilostsalesupper}, we have the desired upper bound.  
\end{proof}

\section{Additional Experimental Results}
\label{sec:addnum}
In this section, we present numerical results that are not covered in Section \ref{sec:Numa} 
to further validate the performance of our proposed algorithms. 
These results come from Poisson demand distributions with other mean demand parameters 
$\{\lambda(t,p)\}_{t\in [t], p \in \mathcal{P}}$, which increases with time.
We also provide numerical results for settings with negative binomial distributions, 
which can have larger variance than Poisson distributions.

\subsection{Another Poisson Distribution Setting}

\subsubsection{Experimental Settings}
The numerical experimental setting in this section has two different points from one in Section \ref{sec:Numa}. 
The first difference is the set of the true mean demand $\{\lambda(t,p)\}_{t\in[T], p\in \mathcal{P}}$ 
of Poisson distributions,   
which are set to $ \lambda(t,p) = 50\exp\left(- \frac{p}{\frac{1}{2} + \frac{5t}{T}}\right)$. 
This mean demand increases monotonically with time unlike the one introduced in Section \ref{sec:Numa}
which decreases with time. 
This demand function is a similar function to that used in \citet{anjos2005optimal, malighetti2009pricing}. 
The second difference is the number of episodes in the experiments for the Independent Prior setting. 
In this case, we conducted the experiment for $S=2000$.

\subsubsection{Numerical Results}
\begin{figure*}[t!]
    \includegraphics[width=1.0\textwidth]{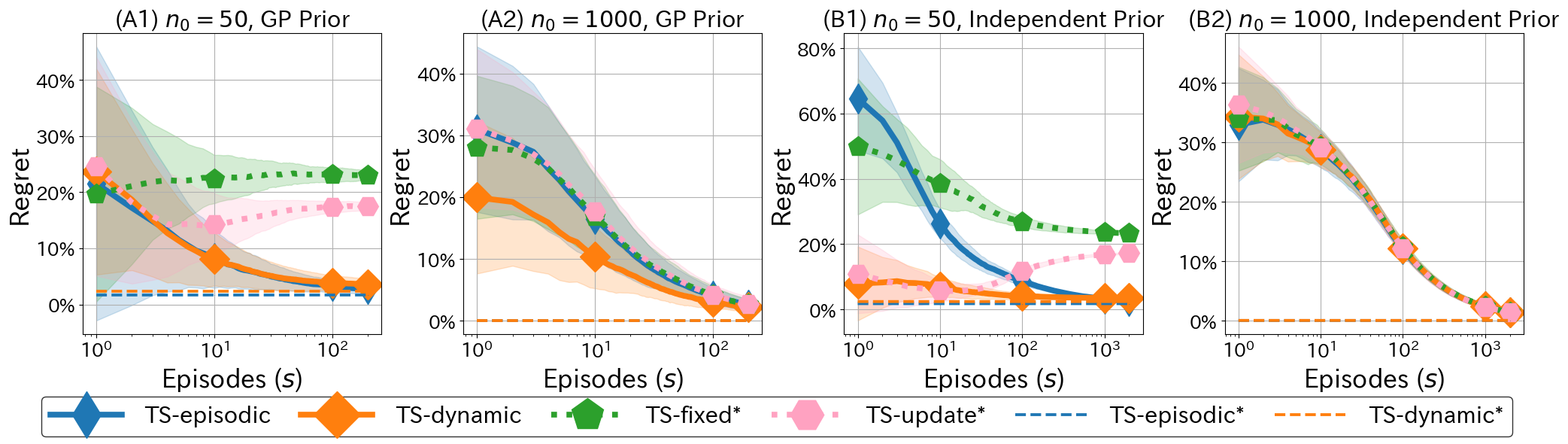}
       \caption{The numerical results for regret of \algone{}, \algtwo{}, {\sf TS-fixed}*, and {\sf TS updated}*,  \algone{}*, and \algtwo{}*. 
       (A1) and (A2) show the results for the GP prior, and (B1) and (B2) show the results for the independent prior.
        (A1) and (B1) show the results for $n_0=50$, and (A2) and (B2) show those for $n_0=1000$.
        The lines represent the averages of the regret and the shaded regions indicate the standard errors across independent $100$ trials. 
        The standard errors for \algone{}* and \algtwo{}* are omitted here and given in Appendix~\ref{sec:benches}.
     }
     \label{fig:benchpoissonv2}
  \end{figure*}

The numerical result is shown in Figure~\ref{fig:benchpoissonv2}, 
which demonstrates that the behavior of the relative regret for the algorithms there 
is almost the same as that shown in Figure~\ref{fig:bench1}. 
To be specific, our proposed algorithms outperform the benchmark algorithms in the case of $n_{0}=50$, 
and have almost the same performance as the benchmark algorithms in the case of $n_{0}=1000$. 
These results show that \algtwo{} can learn a pricing policy more effectively than \algone{},
which is consistent with the results in Figure~\ref{fig:bench1}.  
However, \algtwo{}* has a slightly worse performance than \algone{}* on average 
in the case of $n_{0}=50$.

\subsection{Negative Binomial Settings}
\label{sec:negnum}
A negative binomial distribution $N(k;r,p)$ is followed by the number of failures $k$ in a sequence of Bernoulli trials with the success probability $p$ 
until the trials succeed $r$ times, that is, 
\begin{align*}
N(k;r,p) = \binom{k+r-1}{r-1} p^{r}(1-p)^{k}.  
\end{align*}
This distribution has the mean $\frac{(1-p)r}{p}$ and the variance $\frac{(1-p)r}{p^2}$ and then
has the variance-to-mean ratio $1/p >1$, which is bigger than that of Poisson distributions. 
This setting enables us to examine the performance of our proposed algorithms in a more dispersed environment.

\subsubsection{Experimental Settings}
We consider the same price sets and time horizons as in the numerical setting in Section~\ref{sec:Numa}. 
The true demand distribution is set to negative binomial distributions
$N(k;r,p_{t,k})$ for $t\in[T]$ and $p \in \mathcal{P}$, 
where we assume that the learner knows the parameter $r=10$ 
but does not know the true success probabilities $\{p_{t,k}\}_{t\in[T], p\in \mathcal{P}}$ in advance. 
Here, we use two different true success probabilities
\begin{align}
   \label{eq:successprob}
\text{(PA):}\quad p_{t,k} = 1 - \exp\left(-\frac{t+p_k}{10}\right) \text{ and } 
\text{(PB):}\quad p_{t,k} = 1 - \exp\left(-\frac{p_k}{\left(\frac{1}{2} + \frac{5t}{T}\right)}\right). 
\end{align}
The initial inventory is set to $n_0 = 1000$ and $n_0 = 30$, 
corresponding to the cases where there is enough inventory and where the inventory is quite limited, respectively. 
For the prior on the demand distributions in the proposed algorithms and benchmarks,
we use independent beta distributions ${\rm Beta}(a,b)$ with shape parameters $a=1$ and $b=1$, 
and the success probabilities $\{p_{t,k}\}_{t\in[T], k\in [K]}$ are independently 
and identically distributed for all $k\in[K]$ and $t \in [T]$. 
Since beta distributions are conjugate to negative binomial distributions, the posterior distribution 
remains a beta distribution, which can be easily computed.  
We considered the number of episodes $S = 5000$ and computed all the results based on $100$ independent trials. 

\subsubsection{Numerical Results}

\begin{figure*}[t!]
   \includegraphics[width=1.0\textwidth]{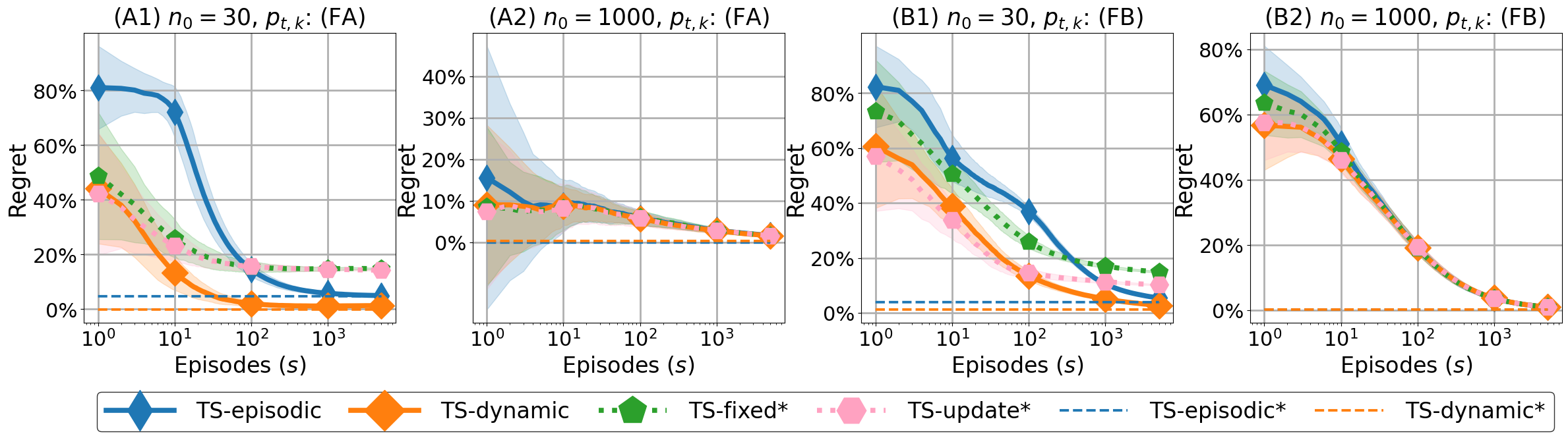}
      \caption{The numerical results for regret of \algone{}, \algtwo{}, {\sf TS-fixed}*, and {\sf TS updated}*,  \algone{}*, and \algtwo{}*. 
      (A1) and (A2) show the results for the success probability (PA) introduced in \eqref{eq:successprob}, and (B1) and (B2) show the results for the success probability (PB) in \eqref{eq:successprob}. 
       (A1) and (B1) show the results for $n_0=30$, and (A2) and (B2) show those for $n_0=1000$.
       The lines represent the averages of the regret and the shaded regions indicate the standard errors across $100$ independent trials. 
       The standard errors for \algone{}* and \algtwo{}* are omitted here and given in Appendix~\ref{sec:benches}.
    }
    \label{fig:benchnegbinom}
 \end{figure*}

Figure~\ref{fig:benchnegbinom} (A2) and (B2) for the case $n_0 = 1000$ show
that our proposed and the benchmark show almost the same performance due to the same reason discussed in Section~\ref{sec:Numa}. 
Figure~\ref{fig:benchnegbinom} (A1) and (B1) show the results for the case of $n_0 =30$. 
They demonstrate that the proposed algorithms successfully learn an efficient pricing policy 
to achieve small regrets unlike the benchmark algorithms. 
Note that for the both different mean demand parameters (PA) and (PB),  
\algtwo{} learns an efficient pricing policy faster than \algone{} and shows better performance as expected.

\section{Details of the Numerical Analysis}
\label{sec:appenum}
\subsection{Comparison Targets}
\label{sec:benches}
We introduce the benchmark algorithms used in Section \ref{sec:Numa}. 
We first present {\sf TS-fixed}* and {\sf TS-update*}, which solve 
the linear optimization problem  ${\rm LP_{avg}}\left(\lambda(\theta), t, \tau, n\right)$ over $\{ x_{k}\}_{k\in[K]} \in [0,1]^{K}$, defined as follows: 
\begin{align*}
\max_{\{x_{k} | k\in[K]\} } &\sum_{k=1}^{K}\lambda_{t,k}(\theta)p_{k}x_{k} 
\\
\text{subject to} &\sum_{k=1}^{K} x_{k}\lambda_{t,k}(\theta) \leq \frac{n}{\tau}, 
\\
&\sum_{k=1}^{K}x_{k} \leq 1. 
\end{align*}
This LP maximizes instantaneous revenue subject to the average inventory
over $\tau$ periods, 
which is a kind of LP relaxation of the revenue optimization problem in the static demand setting.
The solution of $x_{k}(\theta)$ corresponds to the probability of choosing price $p_{k}$.

{\sf TS-fixed}* in Algorithm \ref{alg:Fixed*} samples a demand parameter $\theta_{s}$ 
from the posterior distribution at the beginning of each episode. 
{\sf TS-fixed}* decides a price based on the solution of ${\rm LP_{avg}}(\lambda(\theta_s), t, T, n_{0})$. 
This LP considers the instantaneous revenue at time period $t$ 
subject to the fixed average inventory $n_{0}/T$. 
This constraint means that the initial inventory can be uniformly allocated across the selling season. 
{\sf TS-Update}* in Algorithm \ref{alg:Update*}, on the other hand, decides a price with inventory updated sequentially. 
The algorithm solves ${\rm LP_{avg}}(\lambda(\theta_s), t, T-t+1, n_{t-1})$. 
This constraint means that the remaining inventory can be uniformly allocated across the remaining selling season.

We aim to investigate the effect of an estimate of future inventory consumption on total revenue in the dynamic demand scenario.
To know the effect, we set the sampling timings for {\sf TS-fixed}* and {\sf TS-updated*} to be identical 
to that of \algone{}.  
Then, the difference in the performance mainly arise from solving the LP. 
The numerical results in Section \ref{sec:Numa} can effectively 
demonstrate the effect of estimation of future demand consumption on total revenue in the dynamic demand scenario.

Next, we explain the oracle algorithms \algone{}* and \algtwo{}*, 
given in Algorithm \ref{alg:TS-episodic*} and Algorithm \ref{alg:TS-dynamic*}. 
They determine a schedule of prices by solving the LP with the true demand parameter instead of a randomly sampled one.  
Their expected revenue is independent of the number of episodes since they do not need learning. 
The performance of the oracle algorithms limits that of \algone{} and \algtwo{} 
since \algone{} and \algtwo{} can reach the oracle algorithms' performance after sufficient episodes. 
The details of the numerical results of the relative regret for our numerical experiments are given in Table~\ref{tab:avgandstd} to \ref{tab:avgandstd_4} 
at the end of this section. 

Finally, we briefly explain the optimal pricing policy in hindsight $\pi^*(\theta)$ and its expected revenue ${\rm Rev}^*(T, \theta)$. 
We compute them through dynamic programming based on the recurrence relation of ${\rm Rev}^*(t, \theta)$ for any elapsed time $t \in [T]$ 
\begin{align}
  {\rm Rev}^*(t, \theta, n_{T-t}) = \max_{p \in \mathcal{P} \cup \{p_{\infty}\}}\mathbb{E}\left[\left. p \tilde{D}_{T - t + 1} + {\rm Rev}^*(t - 1, \theta, n - \tilde{D}_{T - t + 1}) \right| P_{t} = p, \theta \right],
\end{align}
where the expectation is taken over $D_{t-T+1}$, ${\rm Rev}^*(t, \theta, 0) =0 $ for $t \in [T]$, ${\rm Rev}^*(0, \theta, m) =0 $ for $m\geq0 $,
 and the dependency on the remaining inventory is revealed for clarity. 
The optimal pricing policy $\pi^*(\theta)$ then consists of the prices that maximize the expectation in the above relation for each elapsed time. 
We show the values of ${\rm Rev}^*(T, \theta)$ in our numerical experiments in Table~\ref{tab:optrev} at the end of this section.

\subsection{Details of Gaussian Process}
\label{sec:gaussianP}
For any given data set $\mathcal{D} = \cup_{i=1}^{\ell} \{(t_{i}, p_{i}, d_{i})\}$, 
when a function $g$:$[T] \times \mathcal{P} \mapsto \mathbb{R}$ follows a gaussian process with a mean function $\mu$ and kernel function $k$, 
then the $\ell$ values of $g$ follows the $\ell$ multivariate Gaussian distribution $\mathcal{N}\left(\bm{g} | \bm{m}, \bm{K}\right)$,
where
\begin{align*} 
\bm{g} =& (g_{1}=g(t_1.p_{1}), \dots, g_{\ell}=g(t_{\ell}, p_{\ell})),\\
\bm{m}= &(\mu_{1}=\mu(t_{1}, p_{1}), \dots, \mu_{\ell}=\mu(t_{\ell}, p_{\ell})), \\
\bm{K} = &\{K_{ij}=K\left((t_{i},p_{i}), (t_{j}, p_{j})\right)\}_{1\leq i,j \leq \ell }. 
\end{align*}
The set of intensities $\{\lambda_{t,k}\}_{t\in[T], k \in [K] }$ is modeled with $g$ following a Gaussian process 
as $\lambda_{t,k}= \exp(g(t,p_{k})) \in (0, \infty)$ for all $t\in[T]$ and $k \in [K]$. 
The posterior distribution of $g$ for a given data set $\mathcal{D}= \cup_{i=1}^{\ell}\{(t_{i}, p_i, d_{i})\}$ is expressed as $\mathbb{P}\left( \bm{g}|\mathcal{D}\right) 
\propto  \prod_{i=1}^{\ell}\left(\text{Poi}\left(d_{i}|\exp(g(p_{i},t_{i}))\right)\right)\mathcal{N}\left(\bm{g} | \bm{m}, \bm{K}\right)$. 
This posterior distribution has no closed form expression and is intractable
and thus we approximated it to a Gaussian distribution with Laplace approximation.

\subsection{Comments on the Implementation}
\subsubsection{Computing Instrument}
The details of our computation instrument is following: 
\begin{description}
\item{OS: Ubuntu 22.04.2 LTS}
\item{CPU model: Intel(R) Xeon(R) Silver 4210 CPU @ 2.20GHz}
\item {Amount of Memory: 128GB}.
\end{description}

The software frameworks for out implementation are following: 
\begin{description}
\item {Python \footnote{https://www.python.org/}: 3.8.13}
\item {Numpy \footnote{https://numpy.org/doc/1.20/contents.html}: 1.20.3}
\item {GPy \footnote{http://sheffieldml.github.io/GPy/}: 1.10.0}
\item {matplotlib \footnote{https://matplotlib.org/}: 3.5.1}
\item {Gurobi \footnote{\citep{gurobi}}: 10.0.0}
\end{description}

\subsubsection{Gaussian Prossess Framework}
The gaussian process prior is implemented using GPy, which is a Gaussian process framework in Python.
This framework covers posterior sampling from a poisson likelihood function with Laplace approximation.

\subsubsection{Gamma Prior Framework }
The independent gamma prior is implemented using Numpy. 
Numpy provides us for a sampling method from a gamma distribution, and its randomness is controlled by setting a seed.

\subsubsection{Solvers of Linear Optimization Problems.}
We solve the linear programming with the GUROBI optimizer, which is a commercial solver. 
We use this solver just for computational time efficiency.  
Another non-commercial solver including (cvxopt \footnote{https://cvxopt.org/}: version 1.2.6) 
can solve the linear programming for the problem size we used in this paper.

\begin{algorithm}[t!]
  \caption{ {\sf TS-fixed*}}
  \label{alg:Fixed*}
  \For{$s=1,\dots, S$} 
  {Sample a demand parameter $\theta_{s} \in \Theta $ from the posterior distribution $f(\cdot| H_{0}^{s})$ of $\theta$.\; 
  \For{$t=1,\dots, T$}
  {
  Solve ${\rm LP_{avg}}(\lambda(\theta_s), t, T, n_{0})$.\;
  Offer price $P_{t;s}= p_{k}$ with probability $x_{k}$
  or offer  $P_{t;s}=p_{\infty}$ with probability $1 - \sum_{k=1}^{K}x_{k}$.\;
  Observe realized demand $D_{t;s}$ and update history  $H_{t}^{s} = H_{t-1}^{s} \cup \{P_{t;s}, D_{t;s}\} $.\;
  }
  }
  \end{algorithm}
  
  \begin{algorithm}[t!]
  \caption{{\sf TS-Update*}}
  \label{alg:Update*}
  \For{$s=1,\dots, S$}
  {
  Sample a demand parameter $\theta_{s} \in \Theta $ from the posterior distribution of $\theta$.\; 
  \For{$t=1,\dots, T$}
  {
  Solve ${\rm LP_{avg}}(\lambda(\theta_{s}),t, T-t+1, n_{t-1;s})$.\;
  Offer price $P_{t;s}= p_{k}$ with probability $x_{k}$
  or offer  $P_{t;s}=p_{\infty}$ with probability $1 - \sum_{k=1}^{K}x_{k}$. \;
  Observe realized demand $D_{t;s}$ and update history  $H_{t}^{s} = H_{t-1}^{s} \cup \{P_{t;s}, D_{t;s}\} $.\;
  }
  }
  \end{algorithm}

  \begin{algorithm}[t!]
  \caption{\algone{}*}
  \label{alg:TS-episodic*}
  Solve LP$(\theta, 1,n_0)$.\;
  \For{$t=1,\dots, T$}
  {
  Offer price $P_{t}= p_{k}$ with probability $x_{k,t}(\theta)$
  and $P_{t}=p_{\infty}$ with probability $1 - \sum_{k=1}^{K}x_{t,k}(\theta)$.\;
  Observe realized demand $D_{t}$.\;
  }
  \end{algorithm}

\begin{algorithm}[t!]
  \caption{\algtwo{}*}
  \label{alg:TS-dynamic*}
  \For{$t=1,\dots, T$}
  {
  Solve LP$(\theta, t, n_{t-1})$.\;
  Offer price $P_{t}= p_{k}$ with probability $x_{t,k}(\theta)$
  and $P_{t}=p_{\infty}$ with probability $1 - \sum_{k=1}^{K}x_{t,k}(\theta)$.\;Observe realized demand $D_{t}$.
  }
\end{algorithm}

\begin{table}[t!]
  \centering
  \begin{tabular}{lrrrr} 
  \toprule
      Algorithms       &  \multicolumn{2}{c}{$n_{0}= 50$} & \multicolumn{2}{c}{$n_{0}= 1000$}
       \\ \midrule
             &   Average (\%) & Standard error (\%) & Average (\%) &  Standard error (\%)\\
  \algone{}*  &  2.63 & 8.59 &  0.07 & 11.83  \\
  \algtwo{}*  & 1.27 & 8.78  &  -0.09 & 11.80 \\ 
  \bottomrule
  \end{tabular}
  \caption{This table shows the empirical results of the relative regrets, $(1 - {\rm Rev}^{\pi}(T,\theta)/ {\rm Rev}^{*}(T,\theta))$, in Figure~\ref{fig:bench1} for \algone{}* and \algtwo{}*. 
  The average and standard errors are obtained across independent 10000 trials. }
  \label{tab:avgandstd}

  \centering
  \begin{tabular}{lrrrr} 
  \toprule
      Algorithms       &  \multicolumn{2}{c}{$n_{0}= 50$}   & \multicolumn{2}{c}{$n_{0}= 1000$} \\ 
      \midrule
             &   Average (\%) & Standard error (\%) & Average(\%)  &  Standard error (\%)\\
  \algone{}* & 1.73 & 8.15 &  0.02 & 8.37  \\
  \algtwo{}* & 2.39 & 6.90 &  0.03 & 8.36 \\ 
  \bottomrule
  \end{tabular}
  \caption{This table shows the empirical results of the relative regrets in Figure~\ref{fig:benchpoissonv2} for \algone{}* and \algtwo{}*. 
  The average and standard errors are obtained across 200000 trials.}
  \label{tab:avgandstd_2}

    \centering
    \begin{tabular}{lrrrr} 
    \toprule
        Algorithms  &  \multicolumn{2}{c}{$n_{0}= 30$}   & \multicolumn{2}{c}{$n_{0}= 1000$} \\ 
        \midrule
               &   Average (\%) & Standard error (\%) & Average(\%)  &  Standard error (\%)\\
    \algone{}* &  4.72 & 12.68 &  -0.11 & 17.63  \\
    \algtwo{}* &  -0.14 & 8.60 &  0.28 & 17.74  \\ 
    \bottomrule
    \end{tabular}
    \caption{This table shows the empirical results of the relative regrets in Figure~\ref{fig:benchnegbinom} (A1) and (A2)  for \algone{}* and \algtwo{}*. 
    The average and standard errors are obtained across 10000 trials.}
    \label{tab:avgandstd_3}

    \centering
    \begin{tabular}{lrrrr} 
    \toprule
        Algorithms       &  \multicolumn{2}{c}{$n_{0}= 30$}   & \multicolumn{2}{c}{$n_{0}= 1000$} \\ 
        \midrule
                &   Average (\%) & Standard error (\%) & Average(\%)  &  Standard error (\%)\\
    \algone{}* &  3.92 & 12.12 &  0.20 & 12.67 \\
    \algtwo{}* &  1.24 & 11.48 &  0.20 & 12.67 \\ 
    \bottomrule
    \end{tabular}
    \caption{This table shows the empirical results of the relative regrets in Figure~\ref{fig:benchnegbinom} (B1) and (B2) for \algone{}* and \algtwo{}*. 
    The average and standard errors are obtained across 10000 trials.}
    \label{tab:avgandstd_4}
    \centering
    \begin{tabular}{lrrrrrrrr} 
    \toprule
                &  \multicolumn{2}{c}{Figure~\ref{fig:bench1}}   & \multicolumn{2}{c}{Figure~\ref{fig:benchpoissonv2}} & \multicolumn{2}{c}{Figure~\ref{fig:benchnegbinom} (PA)}  & \multicolumn{2}{c}{Figure~\ref{fig:benchnegbinom} (PB)}   \\ 
        \midrule
                &   $n_0 = 50$ & $n_0=1000$ & $n_0=50$  &  $n_0 = 1000$  & $n_0 = 30$  &  $n_0=1000$ & $n_{0}=30$ & $n_0=1000$  \\
    ${\rm Rev^{*}}(T,\theta)$ &  330.08  & 359.18 &  383.30  &  594.30  & 258.75 & 320.35 & 141.36 & 278.34 \\ 
    \bottomrule
    \end{tabular}
    \caption{This table shows the values of the optimal expected total revenue ${\rm Rev}^{*}(T, \theta)$ for all our numerical experiments.}
    \label{tab:optrev}
\end{table}

\end{document}